\documentclass[lettersize,journal]{IEEEtran}
\usepackage[utf8]{inputenc}
\usepackage[english]{babel}
\usepackage{microtype}
\usepackage{geometry}
\usepackage{amsmath}
\usepackage{amsfonts}
\usepackage{amssymb}
\usepackage{mathtools}
\usepackage{amsthm}
\theoremstyle{plain}
\newtheorem{theorem}{Theorem}[section]

\newtheorem{corollary}[theorem]{Corollary}
\theoremstyle{definition}

\newtheorem{assumption}[theorem]{Assumption}
\theoremstyle{remark}
\newtheorem{remark}[theorem]{Remark}
\usepackage{graphicx}
\usepackage{subcaption}
\usepackage{textcomp}
\usepackage{booktabs}
\usepackage{stfloats}
\usepackage{array}
\usepackage{algorithm}
\usepackage{algorithmic}
\usepackage{listings}
\usepackage{xcolor}
\usepackage{cite}
\usepackage{url}
\usepackage{verbatim}
\usepackage{hyperref}
\usepackage[capitalize,noabbrev]{cleveref}
\hypersetup{
    colorlinks=true,
    linkcolor=blue,
    citecolor=blue,
    urlcolor=blue
}
\usepackage{listings}
\usepackage{paralist}
\usepackage{verbatim}
\usepackage{textcomp}
\usepackage{float}
\usepackage{listings}
\usepackage{xcolor}
\definecolor{terminalbg}{rgb}{0.95, 0.95, 0.95}
\definecolor{terminalfg}{rgb}{0, 0, 0}

\lstdefinelanguage{Terminal} {
    basicstyle=\ttfamily\footnotesize,
    backgroundcolor=\color{terminalbg},
    frame=single,
    rulecolor=\color{black},
    showstringspaces=false,
    breaklines=true
}
\IEEEoverridecommandlockouts
\usepackage{threeparttable}

\begin{document}
\title{LLM-QFL: Distilling Large Language Model for Quantum Federated Learning}
\author{Dev Gurung and Shiva Raj Pokhrel
\thanks{D.~Gurung and S.~R.~Pokhrel are with School of IT, Deakin University, Australia. dev.gurung@deakin.edu.au, shiva.pokhrel@deakin.edu.au}
\thanks{Manuscript received April 19, 2025; revised August 16, 2025.}}

\markboth{Journal of \LaTeX\ Class Files,~Vol.~14, No.~8, August~2025}%
{Shell \MakeLowercase{\textit{et al.}}: A Sample Article Using IEEEtran.cls for IEEE Journals}

\maketitle
\begin{abstract}
Inspired by the power of large language models (LLMs), our research adapts them to quantum federated learning (QFL) to boost efficiency and performance. We propose a federated fine-tuning method that distills an LLM within QFL, allowing each client to locally adapt the model to its own data while preserving privacy and reducing unnecessary global updates. The fine-tuned LLM also acts as a reinforcement agent, optimizing QFL by adjusting optimizer steps, cutting down communication rounds, and intelligently selecting clients. Experiments show significant efficiency gains. We pioneer a synergy between LLM and QFL, offering: i) practical efficiency: Reduced communication costs and faster convergence.
    ii) theoretical rigor: Provable guarantees for adaptive federated optimization.
    iii) scalability: PEFT methods (LoRA, QLoRA) enable deployment on resource-constrained quantum devices.
Code implementation is available \href{https://github.com/s222416822/llmQFL}{here}\footnote{https://github.com/s222416822/llmQFL}.
\end{abstract}


\section{Introduction}
The convergence of {quantum machine learning} and {large language models (LLMs)} represents a transformative frontier in artificial intelligence (AI) research, offering unprecedented potential for enhancing computational efficiency and intelligence \cite{yeOpenFedLLMTrainingLarge2024a, YeTowardsQuantumMachineLearning2023ICML, liQuantumSupportVector2024}. In this work, we explore the {integration of LLMs over Quantum Federated Learning (QFL)}~\cite{pokhrel2024quantum}, leveraging their capabilities to advance both {distributed quantum computing} and generative AI.

LLMs have demonstrated remarkable proficiency in {understanding and generating human-like responses}, leading to their widespread adoption across diverse applications \cite{yinSurveyMultimodalLarge2024}. Their {development and deployment} present {significant computational challenges} and {ethical concerns}, particularly in {privacy-sensitive and resource-constrained environments}~\cite{guo2025deepseek}. Meanwhile, {quantum computing} offers a paradigm shift, capable of overcoming {classical computational bottlenecks} by processing vast amounts of data at {unprecedented speeds}. By integrating {LLMs into QFL}, we seek to harness the advantages of {both technologies}, where {LLMs can enhance QFL optimization}, and {QFL accelerate LLM training and inference} \cite{wuNetLLMAdaptingLarge2024}.

\begin{figure}
    \centering
    \includegraphics[width=0.68\linewidth]{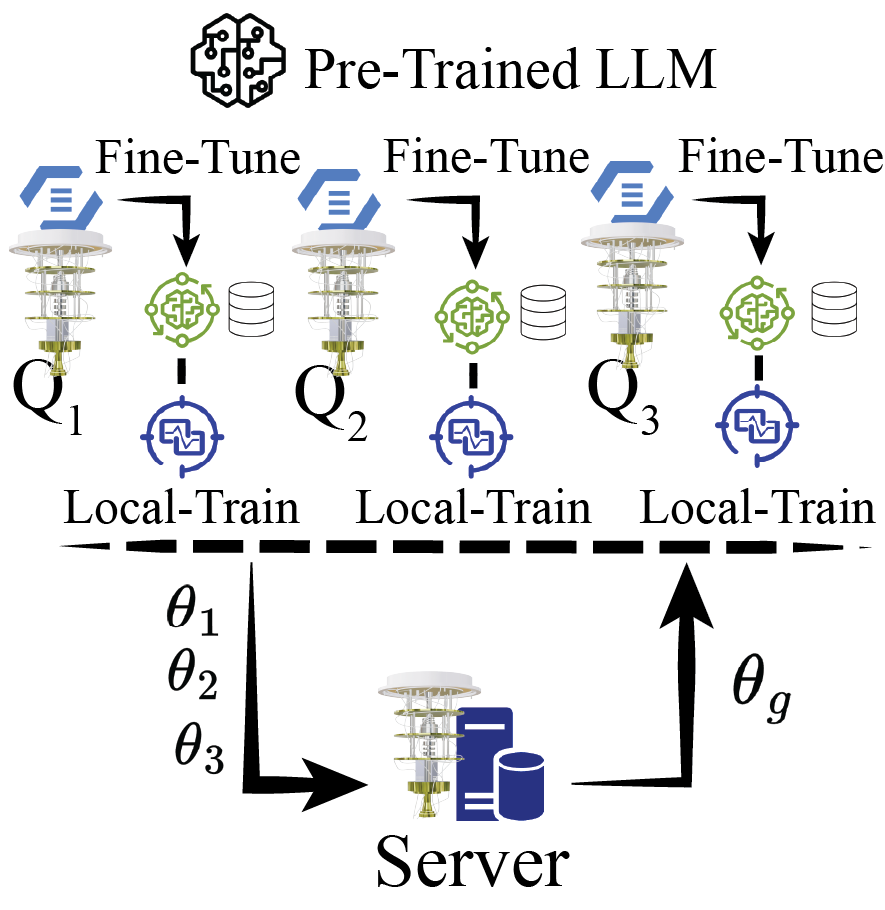}
    \caption{Distilling LLMs over QFL: Locally Fine-Tuned LLMs for enhanced QFL Performance}
    \label{fig:QFL-LLM}
\end{figure}

Despite their potential, key challenges remain in integrating LLMs within the QFL framework. The feasibility of leveraging classically trained LLMs to enhance quantum machine learning remains unexplored, and optimizing QFL algorithms through LLM fine-tuning to mitigate computational and communication bottlenecks requires further investigation \cite{wuFedBiOTLLMLocal2024}. Additionally, with reasoning LLMs like DeepSeek-R1 \cite{guo2025deepseek} accelerating large-scale reinforcement learning, high-quality public training data is expected to be exhausted by 2026, posing a major limitation. This raises the critical question of how to harness privately distributed, high-quality data across organizations while preserving data privacy and security.

\subsection{Literature}
In recent years, there has been great interest in the field of QFL~\cite{pokhrel2024data} followed by integration of LLM into Federated learning. 

In QFL, Xi et al. \cite{xiaQuantumFedFederatedLearning2021b} introduced the FL framework for collaborative quantum training across multiple quantum nodes with local data. Expanding on this, Chehimi et al. \cite{chehimiQuantumFederatedLearning2022} explored training on quantum data, while Li et al. \cite{liQuantumFederatedLearning2021} proposed a privacy-preserving framework using blind quantum computing with a single quantum server. 
To enhance optimization, Qi et al. \cite{qiFederatedQuantumNatural2022} developed a federated quantum natural gradient descent algorithm, and Yun et al. \cite{yunSlimmableQuantumFederated} introduced slimmable QFL to adapt to varying communication and computational constraints. 
A hybrid quantum-classical approach was proposed by Chen et al. \cite{chenFederatedQuantumMachine2021}, integrating a quantum neural network with a pre-trained classical model~\cite{pokhrel2024quantum}. 
Addressing non-IID data challenges, Zhao et al. \cite{zhaoExactDecompositionQuantum2022a} introduced \textit{qFedInf} for one-shot complexity in non-IID scenarios, while Huang et al. \cite{huangQuantumFederatedLearning2022} and Gurung et al. \cite{gurung2024personalized} focused on decentralized learning to improve communication efficiency in non-IID quantum settings.
In recent times, there has been an increasing number of studies in the field, as referenced in \cite{gurungChainedContinuousQuantum2025, gurungBQFLMetaverse, gurungPerformanceAnalysisDesign2025}, suggesting different QFL algorithms such as a chained method, specific application contexts, and a personalized methodology, respectively.

In terms of LLM integration to FL, 
Hou et al. \cite{houPrETextTrainingLanguage2024} proposed Private Evolution-Text (PrE-Text) method to produce differentially private synthetic data to train language models on private federated data setting, blackbox prompt tuning approach FedBPT by Sun et al. \cite{sunFedBPTEfficientFederated2024} and federated full parameter tuning by Qin et al. \cite{qinFederatedFullParameterTuning2024}.
Wu et al. \cite{wuNetLLMAdaptingLarge2024} studied LLM adaptation for networking and presented the NetLLM framework to address networking problems by using the powerful capabilities of LLM.
Ye et al. \cite{yeOpenFedLLMTrainingLarge2024a} build a concise integrated framework named openFedLLM.
Similarly, Wu et al. \cite{wuFedBiOTLLMLocal2024} proposed FedBiOT, an LLM local fine-tuning in FL without the full model to address the distributed nature of data across multiple users. 
Qi et al. \cite{qiFDLoRAPersonalizedFederated2024} proposed a personalized FL of LLM via dual-lora tuning (FDLoRA).
Chen et al. \cite{chenFedDATApproachFoundation2024} proposed a fine-tuning framework tailored to heterogeneous multimodal FL, called FedDAT.
Wang et al. \cite{wangFLoRAFederatedFineTuning2024} proposed FLoRa, a federated fine-tuning LLM with heterogeneous low-rank adaptations that allows federated fine-tuning on heterogeneous LoRA adapters.
Zhang et al. \cite{zhangFedPETuningWhenFederated2023} proposed FedPETuning, a federated learning-based parameter-efficient tuning method of pre-trained language models.
The other works include pFedPrompt \cite{guoPFedPromptLearningPersonalized2023}, personalized federated learning-based prompt learning for vision language models in FL, scaling FL for fine-tuning of LLMs \cite{hilmkilScalingFederatedLearning2021}, 
Titanic \cite{suTitanicProductionFederated2024},
recovering private text in FL of LLMs \cite{guptaRecoveringPrivateText2022}.

In this paper, we introduce a quantum-based environment for LLM fine-tuning in a distributed setting. Unlike previous studies, mentioned above, we explore a unique architectural framework and specialized tools to analyze the nature of quantum distributed networks, enabling a deeper understanding of their structure, optimization dynamics, and scalability in FL settings.
\subsection{Contributions}

To address these challenges, we propose a federated distillation and fine-tuning approach that integrates LLM within QFL, ensuring privacy-preserving, scalable, and communication-efficient learning. We have the following contributions:

\begin{itemize}
    \item \textbf{Federated Fine-Tuning and Distillation of LLMs in QFL}.
     We propose a new method to safely adapt LLMs within QFL while preserving privacy.
 Each quantum device fine-tunes the LLM locally, reducing the need for frequent global updates.
 Knowledge distillation (e.g., using KL divergence) keeps local models aligned with a central global LLM.

    \item \textbf{LLM as a Smart Controller for QFL}.
 The LLM dynamically adjusts optimizer steps (e.g., COBYLA iterations) to optimize training.
 It selects clients with performance closest to the global average, based on deviation $d_i^{t} = |L_i^{t} - L_s^{t}|$.
 Training stops early when improvements are small, i.e., $\frac{\Delta L_s^{t}}{L_s^{t}} < \epsilon$.

    \item \textbf{Theoretical Guarantees}.
 We prove convergence with a rate of $\mathcal{O}(\frac{1}{T})$ under standard assumptions like L-smoothness and bounded variance.
 The method reduces communication rounds by $\frac{\mathbb{E}[K^{t}]}{K}$ and variance by $\left(1 - \frac{k}{N}\right)$.

    \item \textbf{Experimental Results}.
Our method speeds up training on both genomic (DemoHumanOrWorm) and language (TweetEval) tasks.
We validate the approach on IBM Quantum hardware and simulators (AerSimulator, FakeManila).
 Adaptive optimization reduces idle computation by 30\% (see Fig.~\ref{fig:impact_on_ratio_maxiter}).
\end{itemize}

\subsubsection{Overview of Performance Evaluation}
Our study leverages state-of-the-art (SOTA) LLMs and the latest Qiskit Aer simulators, conducting simulations and experiments on real IBM quantum machines using Qiskit, and Google Colab with diverse QPU/GPU configurations. We utilize genomic~\cite{grevsova2023genomic} and TweetEval-sentiment datasets~\cite{rosenthal2017semeval} to perform parameter-efficient distillation and tuning using Low-Rank Adaptation (LoRA), integrating LLMs such as Meta-LLAMA 3.2-1B~\cite{touvron2023llamaopenefficientfoundation}, GPT-2~\cite{radford2019language}, and DeepSeek-LLM-7B-Base~\cite{guo2025deepseek}. Training is conducted on 3–10 quantum devices with both LoRA and quantized LoRA variations. The execution details over real IBM Quantum machines are summarized below, with further elaboration provided in the latter sections of the paper.

\subsubsection{Execution of QFL on a Real Quantum Computer} 
To evaluate the performance of our LLM-QFL framework, we executed our implementation on a real IBM quantum computer. While quantum simulators like Qiskit AerSimulator provide a controlled environment, real quantum hardware introduces practical constraints such as \textit{quantum noise, decoherence, and hardware-specific impairments}, making it essential for a experimental evaluation.  

\begin{figure}[t]
    \centering
    \includegraphics[width=\linewidth]{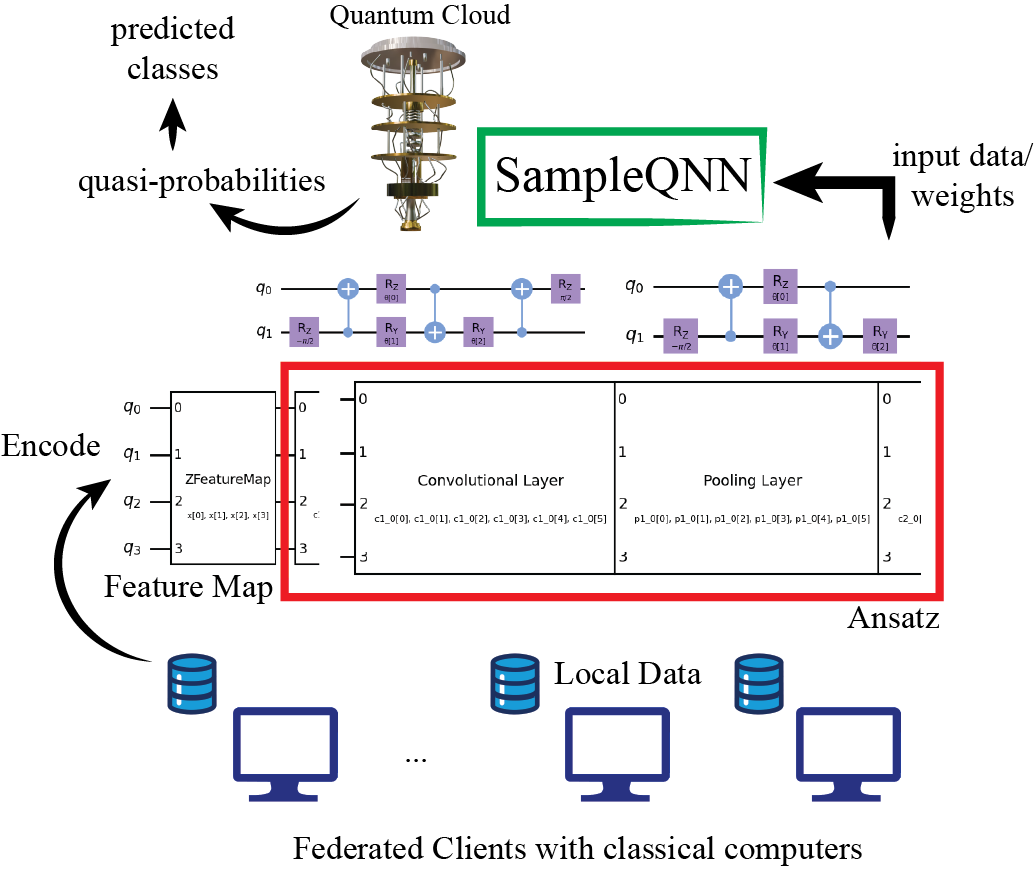}
    \caption{Execution workflow of LLM-QFL on a real IBM quantum computer, detailing data encoding, quantum circuit selection, and result interpretation.}
    \label{fig:qfl_on_r}
\end{figure}

After encoding the data and selecting the quantum circuit structure, we submitted computational jobs to a real quantum processor, as illustrated in Figure~\ref{fig:qfl_on_r}. The input data and parameters were processed using the \textit{SamplerQNN} neural network, which employs a parameterized quantum circuit to encode input features and optimize trainable weights. The \textit{SamplerQNN} then translates the quasi-probabilities estimated by the sampler primitive into discrete predicted classes.  For effective classification, we applied a custom interpret function that computes the parity of bit strings, ensuring accurate label assignments. This approach enabled us to assess the practical feasibility of LLM-QFL  while addressing the real-world challenges associated with quantum computation.  
By integrating {LLMs within QFL}, we {bridge the gap between LLM, quantum intelligence and federated learning}, paving the way for a {scalable, efficient, and privacy-preserving paradigm}.

\section{Background and Preliminaries}
QFL integrates quantum machine learning with federated learning to leverage the advantages of both paradigms. In a typical QFL framework, multiple  $N$ clients 
with quantum processors collaboratively train a global quantum model \(\theta_g\) while keeping their local datasets \(D_i\) private. Each client \(c_i\) updates its local quantum model parameters \(\theta_i\) using a local loss function \(F_i(\theta_i)\) and shares the updated parameters with a central server. The server aggregates these updates to refine the global model using an aggregation rule such as,
\[
\theta_g^{t+1} = \sum_{i=1}^N w_i \theta_i^{t+1}
\]
where, \(w_i\) is the weighting factor for each client, and \(t\) denotes the iteration step. Local parameter updates follow
\[
\theta_i^{t+1} = \theta_i^{t} - \eta \nabla_{\theta_i} F_i(\theta_i^{t})
\]
where, \(\eta\) is the learning rate and \(\nabla_{\theta_i} F_i(\theta_i^{(t)})\) represents the gradient of the loss function with respect to local parameters. QFL extends classical federated learning (FL) principles while incorporating quantum computational advantages, such as enhanced representation power and quantum parallelism. Key differences between classical FL and QFL arise from the intrinsic differences between classical and quantum computing, including data encoding, optimization processes, and quantum noise resilience.

\subsection{LLMs, Distillations \& Fine-tuning}

LLMs have revolutionized the reasoning trained via large-scale reinforcement learning (DeepSeek R1~\cite{guo2025deepseek}) and the natural language processing (NLP) through transformer-based architectures~\cite{vaswaniAttentionAllYou2017}. These LLMs, trained on extensive corpora such as Wikipedia and online text, demonstrate strong generalization and reasoning capabilities across diverse tasks, including text generation, sentiment analysis, and multimodal applications. Other prominent LLMs include GPT-4, PaLM, and LLaMA, which leverage billions of parameters to learn complex linguistic patterns and perform reasoning-based tasks~\cite{zhaoSurveyLargeLanguage2023}.

LLMs undergo two primary training phases: (1) \textit{pre-training} on large-scale textual datasets, where they capture general linguistic structures and statistical relationships, and (2) \textit{fine-tuning} on specific tasks to enhance domain-specific performance. Once pre-trained, LLMs can be adapted to downstream applications such as text classification, summarization, and dialogue systems.

Despite LLMs' capabilities, distilling and fine-tuning them for specific tasks is computationally expensive due to their large parameter count. For instance, LLaMA-3.2-1B contains 1 billion parameters, while DeepSeek-LLM-7B-Base consists of 7 billion parameters. Training all parameters is impractical; thus, Parameter-Efficient Fine-Tuning (PEFT) techniques are employed to optimize adaptation while reducing computational overhead~\cite{hanParameterEfficientFineTuningLarge2024}.

PEFT methods, such as Low-Rank Adaptation (LoRA) and Quantized LoRA (QLoRA), minimize the number of trainable parameters while maintaining model performance. LoRA decomposes large matrices into low-rank components within attention layers, significantly reducing memory requirements~\cite{huLoRALowRankAdaptation2021a}. QLoRA extends this approach by quantizing LoRA adapter weights to lower precision (e.g., 4-bit instead of 8-bit), further decreasing memory footprint and computational costs. These techniques facilitate efficient fine-tuning of LLMs for specialized applications while optimizing resource utilization.

Knowledge distillation, a machine learning technique, aims to transfer knowledge from a large pre-trained model (teacher) to a smaller model (student) \cite{hintonDistillingKnowledgeNeural2015b, buciluaModelCompression2006Distillation}.
It has been extended to studies in the field of classical federated learning such as \cite{morafahPracticalRecipeFederated2024, wuCommunicationefficientFederatedLearning2022Distillation}, where the authors discuss efficient communication FL and address weaknesses of parameter-averaging FL algorithms.

\subsection{Scope and Settings}
We consider a QFL setting with $N$ clients, each possessing a local dataset $\mathcal{D}_i$ and training a quantum model. The local objective of client $i$ is to minimize the loss function

\begin{equation}
F_i(\boldsymbol{\theta}, \phi) = \frac{1}{|\mathcal{D}_i|} \sum_{(\mathbf{x}, \mathbf{y}) \in \mathcal{D}_i} \ell(N (\mathbf{x}; \boldsymbol{\theta}, \phi), \mathbf{y}),
\end{equation}
where, $\boldsymbol{\theta}$ represents the quantum model parameters, $\phi$ denotes the LLM parameters, and $\ell(\cdot, \cdot)$ measures prediction error. QFL seeks to find the optimal parameters $\boldsymbol{\theta}$ and $\phi$ by solving

\begin{equation}
\min_{\boldsymbol{\theta}, \phi} F(\boldsymbol{\theta}, \phi) = \sum_{i \in [N]} w_i F_i(\boldsymbol{\theta}, \phi),
\end{equation}

where, $w_i = \frac{|\mathcal{D}_i|}{|\mathcal{D}|}$ weighs client $i$'s dataset relative to the global dataset $\mathcal{D} = \bigcup_{i \in [N]} \mathcal{D}_i$. The server initializes global parameters $(\boldsymbol{\theta}^{t}, \phi^{t})$ at each communication round $t$, and clients update them locally:

\begin{equation}
\boldsymbol{\theta}_i^{t+1} = \sum_{i \in [N]} w_i \boldsymbol{\theta}_i^{t}.
\end{equation}

We emphasize two key aspects: (a) fine-tuning LLMs over QFL to improve both local and global performance and (b) leveraging LLM distillation to enhance QFL efficiency. 

\begin{remark}[Lower Loss \& Efficient Convergence]
    Consider \( l_{\text{LLM}} (\cdot, \cdot)\) represents the standalone LLM loss, while \( l(\cdot, \cdot) \) reflects the optimized LLM-QFL model loss. By acting as a knowledge distiller, the LLM guides QFL training, improving convergence and efficiency while reducing computational overhead. Therefore, we have an inequality 
$l(\cdot, \cdot) < l_{\text{LLM}}(\cdot, \cdot)$
which implies that the QFL-trained quantum model achieves a lower loss than an LLM used in isolation. 
\end{remark}
Given $l(\cdot, \cdot) < l_{\text{LLM}}(\cdot, \cdot)$, LLMs serve as knowledge distillers, reinforcing quantum model optimization while balancing computational efficiency and predictive accuracy. 
With it, both quantum model parameters \( \boldsymbol{\theta} \) and LLM parameters \( \phi \) are optimized, ensuring a superior balance of predictive accuracy and computational efficiency.

\begin{figure*}[!h]
    \centering
    \begin{subfigure}[b]{0.35\textwidth}
        \centering
        \includegraphics[width=\columnwidth]{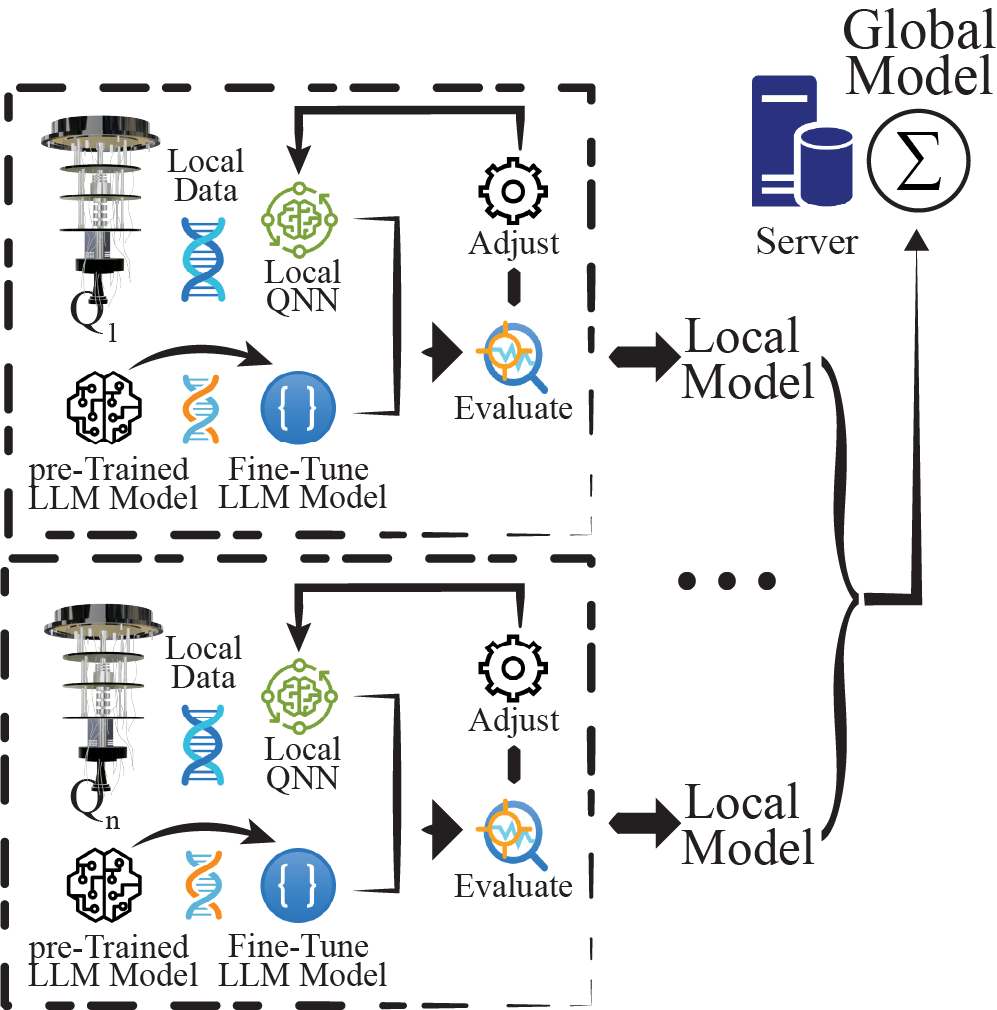}
        \caption{LLM-QFL-Pragmatic View}
        \label{fig:QFL-LLM_proposed}
    \end{subfigure}
    \begin{subfigure}[b]{0.55\textwidth}
        \centering
        \includegraphics[width=\columnwidth]{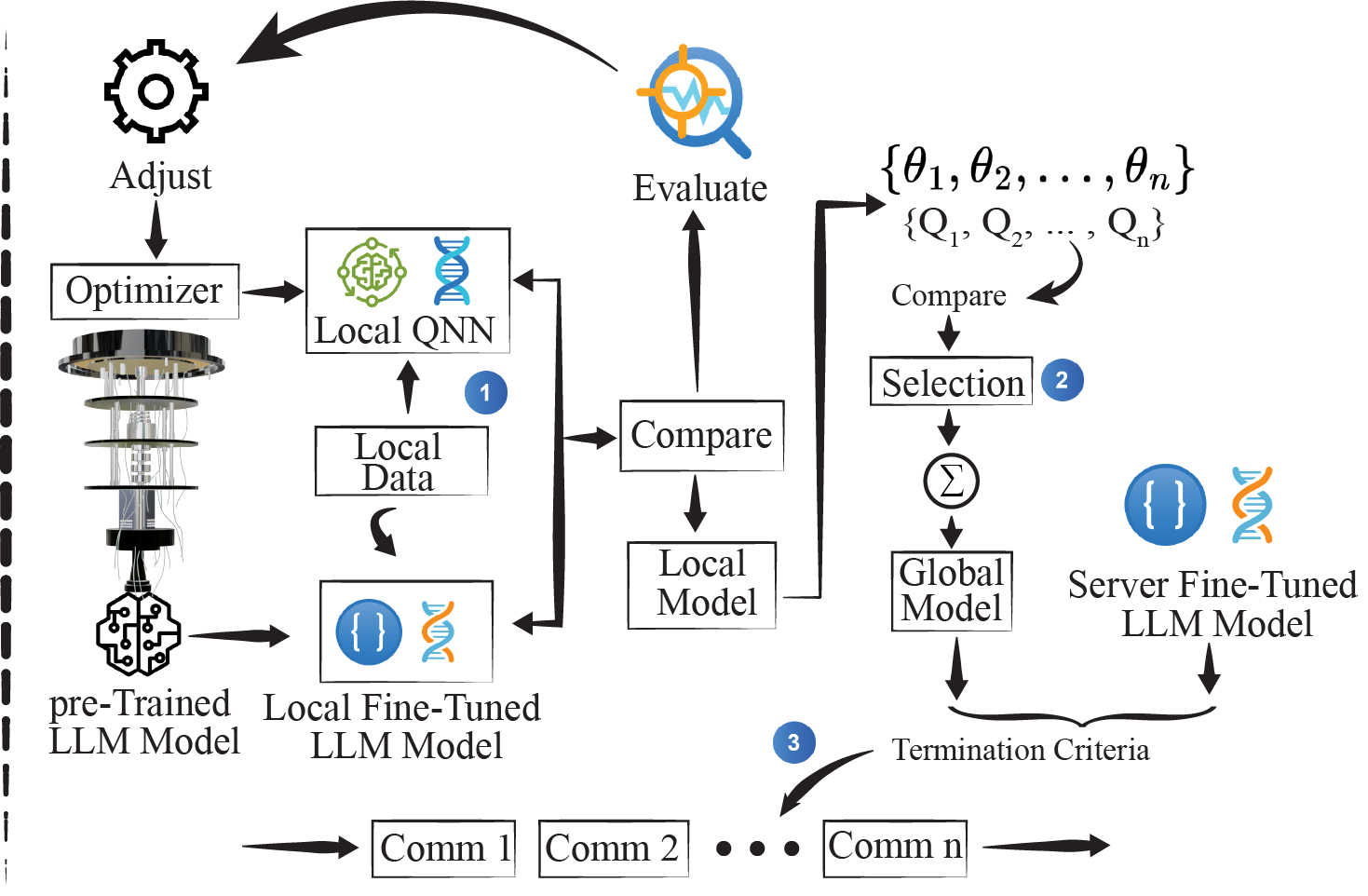}
        \caption{LLM-QFL-Detailed View}
        \label{fig:QFL-LLM_proposed_detail}
    \end{subfigure}
    \caption{Proposed LLM-QFL Framework. Each device fine-tunes its local LLM on its dataset during the initial communication round, followed by training a QCNN on local data. In subsequent rounds, local LLM fine-tuning is skipped, but knowledge distillation from the fine-tuned LLM enhances QCNN adaptation. This enables the QCNN to refine its local optimizer dynamically, leveraging comparative performance analysis to improve efficiency and model convergence.}
    \label{fig:proposed-QFL-LLM}
\end{figure*}

\section{Proposed Framework: LLM-QFL}
This section presents our proposed framework, which integrates LLMs into the QFL paradigm. The framework leverages a single LLM as a foundation model to optimize key QFL tasks, including client selection and optimizer regularization, enhancing overall performance. Additionally, it employs fine-tuning and distilling of LLMs within the QFL setting to adapt them efficiently for intended quantum-driven generative AI applications.

In the proposed LLM-QFL, with \( F_i(\theta_i^{t}) \) as the local training loss function for client \( i \), each client updates its current local model parameters at time step $t$ using fine-tuning \begin{equation}
\theta_i^{t+1} = \theta_i^{t} - \eta \nabla_{\theta_i} F_i(\theta_i^{t})
\end{equation}
where, \( \eta \) is the learning rate, \(\nabla_{\theta_i} \mathcal{F}_i(\theta_i^{t})\)
is the gradient of loss with respect to the parameters, indicating the direction of steepest increase in loss, $\theta_i^{t+1}$ is updated parameters.

Given \( \mathcal{K}(\theta_g, \theta_i) \), knowledge distillation function, typically Kullback-Leibler (KL) divergence between global and local model outputs;
each client further refines its model using distilled knowledge from a global model \( T(\theta_g) \) (acting as a teacher $T$ for distillation)
\begin{equation}
\theta_i^{t+1} = \theta_i^{t+1} + \lambda \mathcal{K}(\theta_g, \theta_i^{t+1})
\end{equation}
where \( \lambda \) controls the influence of the distillation process. 
With this process, each client tries to adjust its local model in line with the performance of fine-tuned LLM model encouraging alignment towards it.

The server aggregates updates from all participating clients and computes $\theta_g^{t+1}$ as 

\begin{equation}
\label{eqn:six}
\theta_g^{t+1} = \sum_{i=1}^{N} w_i \theta_i^{t+1} \notag
\end{equation}

Thus, with \( \mathcal{F}(\theta_i) \) as regularization term ensuring smooth convergence and preventing overfitting, the overall optimization objective for the proposed federated distillation and fine-tuning in QFL is 
\begin{equation}
\min_{\theta} \sum_{i=1}^{N} w_i \left( F_i(\theta_i) + \lambda \mathcal{K}(\theta_g, \theta_i) + \mu \mathcal{F}(\theta_i) \right)
\end{equation}
where, \( \mu \) balances the contribution of fine-tuning to the overall QFL aggregation and $w_i$ reflects each client's contribution. 
Observe that the proposed formulation ensures local adaptation, i.e. minimize $F_i(\theta_i)$ for each client, global coherence, i.e. minimize $\mathcal{K}(\theta_g, \theta_i)$ to align local models with the global model (preventing model drift), and efficient model convergence, i.e., to avoid overfitting and ensure smooth training in federated distilling and fine-tuning of LLM over QFL.
Thus, in the proposed approach, all clients and the server work together to find the parameters $\theta$ that optimize the combined objective.

\begin{algorithm}[!h]
\caption{Proposed LLM over QFL}
\label{alg:qfl_llm}
\begin{algorithmic}[1]
    \REQUIRE $N$ quantum devices, initial global model $\theta_g^0$, number of rounds $T$, optimizer $\mathcal{O}$, optimizer iterations $maxiter$.
    \ENSURE Final global model $\theta_g^T$, locally fine-tuned LLM models $\{\phi_i^T\}_{i=1}^N$
    
    \FOR{each round $t = 1$ to $T$}
        \STATE Broadcast global model $\theta^{t-1}$ to all devices
        \FOR{each device $i = 1$ to $N$ \textbf{in parallel}}
            \STATE \textbf{Step 1: LLM Fine-Tuning} 
            \IF{$t = 1$}
                \STATE Fine-tune local LLM model $\phi_i$ using local dataset $\mathcal{D}_i$
                \STATE Store evaluation metrics (e.g., loss, accuracy)
                \STATE Distill LLM using a global model
            \ENDIF
            \STATE \textbf{Step 2: QNN Training} 
            \IF{$t > 1$}
                \IF{$LLM_l < QNN_l$} 
                    \STATE Compute adjustment ratio, $r = \frac{QNN_l}{LLM_l}$
                    \STATE Update $maxiter = maxiter \cdot r$
                    \STATE Adjust optimizer $\mathcal{O}$ accordingly
                \ENDIF
            \ENDIF
            \STATE Local QNN training to minimize loss $F(\theta_i)$ 
            \STATE Obtain local model update $\theta_i^t$
        \ENDFOR
        \STATE \textbf{Global Aggregation}
         \STATE Select top-performing devices, $N' = \frac{n}{N}$ 
        \STATE Compute global model
       $ \theta_g = \frac{1}{N'} \sum_{i=1}^{N'} \theta_i^t$
        \STATE Broadcast updated global model $\theta_t$ to all devices
        \IF{Termination Criteria} 
        \STATE Stop communication rounds.
        \ENDIF
    \ENDFOR
    \STATE \textbf{Output}: $\theta_g^T$, fine-tuned LLM models $\{\phi_i^T\}_{i=1}^N$
\end{algorithmic}
\end{algorithm}

In Figure \ref{fig:proposed-QFL-LLM}, we show the details of the proposed framework.
Figure \ref{fig:QFL-LLM_proposed} shows an overview of the devices that participate and form local and global models.
Each device has its local dataset, local QNN model, locally fine-tuned model, etc. 
Based on the comparison between quantum and LLM model, we evaluate and adjust the local QNN model.
Similarly, in Figure \ref{fig:QFL-LLM_proposed_detail}, we show a detailed view of the process involved. 
In the first part of local device training, the optimizer is adapted according to the comparison between the local models; in the second step, clients are selected only with the best performance for aggregation; and finally a decision can be made for termination of the communication rounds based on comparison between the server local model and the fine-tuned model performance. The algorithm \ref{alg:qfl_llm} describes the steps involved in the overall proposed method.

\subsection {Key Components of LLM-QFL}
\textit{Distillation and Fine-Tuning LLMs in QFL}.
We design a federated fine-tuning approach for integrating pre-trained LLMs within QFL. Initially, each device fine-tunes its local LLM model using its respective dataset, which is only performed during the first communication round to establish a personalized knowledge base. To ensure adaptability to evolving data distributions, periodic fine-tuning can be reintroduced at specific intervals. Additionally, knowledge distillation is employed to enhance computational efficiency by transferring knowledge from a larger global LLM to smaller, resource-efficient local models. This technique allows local models to remain aligned with the global model while adapting to unique datasets, ensuring robust generalization with minimal computational overhead.

\textit{Adaptive Personalized Optimizer}.
The optimization process in LLM-QFL is dynamically regulated using an adaptive personalized optimizer, where fine-tuned LLMs serve as reinforcement agents for quantum convolutional neural networks (QCNNs). The main objective is to adjust the optimizer's iteration limit based on the performance of the local quantum model relative to the LLM evaluation. If the local quantum model underperforms compared to the LLM benchmark, the optimizer increases the number of iterations to enhance convergence. Various adjustment strategies can be employed, including incremental updates for gradual performance alignment, dynamic weighted adjustments that balance past performance with necessary refinements, and logarithmic scaling that prevents large changes in iteration values. These techniques ensure that poorly performing models receive additional optimization while maintaining computational efficiency and stability.

\textit{Reinforced Adaptations and Termination Criteria}.
The integration of an LLM into QFL enables autonomous decision-making in both client selection and early termination of communication rounds. Unlike conventional federated learning approaches that rely on a fixed number of iterations, our method allows the LLM to dynamically determine when training should stop based on the relative improvement of the global model. If the updates provide diminishing performance gains, training is terminated early to prevent unnecessary computational overhead. Furthermore, the LLM assists in hyperparameter tuning by adapting optimization parameters in real-time, ensuring that computational resources are efficiently allocated for improved learning efficiency. This level of autonomy enhances the overall scalability of QFL while reducing redundant communication rounds.

\textit{Client Selection}.
Client selection plays a crucial role in optimizing communication efficiency in QFL. For each communication round, only the most aligned devices, i.e., those with performance closest to the global model, are selected for participation in model updates. This strategy ensures that clients contributing the most useful updates remain actively involved while minimizing the impact of outliers that could introduce noise into the training process. Additionally, further customization can be introduced using LLM-guided selection, where local device performance is evaluated using multiple weighted comparison metrics rather than a single measure. This approach improves model stability, accelerates convergence, and ensures that only high-quality updates are incorporated into the global model, enhancing the overall effectiveness of federated quantum learning.

\subsection{Algorithmic Details}
Let $N$ represent the total number of devices in the system, each with id $i \in \{1,2,3,...,N\}$ with its own local dataset
$D_i$, local model $\theta_i$ and an associated loss function $F_i(\theta_i)$.
The server has a global model parameterized by $\theta_g$, with global objective function denoted as $F(\theta_g)$.

The fine-tuned LLM model acts as a benchmark model with parameters $\phi_{i}$ and a loss function $F_{i}(\phi)$.
It can be used to act as a regulator as well as a reinforcement agent for various tasks such as selection of optimal subset of devices, to optimize total communication rounds based on their performance alignment with LLM model, etc.

In each communication round, the objective is to 1) optimize device models $\theta_i^{t}$ locally using regulated optimizer, 2) aggregate updates from a subset of devices $S^{t}$ to update global model $\theta_g^{t+1}$ and 3) to ensure termination when global performance is optimal or reach convergence as desired.

With fine tuned LLM, the aim is to minimize the task-specific loss as, 

$$\phi_{LLM}^* = \arg \min_{\phi} F_{LLM}(\phi).$$

This provides a reference loss $\mathcal{L}_{LLM}^*$ to evaluate the performance of the local device and guide the optimization process.
In order to regulate local optimizers, 
the optimizer on device $i$ is regulated based on the difference between LLM's and the device's latest loss value as, 

$$\text{Regulated Iter} = iter * \frac{\mathcal{L}_i^{t}}{L_{LLM}^{t}}$$
where, $iter$ is the initial iterations of the optimizer and $\frac{\mathcal{L}_i^{t}}{\mathcal{L}_{LLM}^{t}}$ is the ratio between them.

If $\mathcal{L}_i^{t} \approx \mathcal{L}_{LLM}^{t}$, we conclude that the convergence is going in the right direction.
Whereas, the deviation between the two indicates a failure to improve the performance and maintain convergence.
In terms of client selection, devices are ranked based on the difference between the performance of server device and the local devices as, 
$$distance = d_i^{t} = |\mathcal{L}_i^{t} - \mathcal{L}_g^{t}|.$$
Now, top $k\%$ devices with the smallest distances are selected as, 
$$S^{t} = \{i\ |\ d_i^{t} \text{ is among the smallest } k\%\}.$$
However, further adaptive customization or use of pre-trained LLM model for this purpose can be used.
This is to make sure that the devices contributing towards better updates are participated and thus reducing the noise in global model updates caused by outliers.

For termination criteria, communication rounds can be terminated when
$$\frac{\Delta \mathcal{L}_s^{t}}{\mathcal{L}_s^{t}} < \epsilon \text{ or } t \geq T_{max}$$
where, 
$\Delta \mathcal{L}_s^{t} = \left|\mathcal{L}_s^{t} - \mathcal{L}_s^{t-1}\right|$ and 
$\epsilon > 0$ is the performance improvement threshold and $T_{max}$ is the maximum number of rounds.

For global model convergence, 
$$\theta_s^{t+1} = \theta_s^{t} - \eta. \Delta \mathcal{L}_s^{t},$$ 
where, 
$$\mathcal{L}_s^{t} = \frac{1}{|S^{t}|} \sum_{i \in S^{t} \mathcal{L}_i^{t}} F_i^{t}$$

as, 
regulated optimizers ensure that each $\mathcal{L}_i^{t}$ improves as $\mathcal{L}_i^{t+1} \leq \mathcal{L}_i^{t}\  \forall\ i, t$.
Similarly, selecting aligned devices reduces variance in $\Delta \mathcal{L}_s^{t}$, leading to faster convergence.

Finally, with termination criteria, ensuring early stopping when model achieves near-optimal performance or faces never-ending plateaus, can help in reducing unnecessary computational processes. A detailed convergence and complexity analysis of the LLM-QFL is provided in the Appendix.

\section{Experimental Evaluation of LLM-QFL}
Our simulations and experiments are conducted on real IBM quantum machines, using Qiskit and Google Colab with varying QPU/GPU configurations, including the A100 GPU (40 GB) and T4 GPU (15 GB), to evaluate the performance of proposed models under distinct computational settings. The study comprises two sets of experiments, each using a different datasets and model architecture.

\textbf{Experiment I.} In the \textit{first experiment}, we utilize the \textit{DemoHumanOrWorm} genomic dataset
from the PyTorch Datasets library, along with the \textit{Meta-LLAMA 3.2-1B}
pretrained language model. The \textit{DemoHumanOrWorm} dataset comprises 75,000 samples, each represented as a 200-nucleotide sequence labeled as either Human (0) or Worm (1), e.g., \texttt{('TCAACTGACTTCCGAG..GTCTTACTCTG', 0)}. To process genomic data for quantum machine learning, the nucleotide sequences are encoded using a standard nucleotide mapping as
\text{nucleotide\_map} $\leftarrow$ \{A: 0, C: 1, G: 2, T: 3\}.

Additionally, dimensionality reduction is applied to transform the 200-sequence features into a 4-qubit representation for quantum processing. A VQC is employed, utilizing the \textit{COBYLA optimizer}. The quantum circuit consists of a \textit{ZZFeatureMap} for data encoding and a \textit{RealAmplitudes ansatz circuit} with tunable parameters. For parameter-efficient tuning, \textit{Low-Rank Adaptation (LoRA)} is applied with the following configuration: rank (r) = 8, LoRA alpha = 16, LoRA dropout = 0.05, and bias set to ``none."

\textbf{Experiment II.} The second experiment utilizes the \textit{TweetEval-Sentiment} dataset
which contains 45,615 training samples, 12,284 test samples, and 2,000 validation samples. Each instance consists of a textual input and a sentiment label (negative, neutral, or positive), for example:

\textit{\text{label}: 2, \text{text}: ``QT @user In the original draft of the 7th book, Remus Lupin survived the Battle of Hogwarts. \#HappyBirthdayRemusLupin"}.

For this task, we employ QCNN, implemented via the \textit{Qiskit} framework. The QCNN architecture comprises feature map encoding, convolutional layers, and pooling layers. The optimization process is initialized with \textit{COBYLA} (max iterations = 10), utilizing 4-qubit encoding across 3–10 quantum devices with a training subset of 1,000 samples. Both \textit{LoRA} and quantized \textit{qLoRA} variations are explored. The language models tested in this experiment include \textit{GPT-2}
and \textit{DeepSeek-LLM-7B-Base}.

For performance benchmarking, we compare our proposed approach with state-of-the-art quantum federated learning techniques, particularly quantum versions of \textit{FedAvg}. The study includes multiple variants such as \textit{LLM-QFL-LoRA}, \textit{LLM-QFL-qLoRA}, and regulation adjustment approaches within \textit{LLM-QFL}. Each variant is analyzed in its respective section to evaluate its impact on model convergence, accuracy, and computational efficiency.

\subsection{Impact of Regulation on Optimizer}
\begin{figure}[!h]
    \centering
    \begin{subfigure}[b]{0.4\columnwidth}
        \centering
       \includegraphics[width=\linewidth]{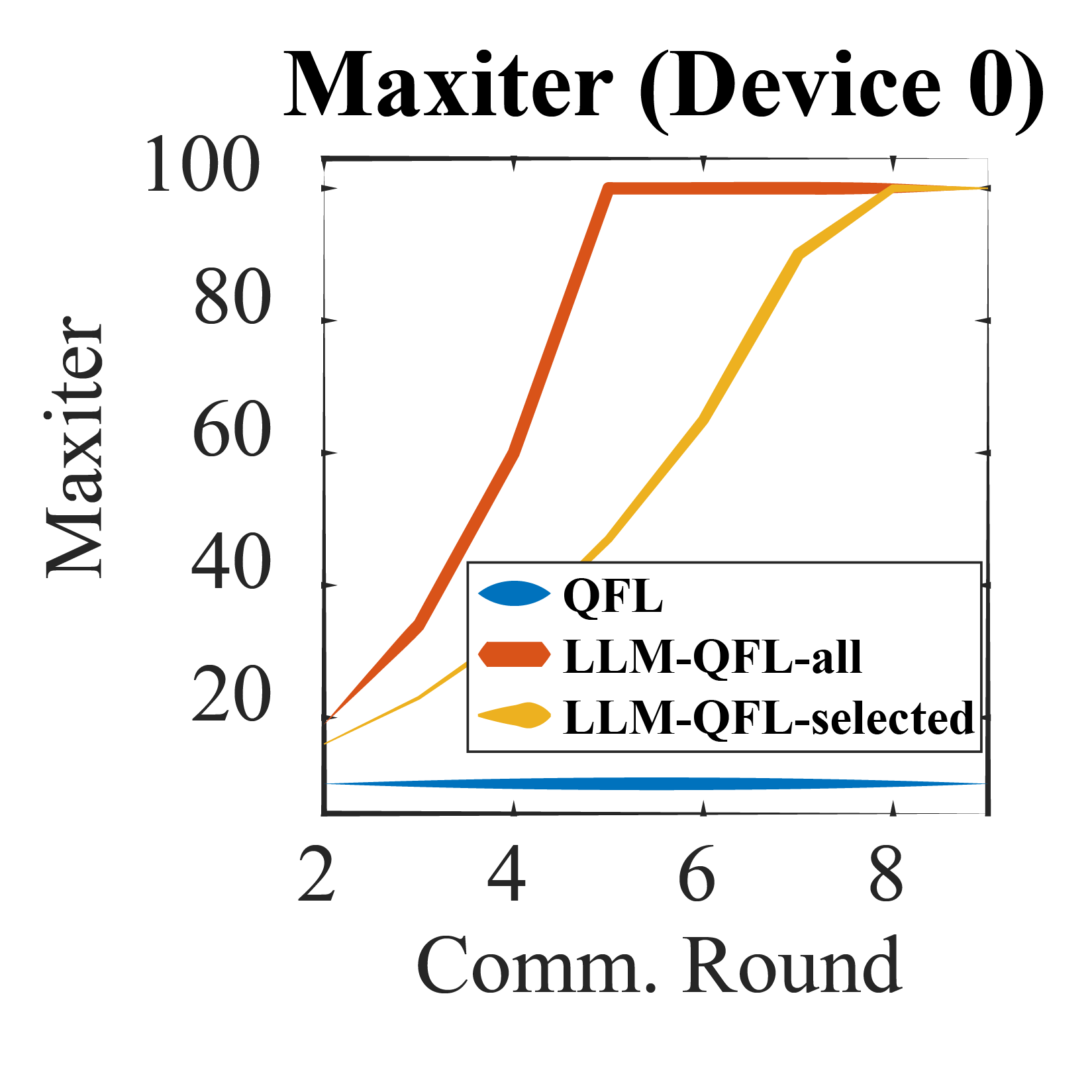}
    \caption{Maxiter}
    \label{fig:maxiter_impact_all_10d}
    \end{subfigure}
    \begin{subfigure}[b]{0.4\columnwidth}
        \centering
       \includegraphics[width=\linewidth]{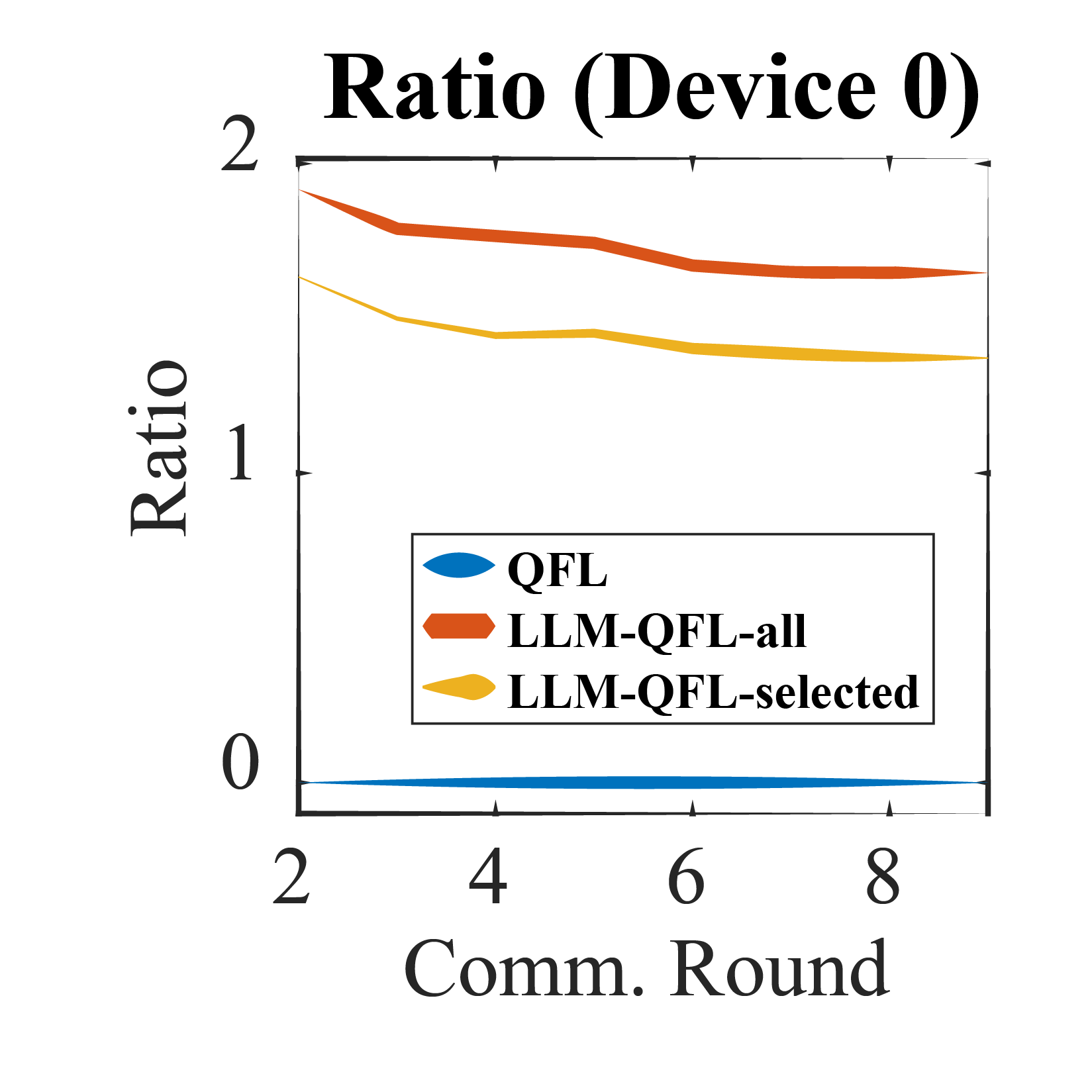}
\caption{Ratio}
\label{fig:ratio_10d}
\end{subfigure}
    \caption{Device 0 observations;
    Decreasing ratio indicates convergence and less gap between performance of LLM model and Quantum Model.}
    \label{fig:impact_on_ratio_maxiter}
\end{figure}

In Figure \ref{fig:impact_on_ratio_maxiter}, 
we can observe the impact on the optimizer and ratio value of methods \textit{QFL},\textit{ LLM-QFL-all} and \textit{LLM-QFL-selected.}
QFL is a default approach without \textit{LLM} integration, whereas \textit{LLM-QFL-all} and \textit{LLM-QFL-selected} are different in sense that ``all" refers to all devices included during FedAvg whereas only selected devices in \textit{LLM-QFL-selected} method.
The results in Figure \ref{fig:maxiter_impact_all_10d} show that the maximum value for the optimizer in QFL remains constant
whereas with other methods, there is a change in the optimizer after the second communication round. 
Adopting a regulated optimizer can serve as an efficient strategy for QFL, in contrast to using a static optimizer across all communication rounds without any adaptive elements. Maintaining a static optimizer may result in inefficiency, generating unnecessary iterations when they are not needed.

\begin{figure}[!h]
    \centering
    \begin{subfigure}[b]{0.4\columnwidth}
        \includegraphics[width=\linewidth]{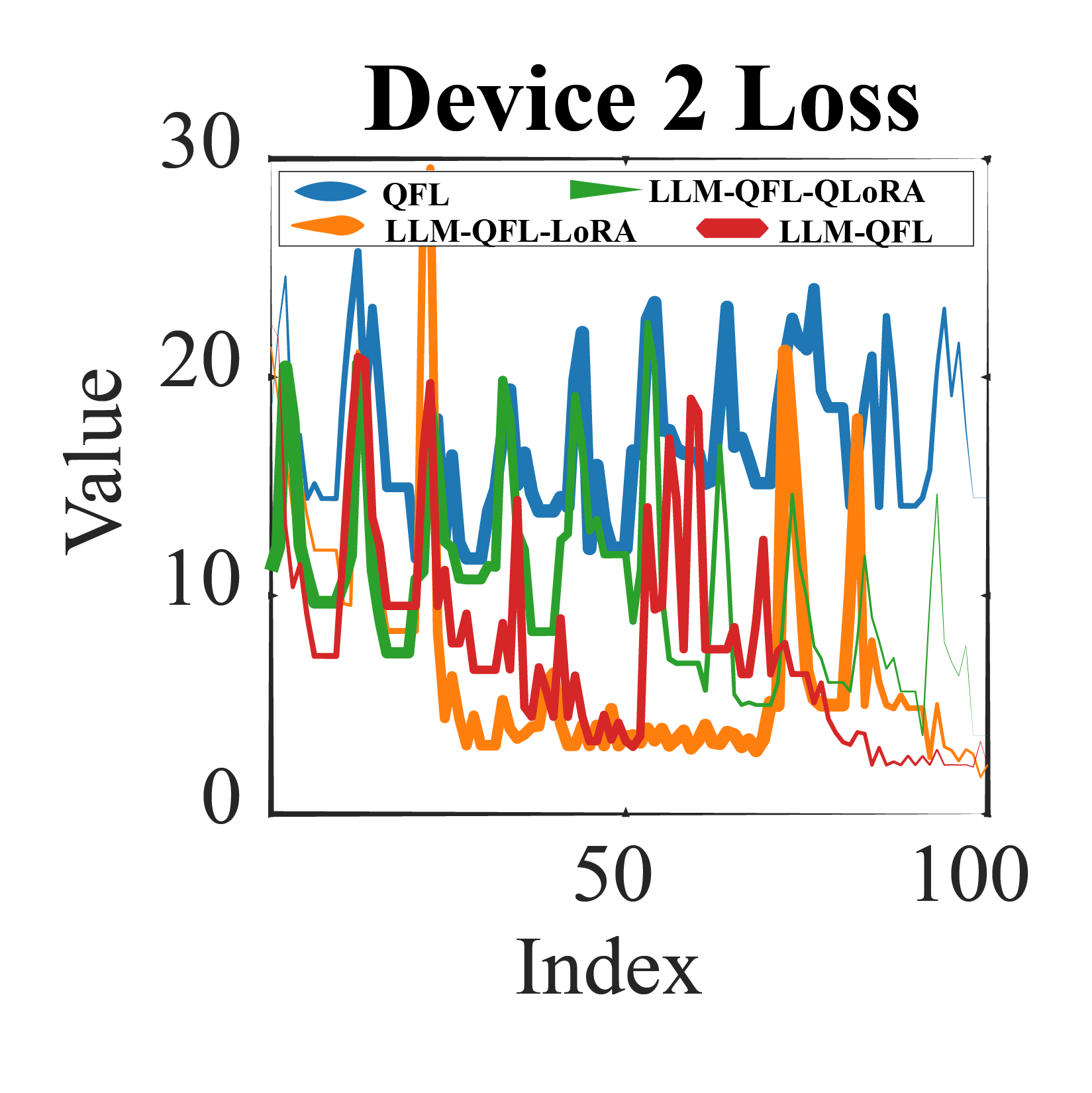}
    \caption{Device}
    \label{fig:device_objective_values}
    \end{subfigure}
    \begin{subfigure}[b]{0.4\columnwidth}
    \centering
    \includegraphics[width=\linewidth]{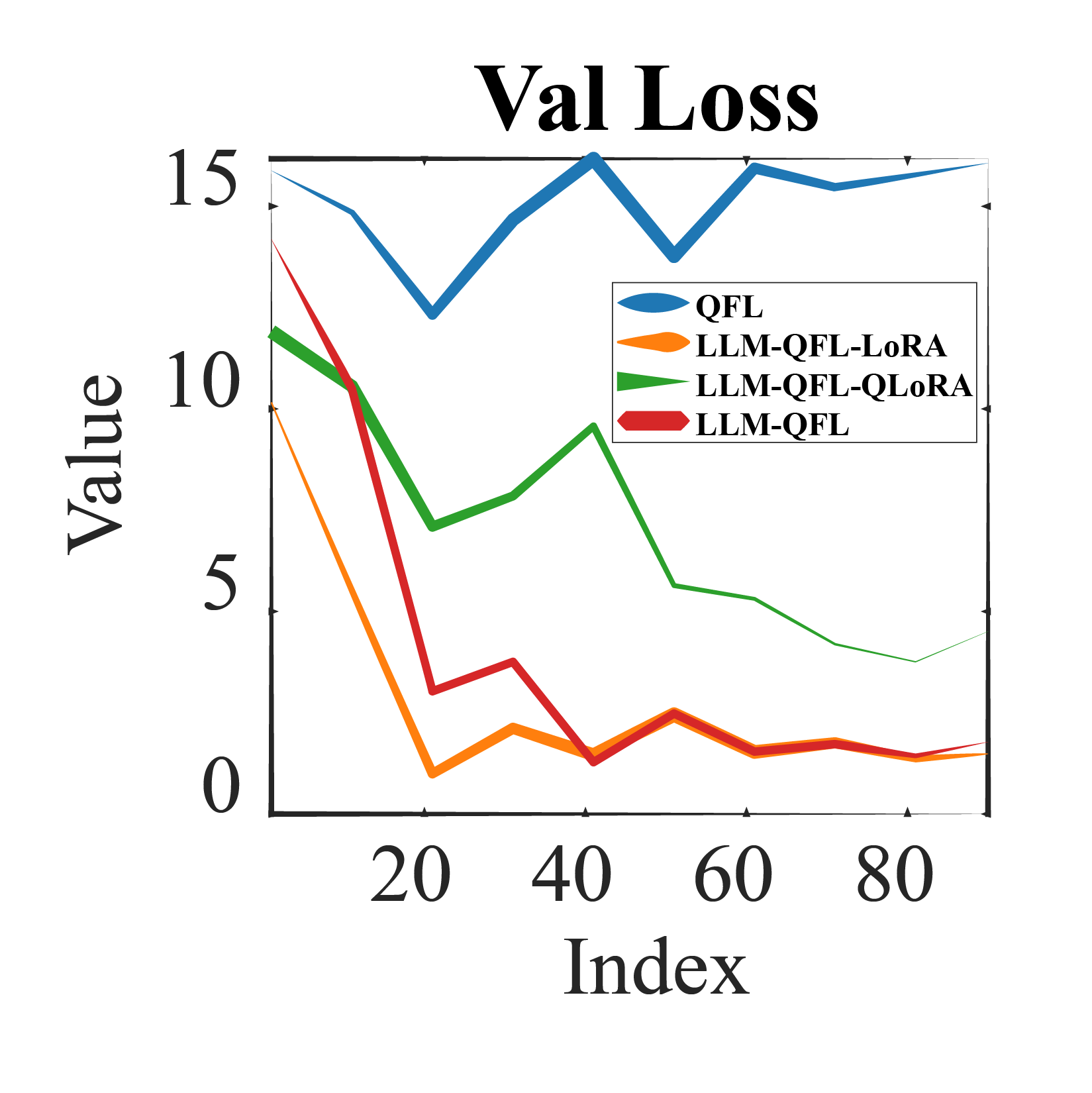}
    \caption{Server}
    \label{fig:server_objective_values}
\end{subfigure}
    \caption{a) Impact on Device performance. b) Impact on Server performance.}
    \label{fig:server_device_performance}
\end{figure}

The impact of the above pattern on the optimizer is evident in Figure \ref{fig:device_objective_values} where, we can observe the objective values for Device 2 with different methods.
The methods compared are \textit{QFL, LLM-QFL, LLM-QFL-QLoRA, LLM-QFL-LoRA}, which represents normal QFL without LLM, QFL with LLM, QFL with QLoRA LLM and QFL with LoRA LLM respectively.
For LLM-integrated QFL, due to optimizer regulation, we can clearly observe the improvement in performance.
The same result is reflected in the performance of the server device, as shown in Figure \ref{fig:server_objective_values}.

\begin{figure}[!h]
    \centering
    \begin{subfigure}[b]{0.33\columnwidth}
        \centering
        \includegraphics[width=\columnwidth]{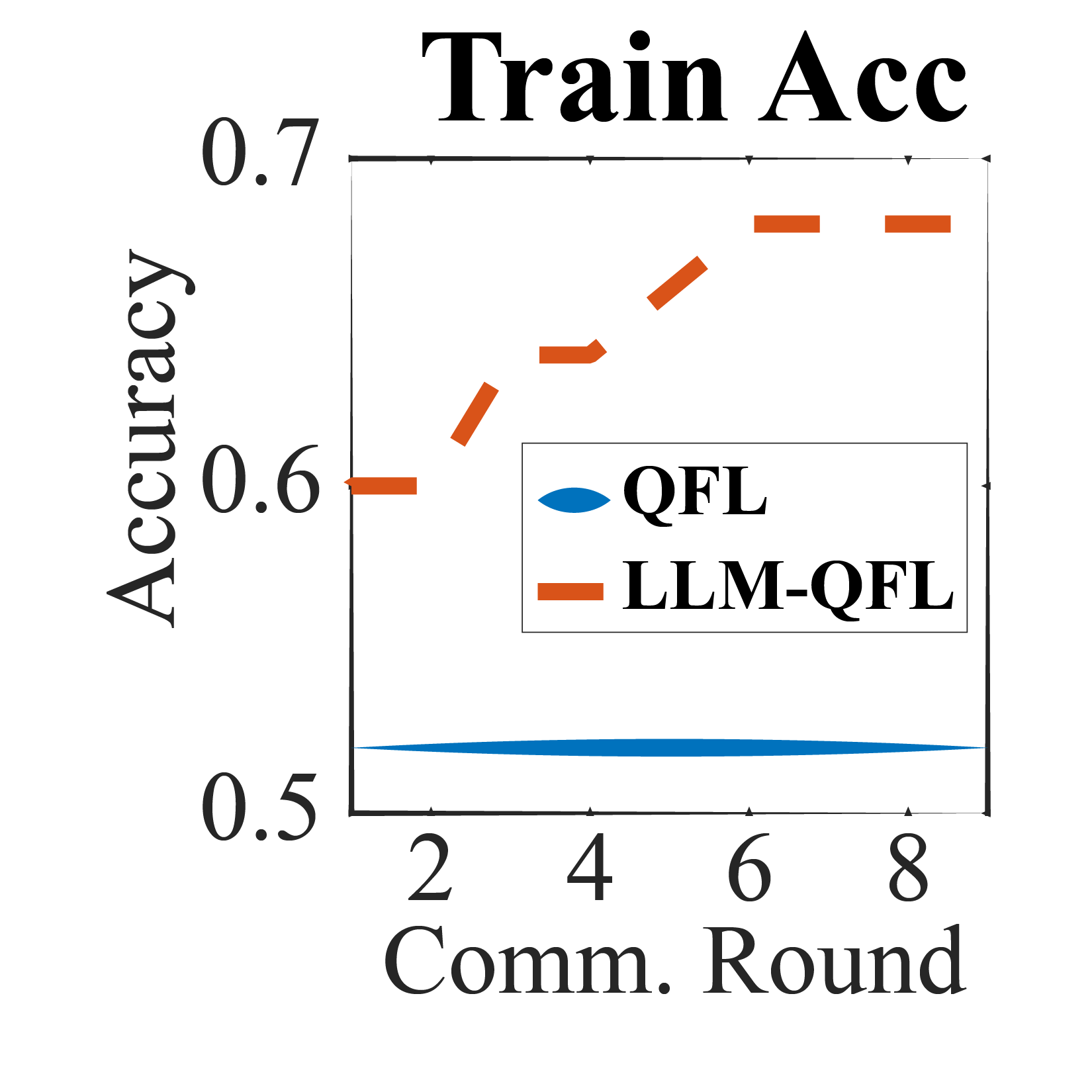}
        \caption{Train Acc}
        \label{fig:train_acc_vqc_genomic_llama_vs_qfl}
    \end{subfigure}
     \hspace{-4mm}
    \begin{subfigure}[b]{0.33\columnwidth}
        \centering
        \includegraphics[width=\columnwidth]{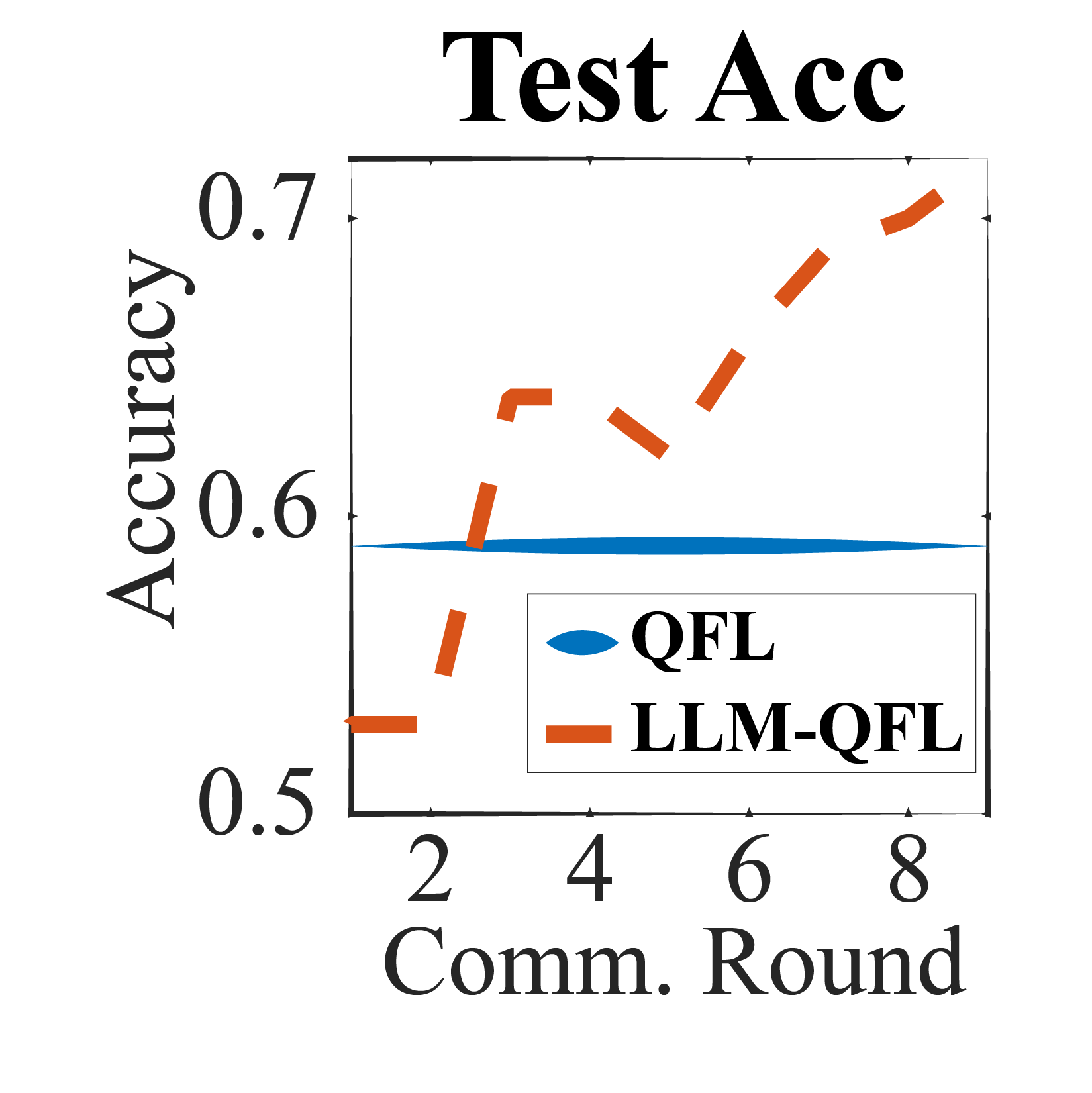}
        \caption{Test Acc}
        \label{fig:test_acc_vqc_genomic_llama_vs_qfl}
    \end{subfigure}
     \hspace{-4mm}
    \begin{subfigure}[b]{0.33\columnwidth}
        \centering
        \includegraphics[width=\columnwidth]{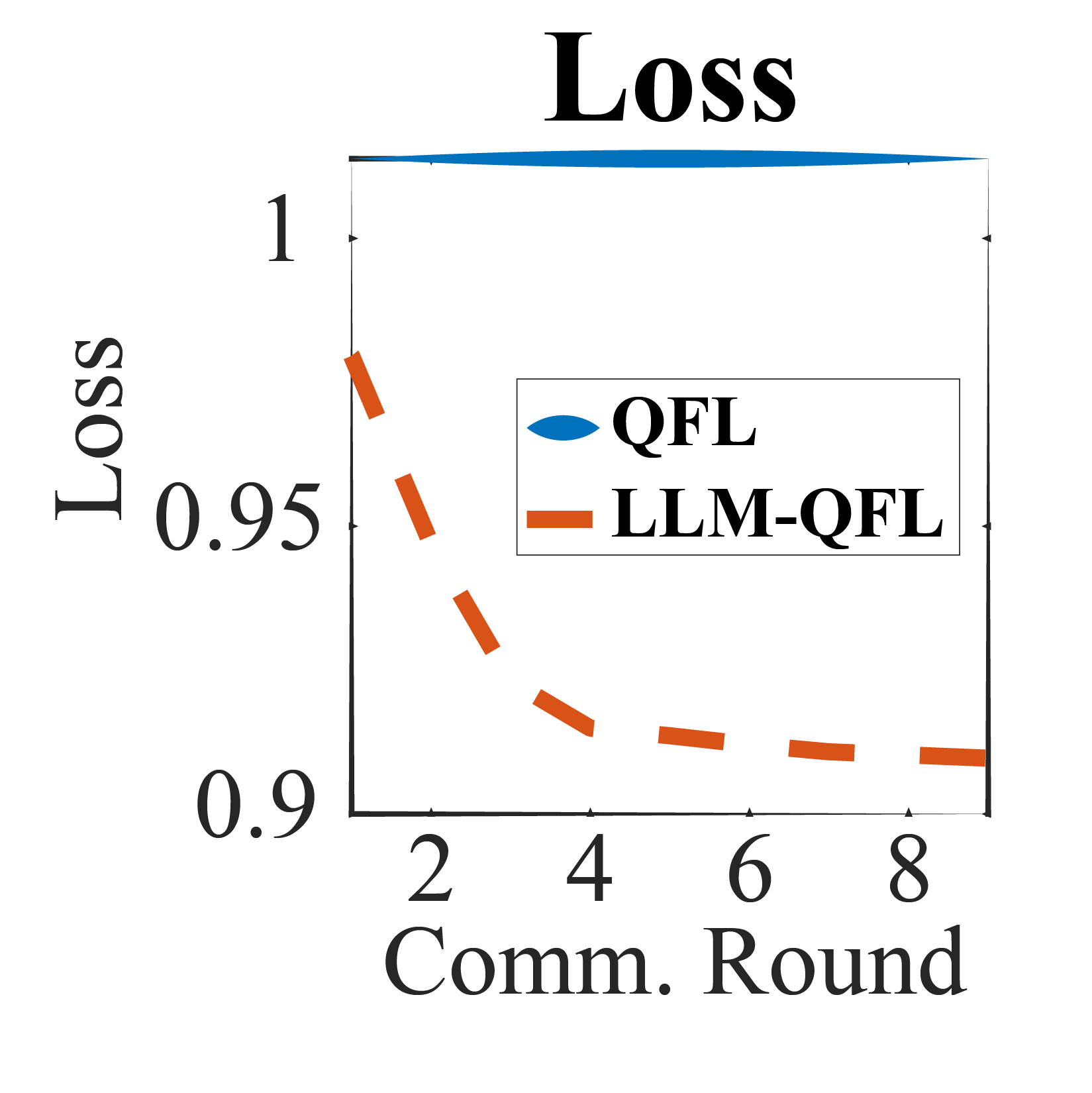}
        \caption{Loss}
        \label{fig:loss_vqc_genomic_llama_vs_qfl}
    \end{subfigure}
    \caption{Device performance: LLM vs. LLM-QFL.}
    \label{fig:device_performance_device_3devices1}
\end{figure}

In Figure \ref{fig:device_performance_device_3devices1}, the convergence on QFL using a regulated optimizer is evident in terms of
train accuracy, test accuracy and the loss of the train for device 3. Although varying optimizer parameters can cause communication rounds to last longer compared to the default QFL lacking LLM integration, convergence is ultimately achieved both sooner and more rapidly.

\subsection{Impact on Client Selection}
\begin{figure}[!h]
    \centering
    \begin{subfigure}[b]{0.4\columnwidth}
        \centering
       \includegraphics[width=\linewidth]{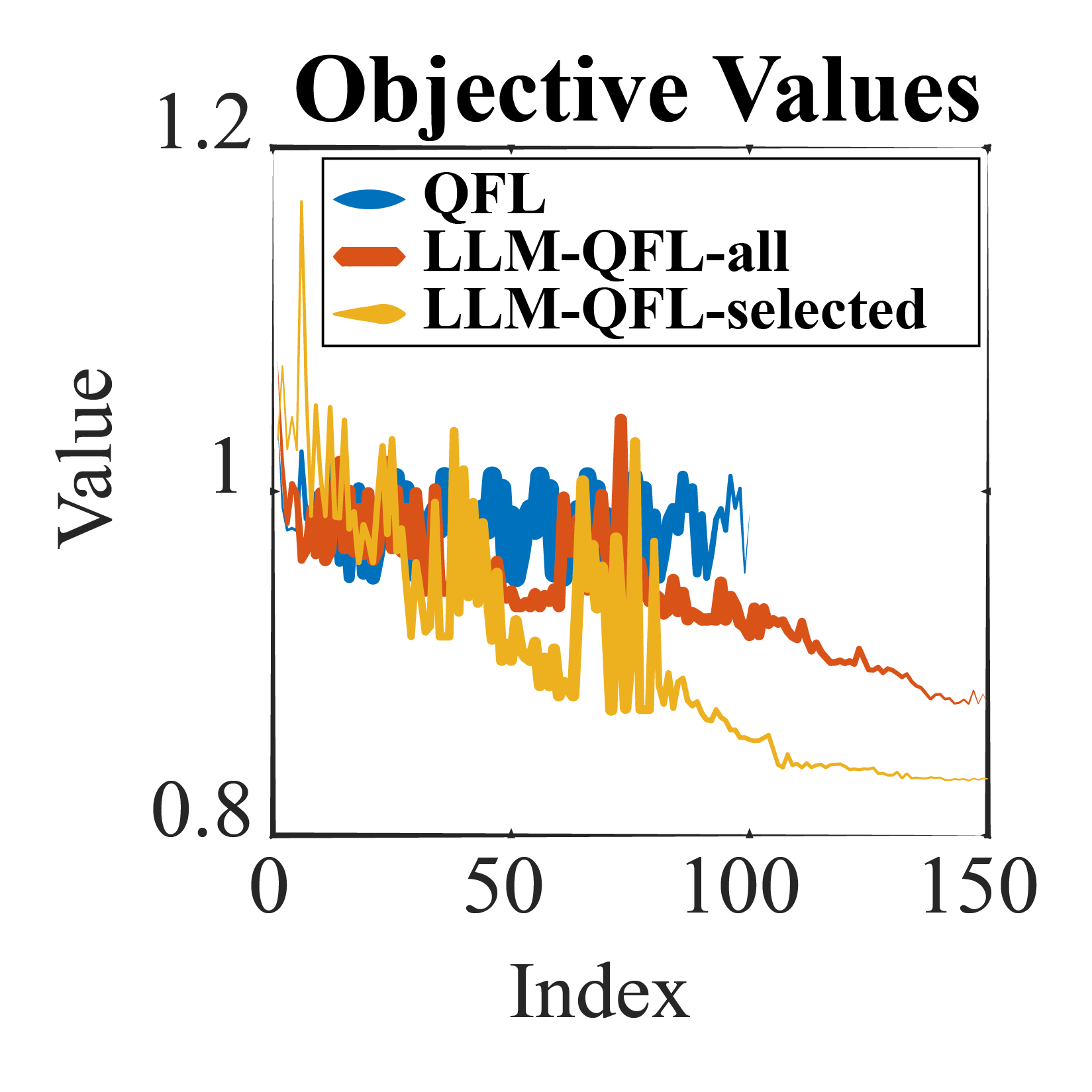}
    \caption{Till 150 Iter}
    \label{fig:client_selection_impact_genomic_device_objective_value}
    \end{subfigure}
    \hspace{-2mm}
    \begin{subfigure}[b]{0.4\columnwidth}
        \centering
       \includegraphics[width=\linewidth]{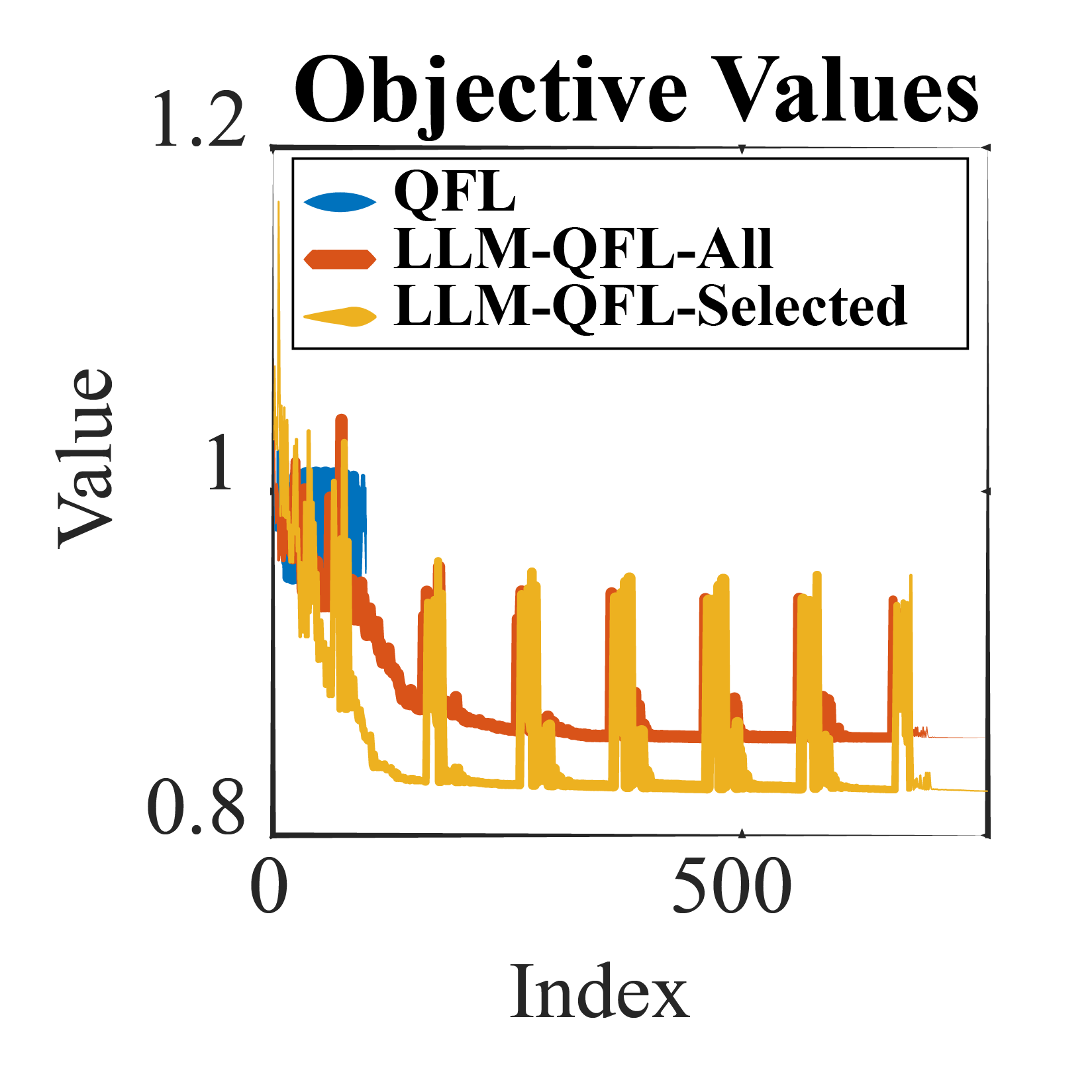}
    \caption{All Iter}
    \label{fig:client_selection_impact_genomic_device_objective_value_all}
    \end{subfigure}
 
    \caption{Device 8 performance: All vs Selected}
    \label{fig:client_selection_genomic_main_figure}
\end{figure}

Figure \ref{fig:client_selection_impact_genomic_device_objective_value} illustrates the impact of choosing a few selected clients to optimize server performance and convergence. 
The comparison is made between the standard \textit{QFL}, \textit{LLM-QFL-all}, which includes all devices, and \textit{LLM-QFL-selected}, which restricts the averaging process to just $10\%$ of the devices.

For a Device $8$, for QFL, we can see that it stops at $100$ iterations, which is a total of $10$ communication rounds for the device with $10$ local optimizer iterations.
However, with LLM assisted QFL, the convergence for both client selection and without client selection converges more within $100$ iterations, which is not spread along all $10$ communication rounds for the device
since the optimizer iterations is updated according to the need automatically.
This can reduce the need for multiple communication rounds that are not required.

While Device $8$ completes its all communication rounds, i.e. $10$ in default QFL, with LLM assisted approach, we have device performing more than $600$ iterations where MAX\_ITER is capped to $100$ per communication round. 
As shown in Figure \ref{fig:client_selection_impact_genomic_device_objective_value_all}, mostly after $200$ iterations, the convergence is almost flat with no steep convergence.
This could be used as an indication to stop the training procedure by comparing the server local performance and the server LLM model performance, or the repeated pattern from the last iterations or communication rounds, which we aim to further study extensively in our future work.

\begin{figure}[!h]
    \centering
    \begin{subfigure}[b]{0.33\columnwidth}
        \centering
       \includegraphics[width=\linewidth]{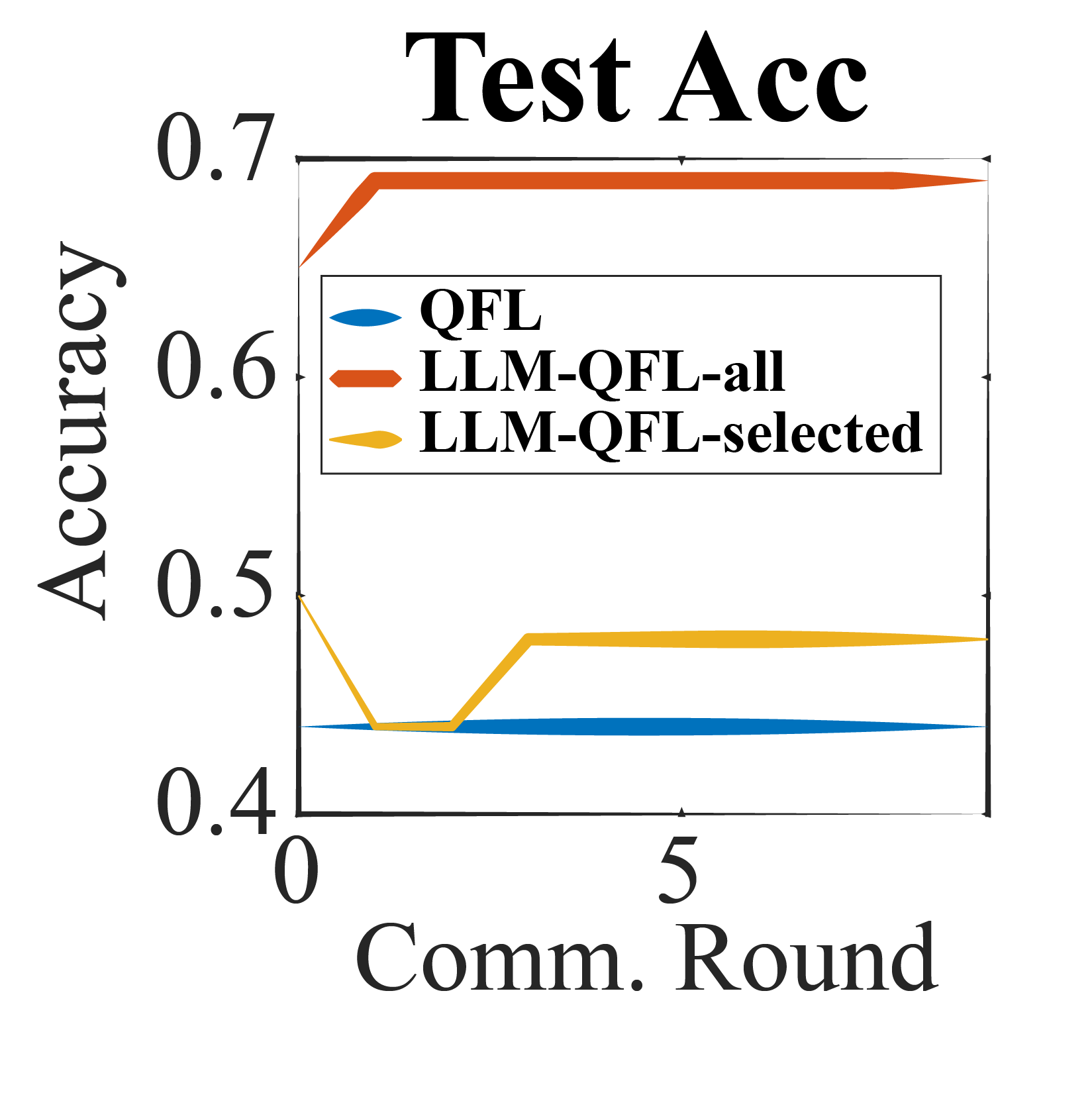}
    \caption{Test Acc}
    \label{fig:server_test_acc}
    \end{subfigure}
    \hspace{-4mm}
    \begin{subfigure}[b]{0.33\columnwidth}
        \centering
       \includegraphics[width=\linewidth]{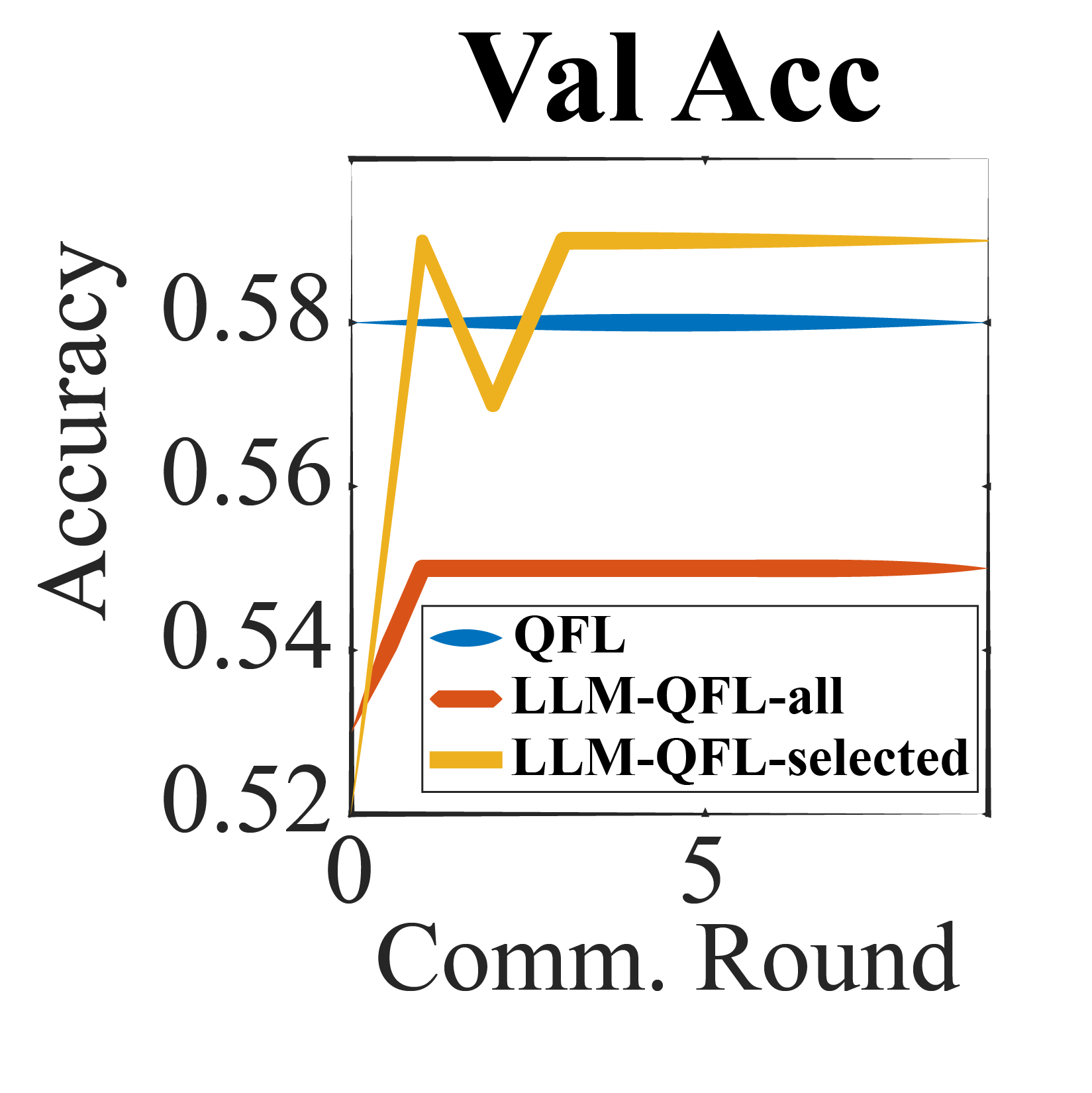}
    \caption{Val Acc}
    \label{fig:server_val_acc}
    \end{subfigure}
    \hspace{-4mm}
  \begin{subfigure}[b]{0.33\columnwidth}
        \centering
       \includegraphics[width=\linewidth]{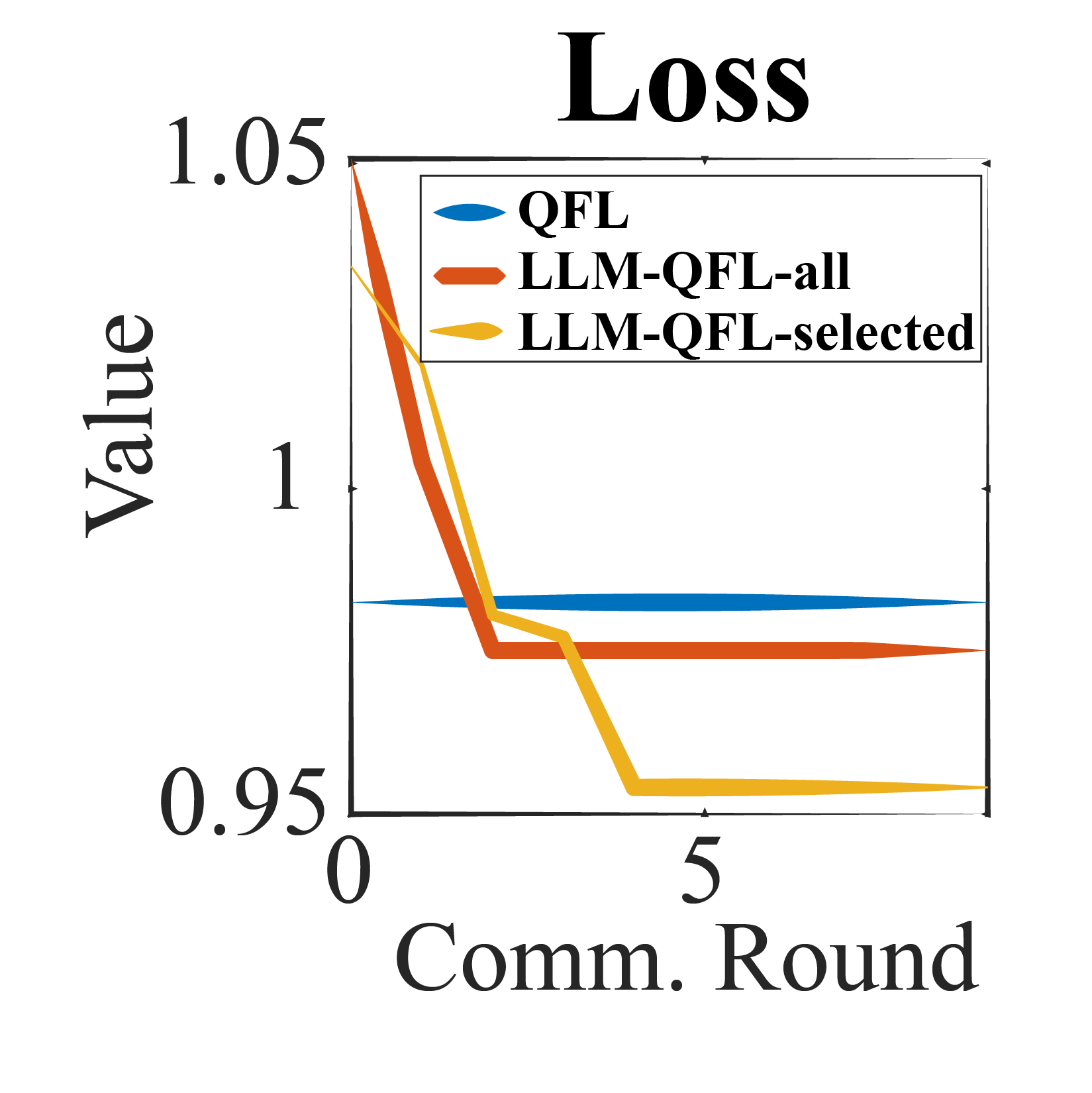}
    \caption{Val Loss}
    \label{fig:vaL_loss_server}
    \end{subfigure}
    \caption{Server performance: QFL, LLM-QFL-all vs. LLM-QFL-selected.}
    \label{fig:server_performance}
\end{figure}

Similarly, to observe the impact on server performance, in Figures \ref{fig:server_test_acc}, \ref{fig:server_val_acc} and \ref{fig:vaL_loss_server},
it is evident that the client selection method \textit{LLM-QFL-selected} performs slightly better overall compared to both the standard \textit{QFL} and \textit{LLM-QFL-all} approaches.

\subsection{Impact of Noise and Real Quantum Computer}
\begin{figure}[!h]
    \centering
    \begin{subfigure}[b]{0.32\columnwidth}
        \centering
       \includegraphics[width=\linewidth]{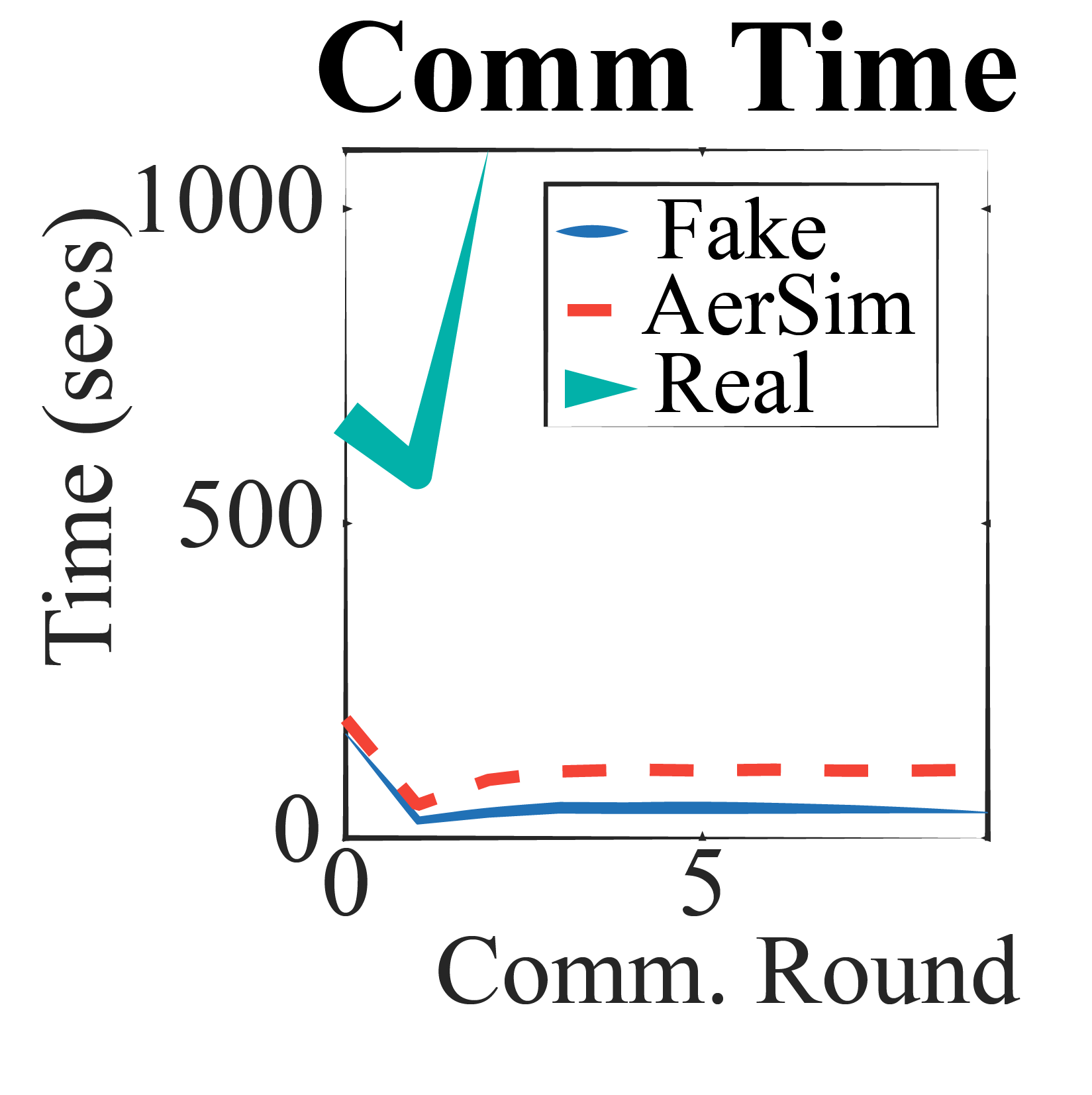}
    \caption{Comm Time}
    \label{fig:comm_time_simulators}
    \end{subfigure}
    \hspace{-2mm}
    \begin{subfigure}[b]{0.32\columnwidth}
        \centering
       \includegraphics[width=\linewidth]{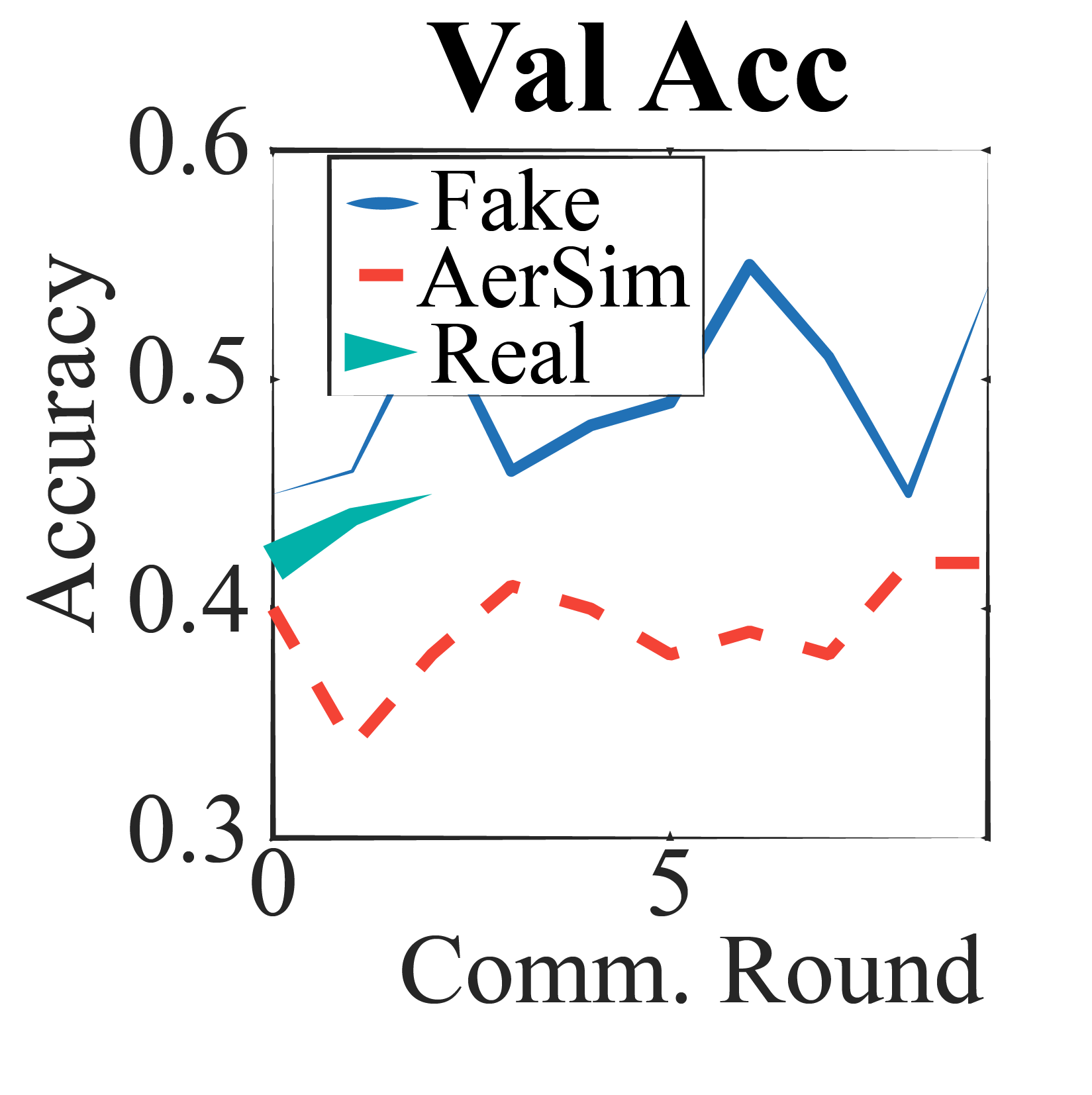}
    \caption{Val Acc}
    \label{fig:server_val_acc_simulators}
    \end{subfigure}
    \hspace{-2mm}
  \begin{subfigure}[b]{0.32\columnwidth}
        \centering
       \includegraphics[width=\linewidth]{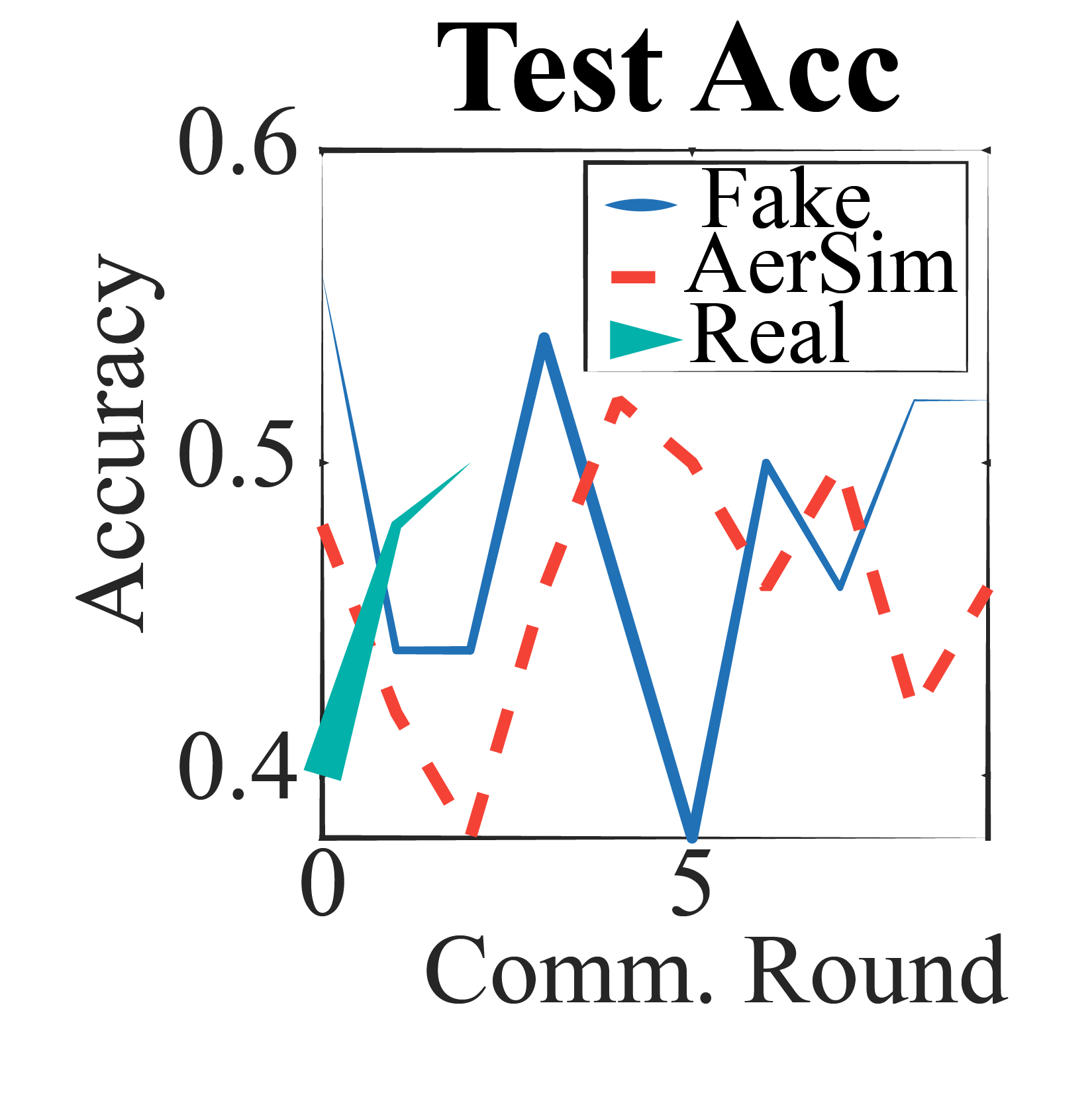}
    \caption{Test Acc}
    \label{fig:server_performance_simulators_r}
    \end{subfigure}
 
    \caption{Server Performance: Simulators vs Real Quantum Computer}
    \label{fig:server_comm_time_r}
\end{figure}

\begin{figure}[!h]
    \centering
    \begin{subfigure}[b]{0.32\columnwidth}
        \centering
       \includegraphics[width=\linewidth]{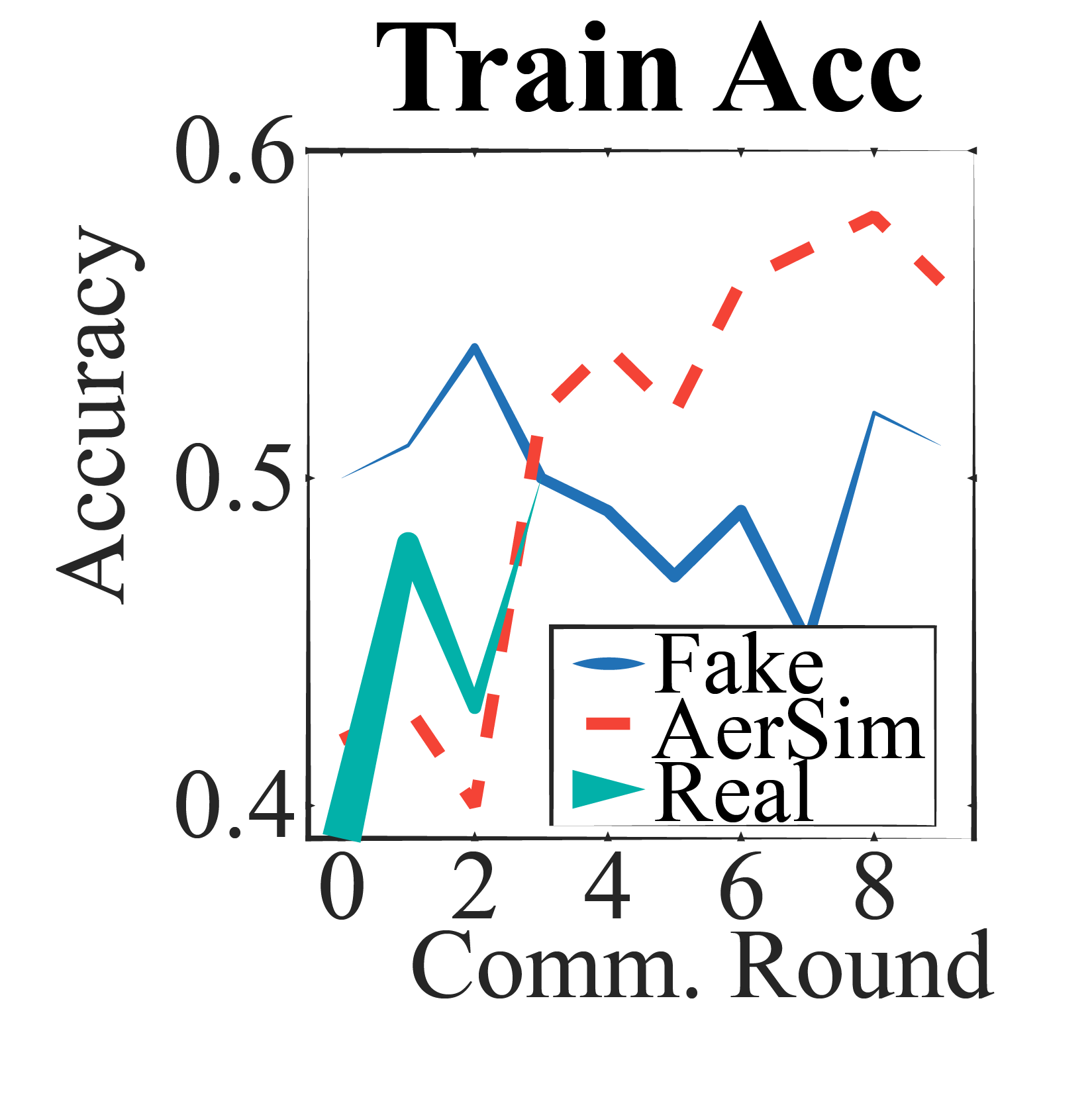}
    \caption{Train Acc}
    \label{fig:device_train_acc_simualtors_0}
    \end{subfigure}
     \hspace{-2mm}
    \begin{subfigure}[b]{0.32\columnwidth}
        \centering
       \includegraphics[width=\linewidth]{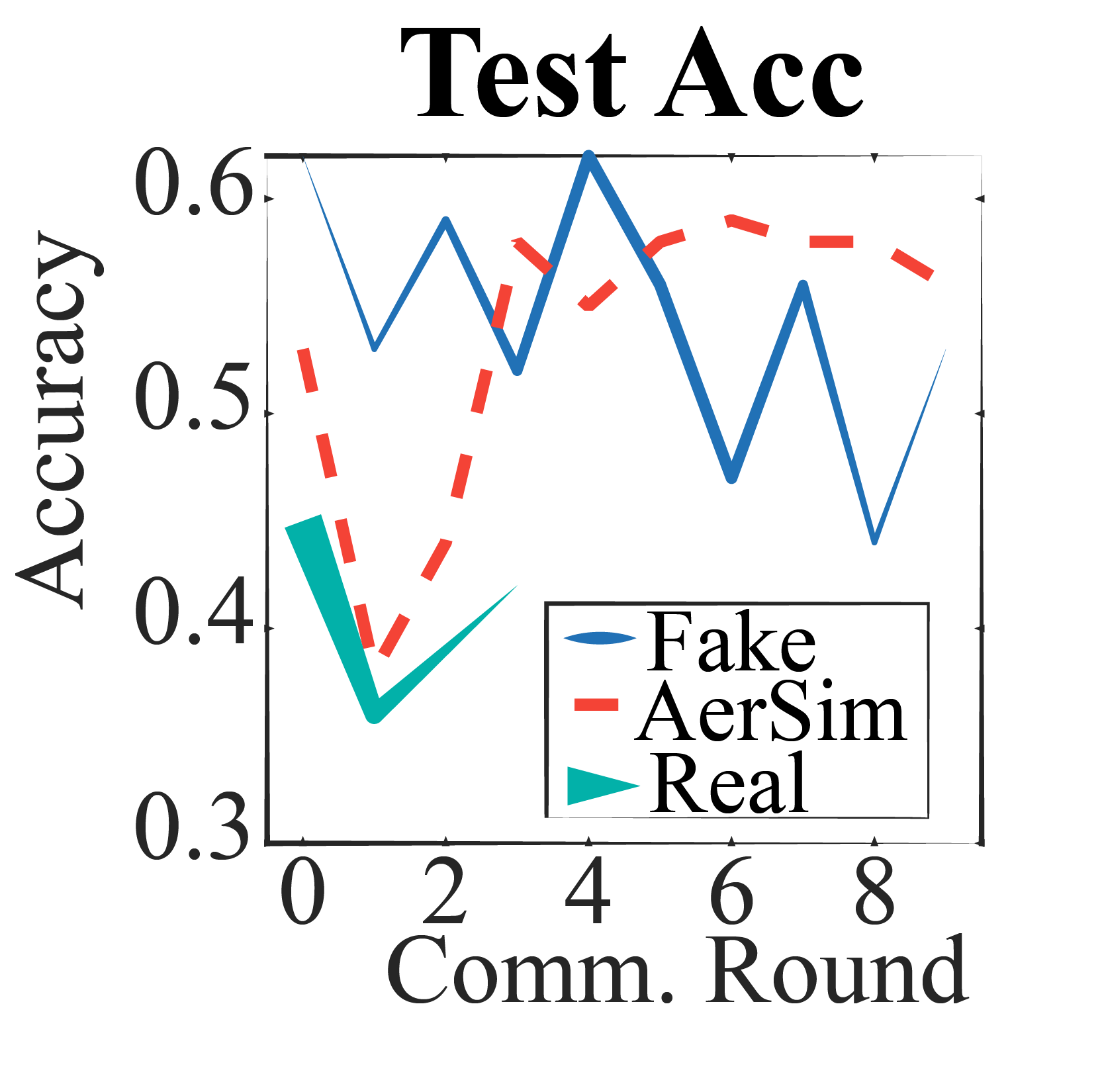}
    \caption{Test Acc}
    \label{fig:device_test_acc_simulators_0}
    \end{subfigure}
     \hspace{-2mm}
  \begin{subfigure}[b]{0.32\columnwidth}
        \centering
       \includegraphics[width=\linewidth]{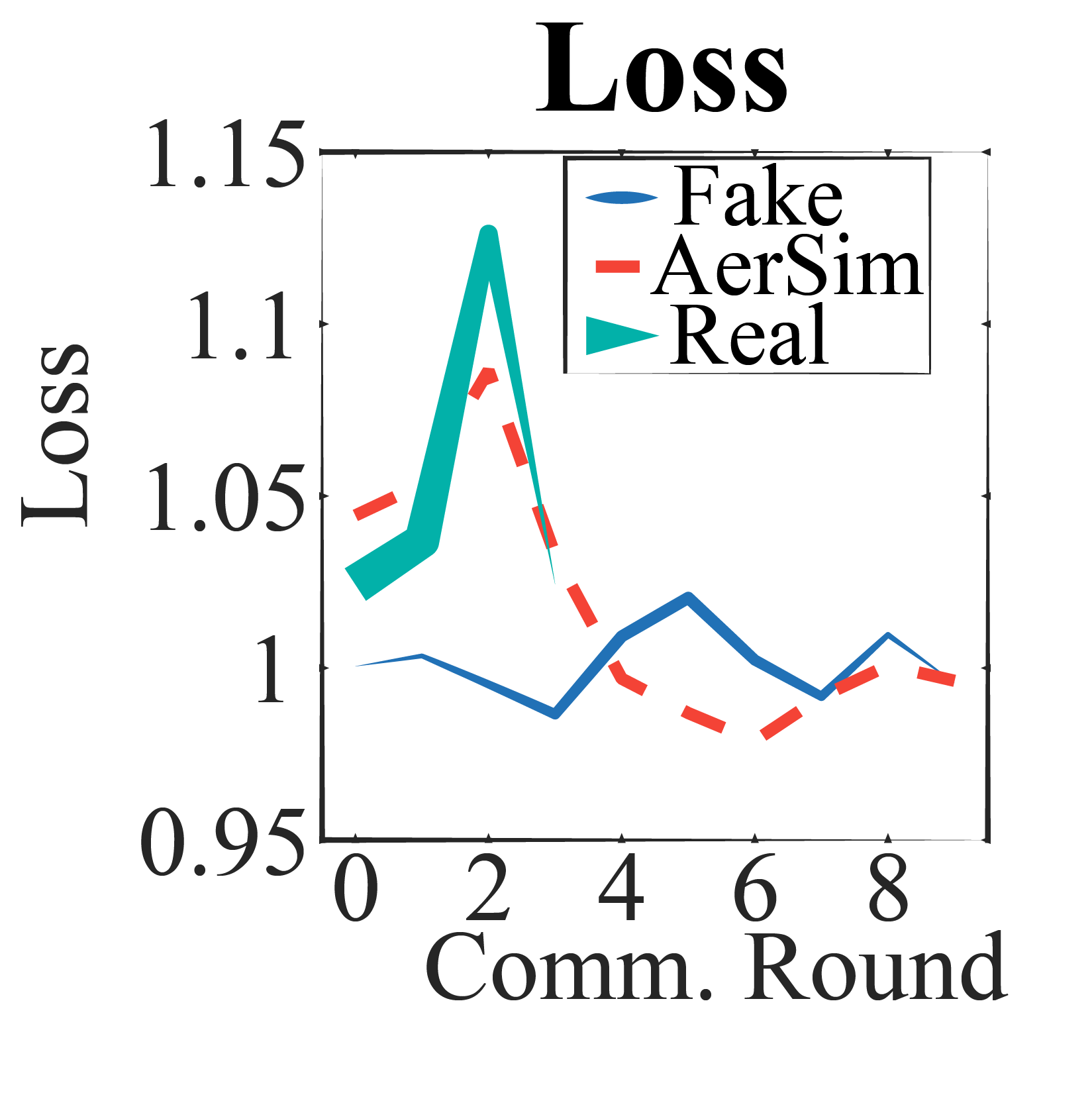}
    \caption{Train Loss}
    \label{fig:device_loss_simulators_0}
    \end{subfigure}
    \caption{Device Performance: Simulators vs Real Quantum Computer}
    \label{fig:device_performance_device_3devices_r}
\end{figure}

Figures \ref{fig:server_comm_time_r} and \ref{fig:device_performance_device_3devices_r} illustrate a performance comparison between simulation executions and those on actual quantum computers.
In these experiments, the parameters are as follows: the classifier employed is VQC, the dataset is genomic, and the simulators are \textit{FakeManila} (Fake) and a \textit{AerSimulator} with \textit{``IBM\_Brisbane"} noise (AerSim); the real quantum computer used is \textit{``IBM\_Brisbane}" (Real).
Figure \ref{fig:qfl_on_r} illustrates the procedure for executing the experiment on the real quantum computer.

In Figure \ref{fig:comm_time_simulators}, we observe that the execution time of the running experiment is slow due to various reasons such as queue time, wait times etc. than in comparison to running on simulators such as \textit{AerSimulator}, \textit{FakeManila} etc.
In terms of accuracy, the actual noise in real quantum computer is different than in simulator and thus as shown in Figure \ref{fig:server_val_acc_simulators} and \ref{fig:server_performance_simulators_r}, the performance doesn't match to that of in the simulators.

The results of performance in real quantum computer on device accuracy, test and loss in shown in Figures \ref{fig:device_train_acc_simualtors_0}, \ref{fig:device_test_acc_simulators_0} and \ref{fig:device_loss_simulators_0}.
In all instances, the performance starts lower than those in simulators.
This highlights the importance of research in the field in understanding the underlying causes of differences in the results of simulators and real quantum computers.

\begin{figure}[!h]
    \centering
    \begin{subfigure}[]{0.3\columnwidth}
        \centering
       \includegraphics[width=\linewidth]{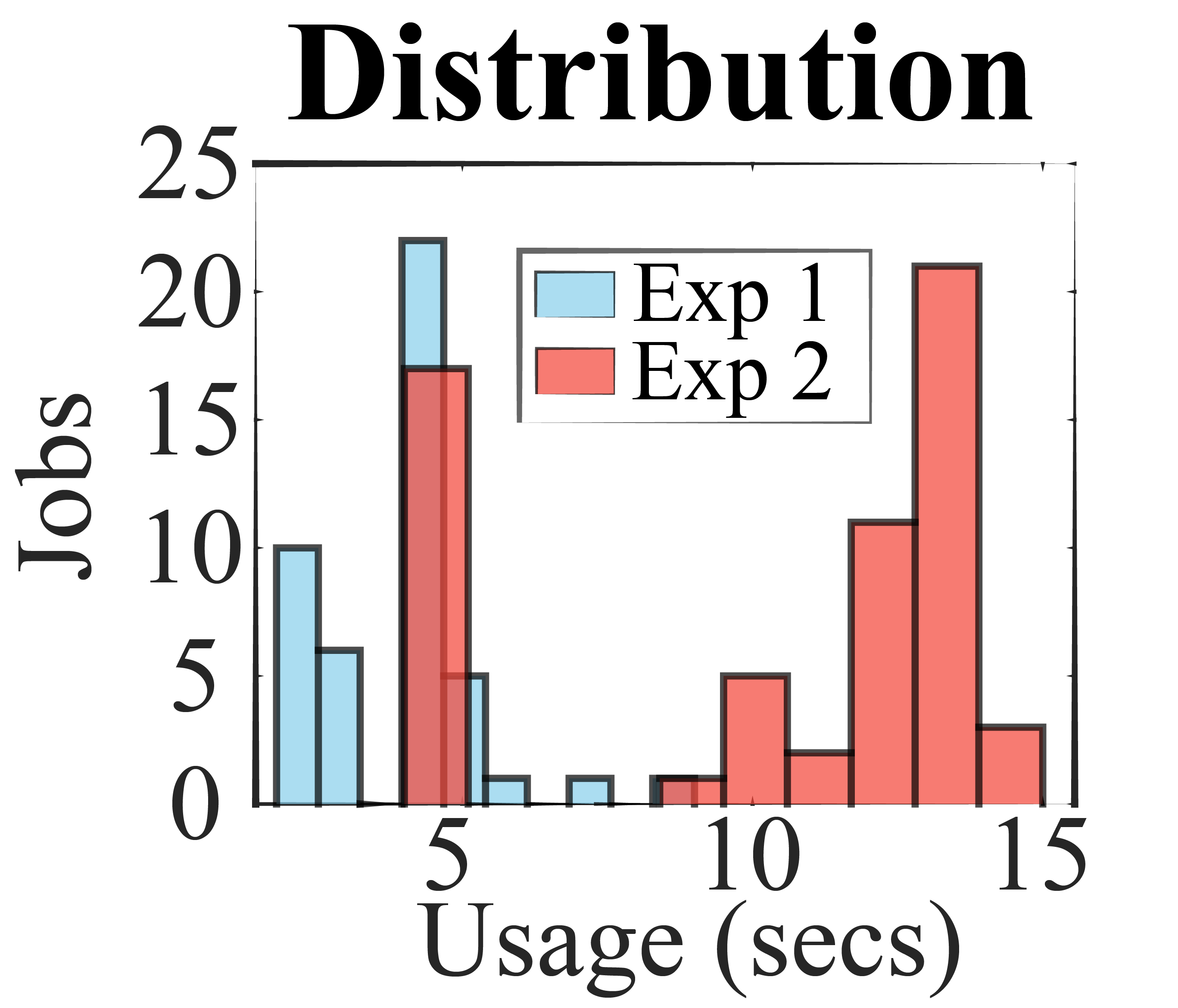}
    \caption{Usage Time}
    \label{fig:usage_times}
    \end{subfigure}
       \hspace{-3mm}
    \begin{subfigure}[]{0.55\columnwidth}
        \centering
       \includegraphics[width=\linewidth]{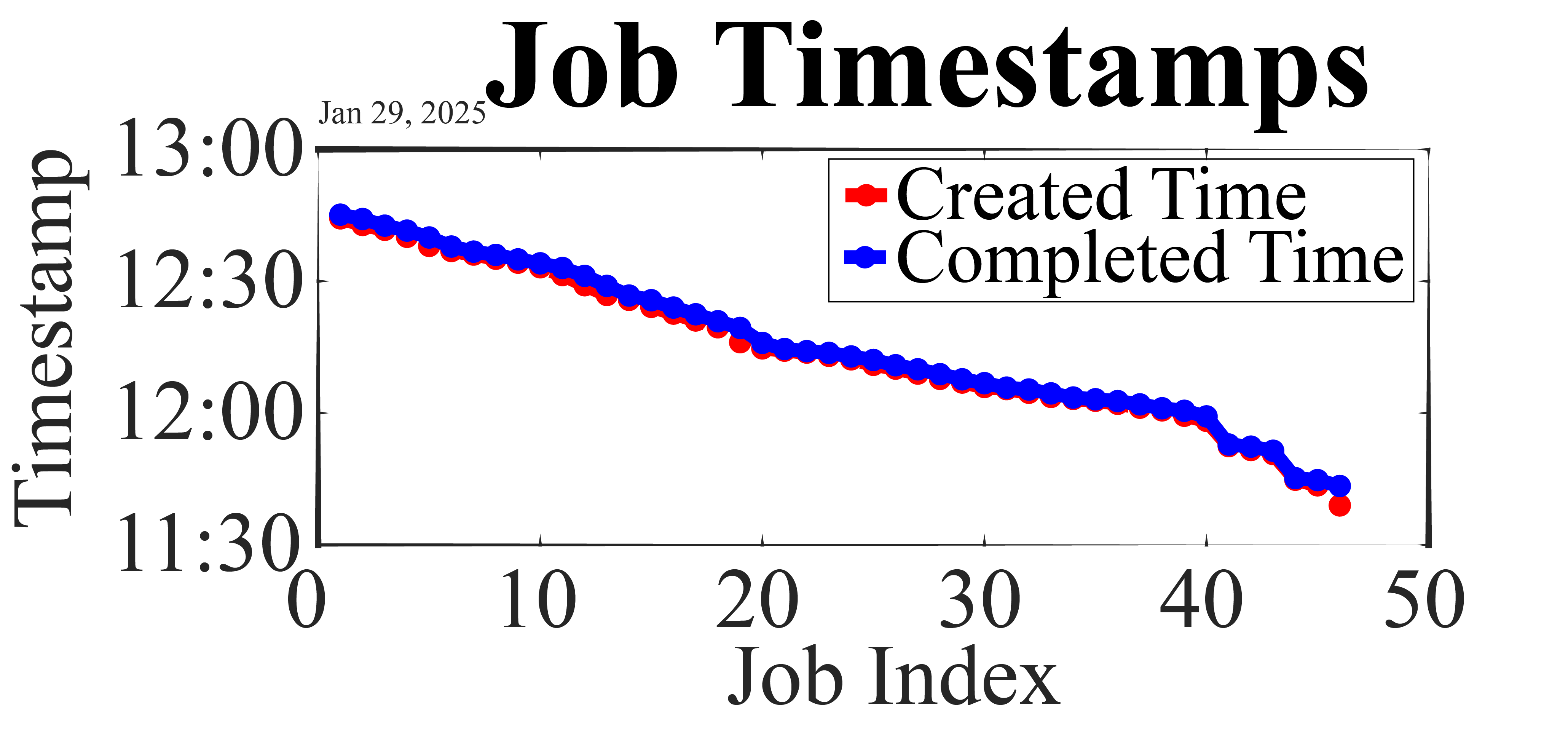}
    \caption{Timestamps}
    \label{fig:timestamps}
    \end{subfigure} 
    \caption{Usage times and Timestamps on execution on IBM quantum Computer.}
    \label{fig:execution_times}
\end{figure}

Due to limited usage limit, the results on real quantum computer are only for two (Exp 1) communication rounds.
The number of devices used is 4 including server device.
The maximum maxiter value is 5 with the initial local iteration starting at 1. 
The execution times are shown in Figure \ref{fig:execution_times} showing the usage times and cumulative timestamps for the jobs sent during the QFL execution on the real quantum computer.
Due to the low sampler shots set to 10, only 4 devices, with varying iteration from 1 to 5, more than 20 jobs required around 4 seconds to complete as shown in Figure \ref{fig:usage_times} whereas we can observe timestamps in Figure \ref{fig:timestamps} showing the job creation time and completion time showing somehow similar execution times for all jobs. Further experimental analysis can be found in the Appendix.

We further performed experimental analysis (Exp 2)
with following settings.
Dataset used: Genomic; both train and test set downloaded from the DemoHumanWorm Dataset.
Train size used is 1000 and test size is 200.
Number of devices using train sample is 3.
LLM model used is Llama-3.2-1B; adaptive approach is utilized such as $ratio * maxiter$. 
Selection method chooses the best devices aligning with server performance with various devices used for experimentation including simulators, real quantum device, FakeManila and AerSimulator.
The number of sample shots is set to 100.
The whole experimentation with 3 rounds utilized all 10 minutes free usage time limit that was provided with free plan.
From Table \ref{tab:performance_comparison_simulators}, 
we can conclude that the performance is better with AerSimulator (AerSim) in comparison to results from real IBM quantum computer (Real).
The results are obtained from two sets of experiments.

\begin{table}[h]
\centering
\caption{Performance Comparison Across Simulators}
\label{tab:performance_comparison_simulators}
\begin{threeparttable}
\resizebox{0.8\columnwidth}{!}{%
\begin{tabular}{lccccc}
\toprule
 & \multicolumn{2}{c}{Device Performance} & \multicolumn{2}{c}{Server Performance} & \\
\cmidrule(lr){2-3} \cmidrule(lr){4-5}
Simulator & Train Acc & Test Acc & Val Acc & Test Acc & Comm Time (s) \\
\midrule
\texttt{Fake} & 0.4957 & \textbf{0.5183} & 0.4680 & 0.5320 & \textbf{162.8943} \\
\texttt{AerSim} & \textbf{0.5093} & 0.5100 & \textbf{0.5350} & \textbf{0.5640} & 325.0069 \\
\texttt{Real}\tnote{*} & 0.5072 & 0.4928 & 0.4750 & 0.4700 & 1395.9734 \\
\midrule
\texttt{Fake} & \textbf{0.5087} & \textbf{0.5310} & \textbf{0.4920} & \textbf{0.4820} & \textbf{50.4605} \\
\texttt{AerSim} & 0.4637 & 0.4543 & 0.3920 & 0.4600 & 108.7150 \\
\texttt{Real}\tnote{**} & 0.4733 & 0.4525 & 0.4367 & 0.4600 & 780.1536 \\
\bottomrule
\end{tabular}
}
\end{threeparttable}
\end{table}

\section{Conclusion}
This paper proposed an LLM-QFL framework, distilling LLM with FL and leveraging LLMs to address key QFL challenges. Through theoretical and experimental analysis, we demonstrate its robustness and practicality. The fusion of QFL and LLMs marks a technological leap, opening new frontiers in AI and quantum computing. Continued research and ethical oversight are essential to unlocking their transformative potential.
\\\textit{Future Directions}.
Despite its potential, integrating QFL with LLMs faces several challenges. Quantum hardware limitations and algorithmic complexities, along with ethical concerns, demand responsible AI development. Further research is needed to explore how pre-trained LLMs can enhance quantum models, leveraging their learning and performance capabilities. Additionally, LLM-inspired quantum Transformers could be a breakthrough in QML, unlocking significant computational and performance advantages.

\section{Acknowledgement}
This research was supported by an Australian Government Research Training Program (RTP) Scholarship. 


\clearpage
\section*{Appendix} \label{sec:appendix}
\section*{A. Convergence \& Complexity of LLM-QFL}
We made the following assumptions.

\begin{assumption}[Lipschitz Smoothness]
Each local loss function $F_i(\boldsymbol{\theta}, \phi)$ is L-smooth:
\[
\|\nabla F_i(\boldsymbol{\theta}, \phi) - \nabla F_i(\boldsymbol{\theta}', \phi')\| \leq L \|(\boldsymbol{\theta}, \phi) - (\boldsymbol{\theta}', \phi')\|
\]
for all $(\boldsymbol{\theta}, \phi), (\boldsymbol{\theta}', \phi') \in \mathbb{R}^d$.
\end{assumption}

\begin{assumption}[Bounded Variance and Expected Squared Norm \cite{li_convergence_2020}]
The variance of local gradients is bounded:
\[
\mathbb{E}[\|\nabla F_i(\boldsymbol{\theta_i^t}, \phi_i) - \nabla F(\boldsymbol{\theta_i^t}, \phi_i)\|^2] \leq \sigma_i^2
\]
for all $i \in [N]$ and $(\boldsymbol{\theta}, \phi) \in \mathbb{R}^d$.

Similarly, the expected squared norm for gradients is uniformly bounded 
\[
\mathbb{E}\|\nabla F_i(\theta_i^t, \xi_i^t)\|^2 \leq G^2\ \forall\ i,t.
\]
with random sample $\xi_i^t$.
\end{assumption}

\begin{assumption}[Strong Convexity]
The global loss function $F(\boldsymbol{\theta}, \phi)$ is $\mu$-strongly convex:
\begin{eqnarray}
   F(\boldsymbol{\theta}', \phi') &\geq& F(\boldsymbol{\theta}, \phi) + \langle \nabla F(\boldsymbol{\theta}, \phi), (\boldsymbol{\theta}', \phi') \\&&- (\boldsymbol{\theta}, \phi) \rangle\nonumber + \frac{\mu}{2}\|(\boldsymbol{\theta}', \phi') - (\boldsymbol{\theta}, \phi)\|^2 
\end{eqnarray}
\end{assumption}

\begin{theorem}[Convergence of LLM-QFL]
Under Assumptions 1-3, if we set the learning rate $\eta_t = \frac{2}{\mu(t+\gamma)}$ where $\gamma = \max\{8L/\mu, E\}$, then after $T$ communication rounds, the output $(\boldsymbol{\theta}^{T}, \phi^{T})$ of Algorithm \ref{alg:qfl_llm} satisfies \cite{li_convergence_2020}:
\[
\mathbb{E}[F(\boldsymbol{\theta}^{T}, \phi^{T})] - F^* \leq \frac{2L}{\mu} \cdot \frac{\Psi}{T + \gamma}
\]
\end{theorem}
where, $\Psi = \frac{B + C}{\mu} + 2L\|(\boldsymbol{\theta}^{0}, \phi^{0}) - (\boldsymbol{\theta}^*, \phi^*)\|^2$, $B = \sum_{i=1}^M w_i^2 \sigma_i^2 + 6L\Gamma + 8(E-1)^2G^2$, and $C = \frac{4}{S^t} E^2 G^2$ with $\Gamma = F^* - \sum_{i=1}^M w_i F_i^*$, iteration per device $E_i^t = iter * \frac{\mathcal{L}_i^t}{\mathcal{L}_{LLM}}$ for $i \in S^t$ and  $S^t = \underset{\substack{S \subseteq \{1, \dots, N\}, |S| = K}}{\arg\min} \sum_{i \in S}|\mathcal{L}_i^t - \mathcal{L}_g^t|.$

$\Gamma$ quantifies degree of non-iid, $G^2$ bounds gradient norms, $\sigma^2$ bounds local gradient variance and $w_i$ is client weight factor.

\begin{proof}
Let $(\boldsymbol{\theta}^*, \phi^*)$ be the optimal solution.
For each client $i$, the local update can be written as,
\begin{eqnarray}
  (\boldsymbol{\theta}_i^{t+1}, \phi_i^{t+1}) &=& (\boldsymbol{\theta}^{t}, \phi^{t}) - \eta_t \sum_{k=0}^{K_i^{t}-1}\nonumber\\&& \nabla F_i(\boldsymbol{\theta}_i^{(t,k)}, \phi_i^{(t,k)})  
\end{eqnarray}

where, $K_i^{t}$ is the adaptive number of local steps determined by the LLM regulation.
Similarly, 
for global update we have,
\[
(\boldsymbol{\theta}^{t+1}, \phi^{t+1}) = \sum_{i \in S^{t}} w_i (\boldsymbol{\theta}_i^{t+1}, \phi_i^{t+1})
\]
where, $S^{t}$ is the selected client set.

With error decomposed, we can have
\begin{eqnarray}
   && \mathbb{E}\|(\boldsymbol{\theta}^{t+1}, \phi^{t+1}) - (\boldsymbol{\theta}^*, \phi^*)\|^2 \leq (1 - \eta_t \mu)\mathbb{E}\|\nonumber\\&&(\boldsymbol{\theta}^{t}, \phi^{t}) - (\boldsymbol{\theta}^*, \phi^*)\|^2 + \eta_t^2(B + C)
\end{eqnarray}

Applying this recursively and using the learning rate schedule gives the final bound.

\end{proof}

\begin{theorem}[Communication Complexity]
Under the same assumptions as Theorem 1, to achieve $\mathbb{E}[F(\boldsymbol{\theta}^{T}, \phi^{T})] - F^* \leq \epsilon$, the required number of communication rounds is:
\[
T = \mathcal{O}\left(\frac{L}{\mu} \log \frac{1}{\epsilon} + \frac{B + C}{\mu \epsilon}\right)
\]
\end{theorem}

\begin{proof}
From Theorem 1, we have:
\[
\frac{2L}{\mu} \cdot \frac{\Psi}{T + \gamma} \leq \epsilon
\]
Solving for $T$ gives the result.
\end{proof}

\begin{theorem}[Computation Complexity]
The total number of gradient computations across all clients to reach $\epsilon$-accuracy is:
\[
\mathcal{O}\left(\left(\frac{L}{\mu} + \frac{B + C}{\mu \epsilon}\right) \cdot \mathbb{E}[K_i^{t}]\right)
\]
where, $\mathbb{E}[K_i^{t}]$ is the average number of local steps.
\end{theorem}

\begin{remark}
We can see that the LLM regulation mechanism reduces both communication and computation complexity by:
\begin{itemize}
\item Adaptively increasing $K_i^{t}$ for clients with higher loss, accelerating their convergence
\item Selecting clients with smaller $d_i^{t}$, reducing the variance term $B$
\item Early stopping when $\frac{\Delta L_s^{t}}{L_s^{t}} < \epsilon$, avoiding unnecessary rounds
\end{itemize}
\end{remark}

\begin{corollary}[Efficiency Gains of LLM-QFL]
Let $T_{\text{QFL}}$ and $T_{\text{LLM-QFL}}$ be the number of communication rounds needed to reach $\epsilon$-accuracy for standard QFL~\cite{pokhrel2024quantum} and LLM-QFL respectively. Under the same assumptions as in Theorem 1, we have the following.

1. (Adaptive Step Size Efficiency)
\[
\frac{T_{\text{QFL}}}{T_{\text{LLM-QFL}}} \geq \frac{\mathbb{E}[K_i^{t}]}{K}
\]
where, $K$ is the fixed number of local steps in QFL and $\mathbb{E}[K_i^{(t)}]$ is the average adaptive steps in LLM-QFL.

2. (Variance Reduction)
\[
\text{Var}(\nabla F_{\text{LLM-QFL}}) \leq \left(1 - \frac{k}{N}\right)\text{Var}(\nabla F_{\text{QFL}})
\]
where, $k/N$ is the fraction of selected clients.
\end{corollary}

\begin{proof}
1. From the convergence rate in Theorem 1, the effective progress per round scales with the number of local steps. The adaptive mechanism ensures:
\[
\mathbb{E}[K_i^{(t)}] = \mathbb{E}\left[\text{iter} \cdot \frac{L_i^{t}}{L_{LLM}^{t}}\right] \geq K
\]
whenever $L_i^{t} \geq L_{LLM}^{t}$. Thus,
\[
T_{\text{LLM-QFL}} = \mathcal{O}\left(\frac{1}{\mathbb{E}[K_i^{t}]}\right) \leq \frac{K}{\mathbb{E}[K_i^{t}]} T_{\text{QFL}}
\]

2. For client selection variance, let $S_{\text{rand}}^{t}$ be random selection, and $S_{\text{align}}^{t}$ be our alignment-based selection. The variance decomposes as,
\[
\text{Var}(\nabla F) = \mathbb{E}[\|\nabla F_i - \nabla F\|^2] = \frac{1}{N}\sum_{i=1}^N d_i^{(t)2}
\]
Our selection criterion $S^{(t)} = \{i | d_i^{t} \text{ smallest } k\%\}$ minimizes
\[
\text{Var}(\nabla F_{\text{LLM-QFL}}) = \frac{1}{k}\sum_{i \in S_{\text{align}}^{(t)}} d_i^{(t)2} \leq \frac{1}{k}\sum_{i \in S_{\text{rand}}^{(t)}} d_i^{(t)2}
\]
By Markov's inequality on the sorted $\{d_i^{t}\}$,
\[
\mathbb{E}[\text{Var}(\nabla F_{\text{LLM-QFL}})] \leq \left(1 - \frac{k}{N}\right)\text{Var}(\nabla F_{\text{QFL}})
\]
\end{proof}

\section*{B. Details of the Evaluation}
\begin{table*}[h]
\centering
\caption{Summary of Experimental Setup and HyperParameters}
\resizebox{\linewidth}{!}{
\begin{tabular}{|l|l|l|}
\hline
\textbf{Aspect} & \textbf{Experiment I} & \textbf{Experiment II} \\ \hline
\textbf{Dataset} & DemoHumanOrWorm (75,000 samples, 200-nucleotide sequences) & TweetEval-Sentiment (45,615 train, 12,284 test, 2,000 val) \\ \hline
\textbf{Data Type} & Genomic (Human: 0, Worm: 1) & Textual (Sentiment: negative, neutral, positive) \\ \hline
\textbf{Encoding} & Nucleotide mapping \{A: 0, C: 1, G: 2, T: 3\}, 4-qubit representation & 4-qubit encoding \\ \hline
\textbf{Model} & Meta-LLAMA 3.2-1B (pretrained) & GPT-2, DeepSeek-LLM-7B-Base \\ \hline
\textbf{Quantum Architecture} & VQC (ZZFeatureMap + RealAmplitudes ansatz) & QCNN (feature map, conv. layers, pooling layers) \\ \hline
\textbf{Optimizer} & COBYLA & COBYLA (max iter = 10) \\ \hline
\textbf{Tuning Method} & LoRA (r = 8, alpha = 16, dropout = 0.05, bias = none) & LoRA, qLoRA \\ \hline
\textbf{Quantum Devices} & IBM quantum machines  & 3–10 quantum devices \\ \hline
\textbf{Training Subset} & Full dataset & 1,000 samples \\ \hline
\textbf{Computational Setup} & Qiskit, Google Colab (A100 GPU: 40 GB, T4 GPU: 15 GB) & Qiskit, Google Colab (A100 GPU: 40 GB, T4 GPU: 15 GB) \\ \hline
\textbf{Benchmarking} & N/A & QFL variants (LLM-QFL-LoRA, LLM-QFL-qLoRA, FedAvg) \\ \hline
\end{tabular}
}
\end{table*}

\subsection{Datasets}
\subsubsection{TweetEval, Sentiment}
\begin{figure}
    \centering
    \includegraphics[width=0.7\linewidth]{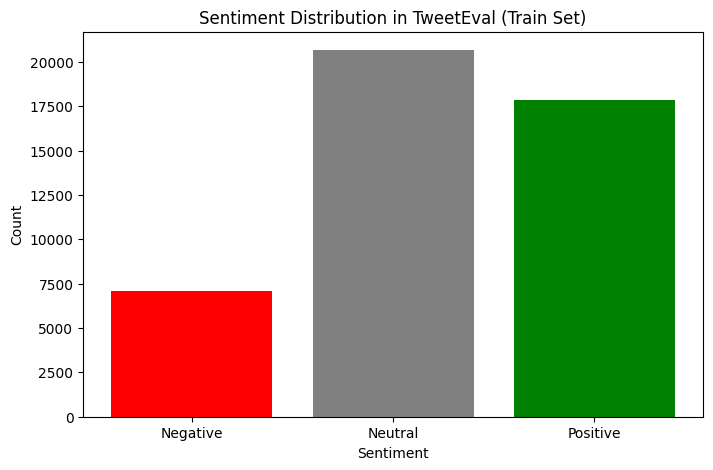}
    \caption{TweetEval-Sentiment Train Set}
    \label{fig:tweet_eval_sentiment}
\end{figure}

TweetEval consists of heterogeneous tasks in Twitter which is used as multiclass dataset with tasks like \textit{irony, hate, offensive,  stance, emoji, emotion} and \textit{sentiment}.
For the \textit{ sentence} task, there are three classes \textit{negative, positive,} and \textit{neutral} with varying sample as shown in Figure \ref{fig:tweet_eval_sentiment}.

\begin{lstlisting}[language=Terminal]
    DatasetDict({
    train: Dataset({
        features: ['text', 'label'],
        num_rows: 45615
    })
    test: Dataset({
        features: ['text', 'label'],
        num_rows: 12284
    })
    validation: Dataset({
        features: ['text', 'label'],
        num_rows: 2000
    })
})
\end{lstlisting}

\subsubsection{Genomic-DemoHumanOrWorm Dataset}
\begin{figure}
    \centering
    \includegraphics[width=0.7\linewidth]{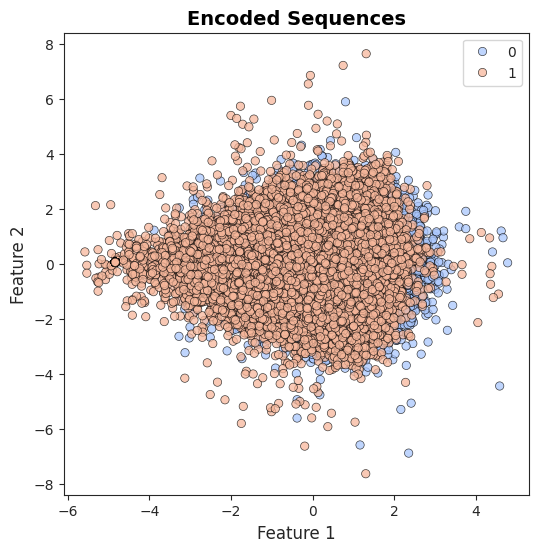}
    \caption{Encoded and PCA applied Genomic Dataset}
    \label{fig:encoded_genomic_data}
\end{figure}

DemoHumanOrWorm is a benchmark genomic sequence dataset for classification purposes which consists of 75,000 train samples and 25,000 test set samples with data instance as, 
\begin{lstlisting}[language=Terminal]
('TCAACTGACTTCCGAGGGAATAAGTGTTTCGCCAT
CTCGAACTGTATACTCTGCTATCAA
GACCGTTACTGTAAGTGTTGTTTTCAAACTGC
AAGTTTAAAACTGAAAATATTTTCAGCA
CAATCTCCTAAACCGTGGCAACTGGGGAAAT
TGTTCAGAAAATTGGCCT
GAAGGGTATAACACGTTCCTGTCTTACTCTG', 0)
\end{lstlisting}
and labels as 0 or 1.
The encoded and PCA (n=4) applied dataset is shown in Figure \ref{fig:encoded_genomic_data}.

\subsubsection{Data Preparation of LLM fine-tuning and VQC training}
Both the architecture of LLM and VQC is different.
LLM is based on the Transformer architecture, while VQC is based on the Quantum Neural Network architecture. 
We cannot feed the same data format to both models.
Thus, there are some necessary steps involved in preprocessing the dataset for both the LLM model and the VQC model.

\begin{enumerate}
    \item First, we load and shuffle dataset. Based on the required data sample size, we deduct small samples.
    \item Then, we convert dataset into dataframe with columns ``dset" for train or ``test" set, cat to store labels (Human or Worm), ``seq" for nucleotide sequences.
    \item For LLaMA model, first we convert dataframe to HuggingFace Dataset format. For tokenization, we first define k-mer tokenization (substrings of length k=6), load LLaMA tokenizer (meta-llama/Llama-3.2-1B), ensure padding, load pretrained model for sequence classification (2 labels), resize token embeddings to include new pad token and apply tokenization to the dataset using map(). Finally, we create DatasetDict (`train': , `test':) for LLM.
    \item For VQC model, one-hot encoding is applied to nucleotide sequences as A=[1,0,0,0], C=[0,1,0,0], G=[0,0,1,0], T=[0,0,0,1] and are converted to one-hot vectors. Also, the Principal component analysis is applied for dimensionality reduction with PCA (n\_components=4) which reduces 200 nucleotides feature sequence into 4 features representations.
\end{enumerate}

\begin{figure*}[!h]
    \centering
    \includegraphics[width=0.8\textwidth]{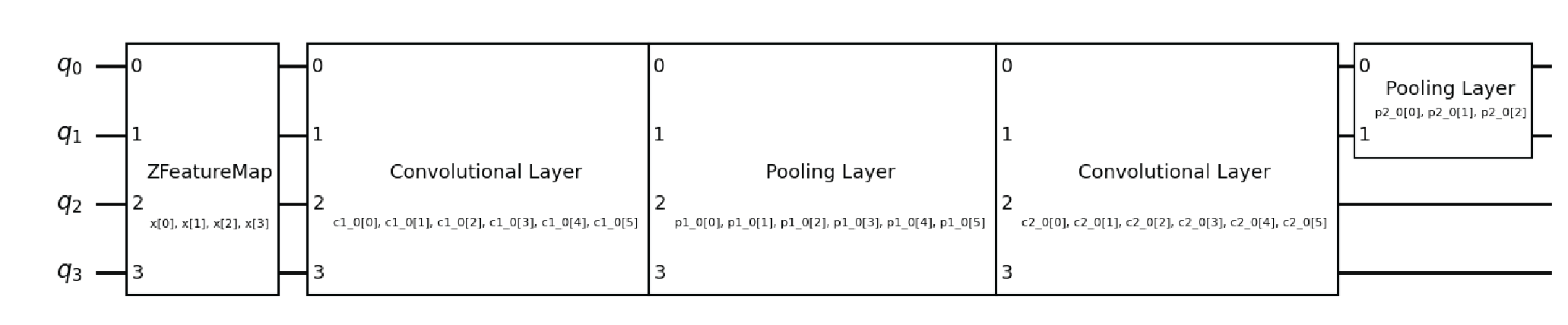}
    \caption{Quantum Convolutional Neural Network (QCNN); Example of QCNN with 4 qubits. At each layer, dimensionality is reduced from four to two qubits by disregarding the first two qubits.}
    \label{fig:QCNN}
\end{figure*}

\begin{figure*}[!h]
    \centering
    \includegraphics[width=0.8\textwidth]{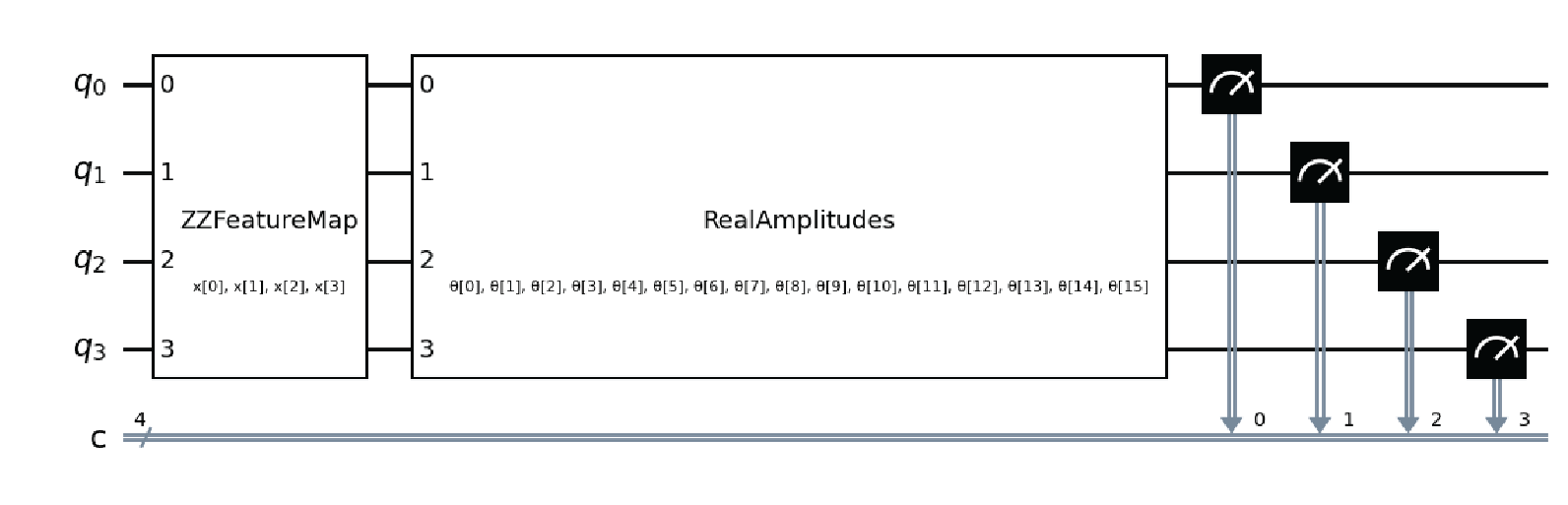}
    \caption{Variational Quantum Classifier with 4 Qubits; Consists of feature map using ZZFeatureMap, ansatz using RealAmplitudes and finally the measurement for classification.}
    \label{fig:VQC}
\end{figure*}

\subsection{IBM QPU execution}
\begin{figure}
    \centering
    \includegraphics[width=0.9\linewidth]{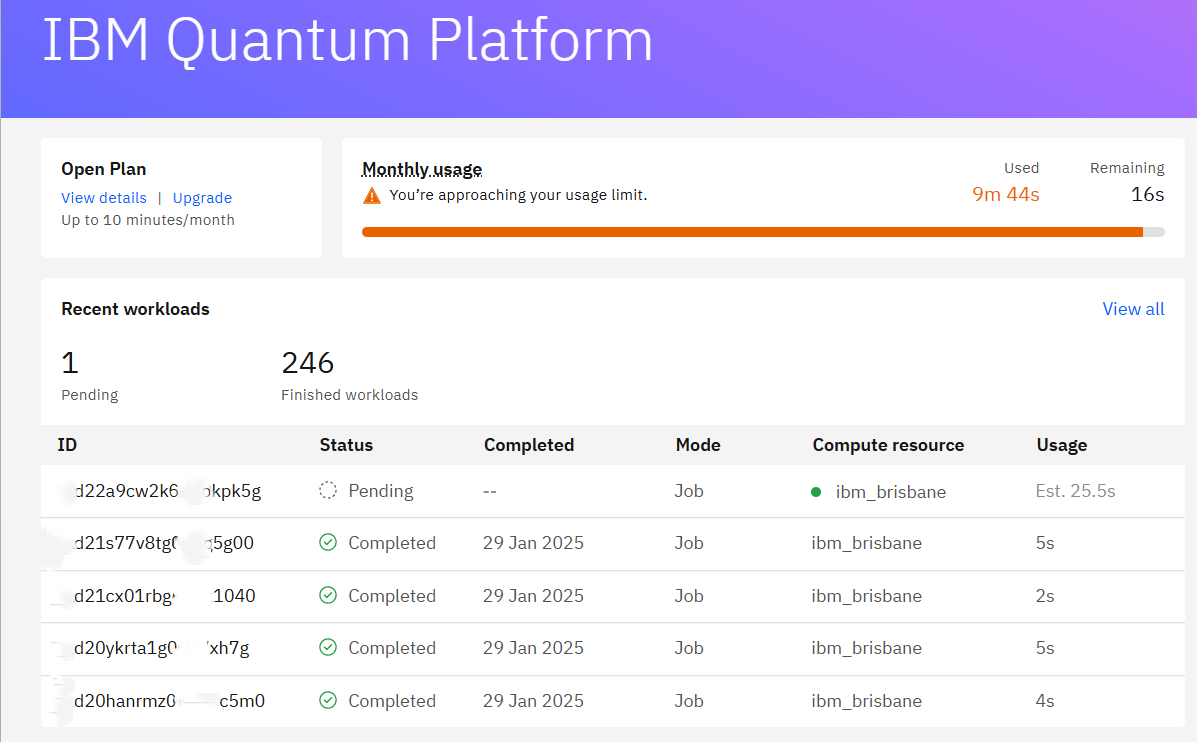}
    \caption{IBM Quantum Platform Dashboard showing all status of jobs pending, canceled, completed etc.}
    \label{fig:ibm_run_dashboard}
\end{figure}

Execution of QFL in an actual quantum computer is possible through providers such as the IBM quantum platform. 
Real quantum experiments can presently be conducted with some ease, as we remain in a preliminary phase of the process pending the widespread availability of quantum hardware.
With limited usage limit, errors like \textit{``Job create exceeds open plan job usage limits"} is common, and execution is one done conservatively. 
However, paid versions exist, which are quite expensive.
Once the tasks are completed, we can observe the dashboard of all the jobs and information about when they were created and completed along with the usage time, as shown in Figure \ref{fig:ibm_run_dashboard}.

\begin{lstlisting}[language=Terminal] 
IBMRuntimeError: 'Failed to run program: \'403 Client Error: Forbidden for url: https://api.quantum.ibm.com/runtime/jobs. {"errors":[{"message":"Job create exceeds open plan job usage limits","code":4317,"solution":"Please wait until the beginning of next month to submit more jobs when your quota will reset.","more_info":"https://docs.quantum-computing.ibm.com/errors"}]}\''
\end{lstlisting}

\subsection{Simulators and Real Quantum Computer}
Quantum experiments can be carried out both in Simulator as well as Real Quantum Computers. 
Simulators are useful for testing and debugging and are free while the real IBM quantum computer incurs fees associated after its free limit usage.
Some of the local simulators provided by \textit{qiskit-ibm-runtime} are fake backends such as \textit{FakeManila} or \textit{Qiskit AerSimulator} backends which can be used to replicate both noise free or noise model based on real quantum hardware.
For real quantum hardware, there are various cloud providers, with IBM providing quantum hardware like \textit{``IBM\_Bribane", ``IBM\_Kyiv" }etc. 
In terms of simulators, \textit{AerSimulator} is great for general purpose, high-performance simulation, while a fake backend such as FakeManila is better for mimicking the specific hardware IBM QPUs by using snapshots.
The number of qubits supported by \textit{AerSimulator} matches to the origin circuit, whereas for the fake backend it supports 5 qubits.
IBM systems like \textit{``IBM\_Brisbane"} support upto 127 qubits.

\begin{figure}[!h]
    \centering
    \begin{subfigure}[b]{0.33\columnwidth}
        \centering
       \includegraphics[width=\linewidth]{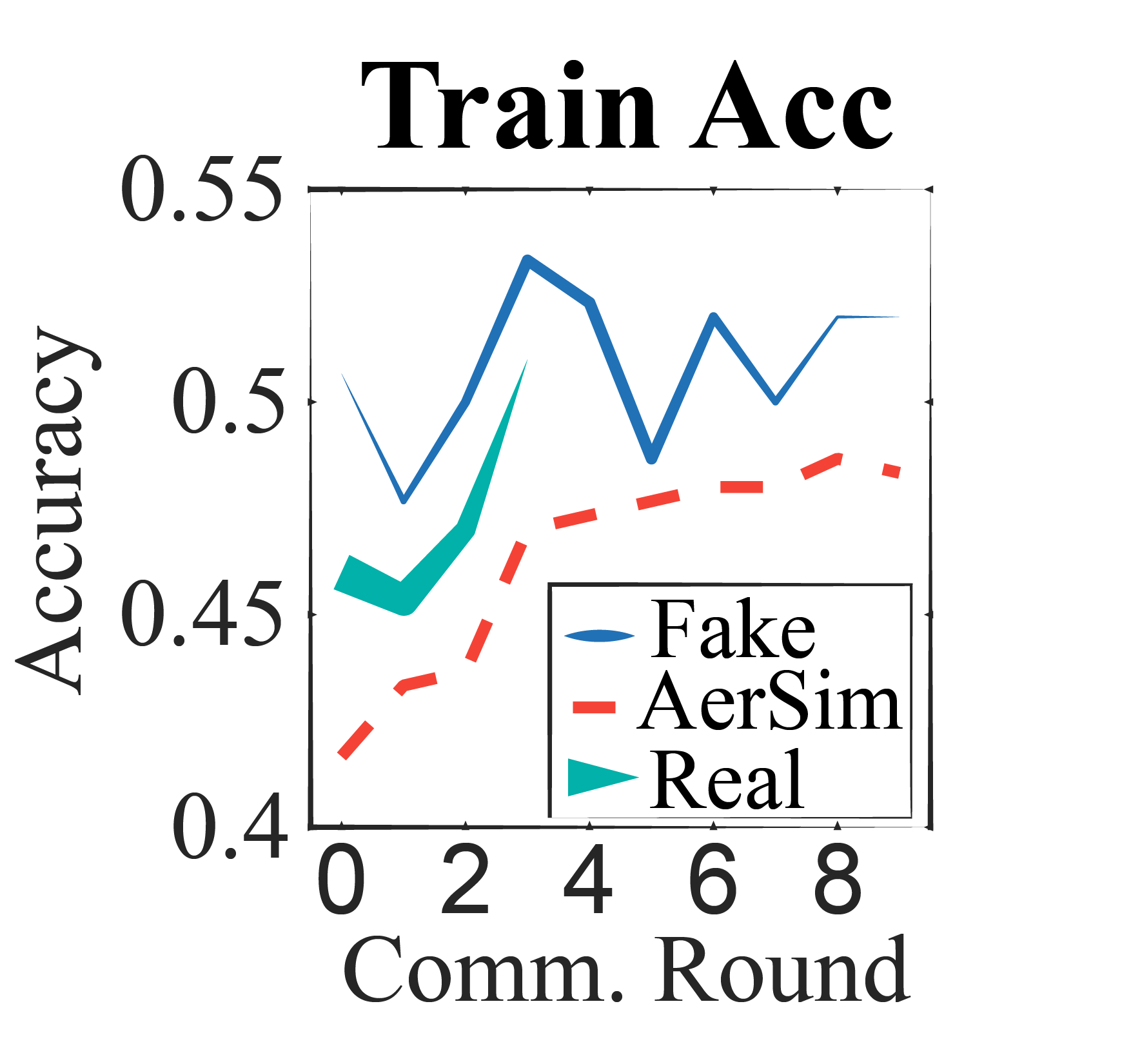}
    \caption{Train Acc}
    \label{fig:avg_train_acc_devices_real}
    \end{subfigure}
     \hspace{-2mm}
    \begin{subfigure}[b]{0.33\columnwidth}
        \centering
       \includegraphics[width=\linewidth]{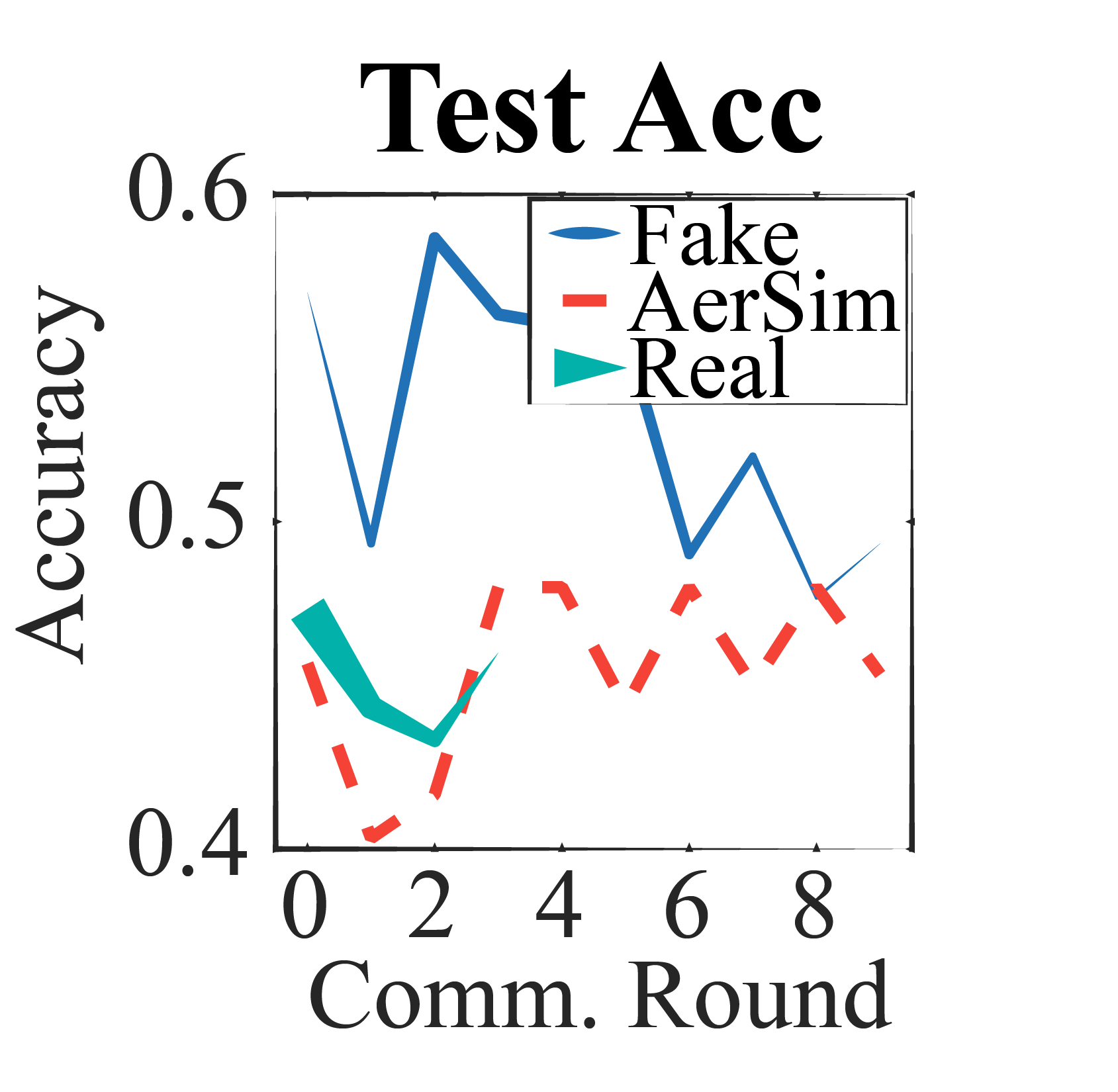}
    \caption{Test Acc}
    \label{fig:avg_test_acc_devices_real}
    \end{subfigure}
     \hspace{-2mm}
  \begin{subfigure}[b]{0.33\columnwidth}
        \centering
       \includegraphics[width=\linewidth]{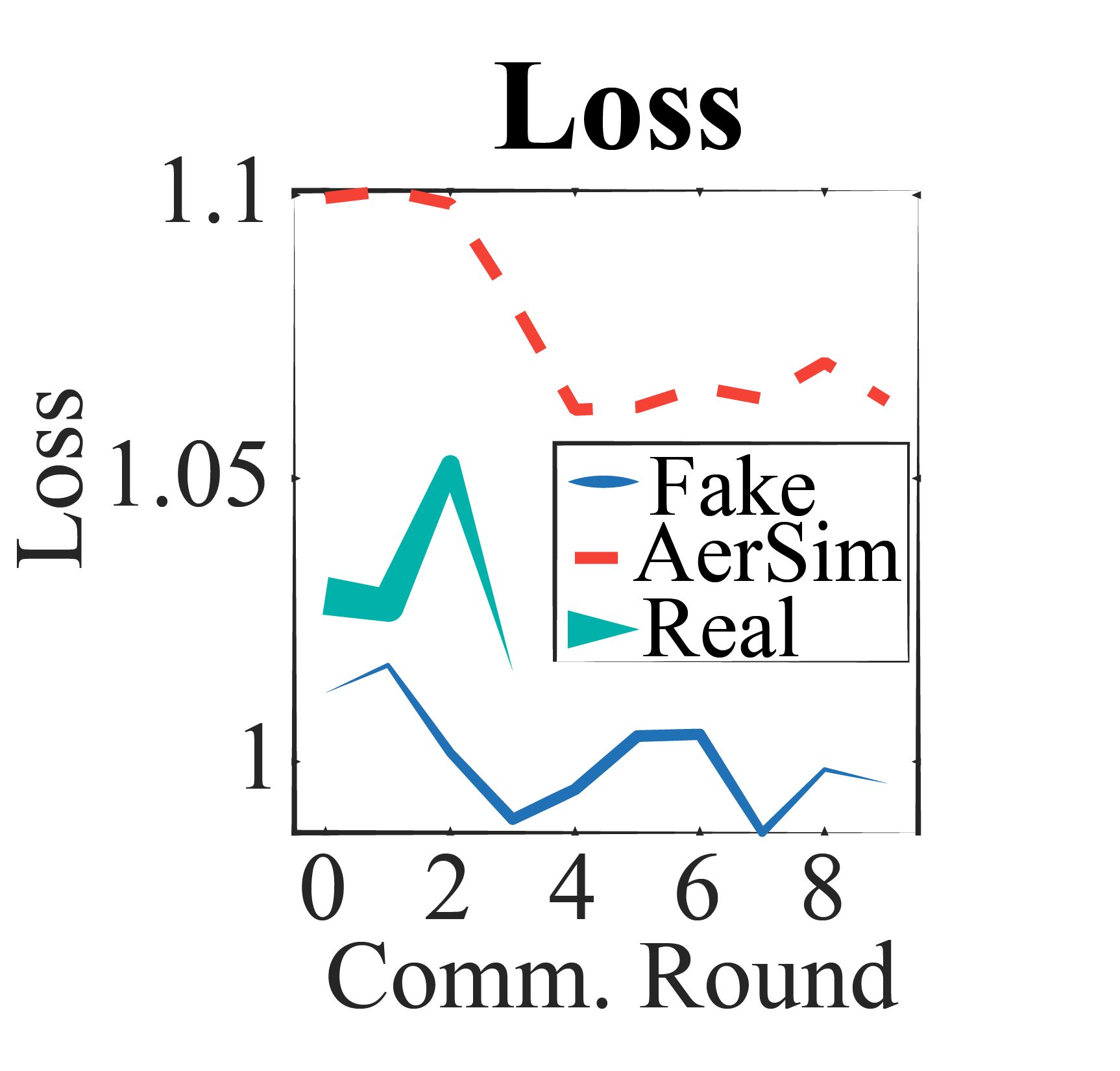}
    \caption{Train Loss}
    \label{fig:avg_loss_devices_real}
    \end{subfigure}
    \caption{Average Device Performance: Simulators vs Real Quantum Computer}
    \label{fig:device_avg_performance_device_3devices_r}
\end{figure}

Average devices performance comparison between results from real quantum hardware (Real) and simulators (Fake, AerSim) are shown in Figures \ref{fig:avg_train_acc_devices_real}, 
\ref{fig:avg_test_acc_devices_real} and \ref{fig:avg_loss_devices_real} for average train accuracy, average test accuracy and average loss respectively.
Due to hardware limitations, results from quantum hardware are only for 3 communications rounds. 
We can conclude that the results from real quantum hardware show performance is in between Fake and AerSim results with results in AerSim performing worst. 

\begin{figure}[!htb]
    \centering
    \includegraphics[width=\linewidth]{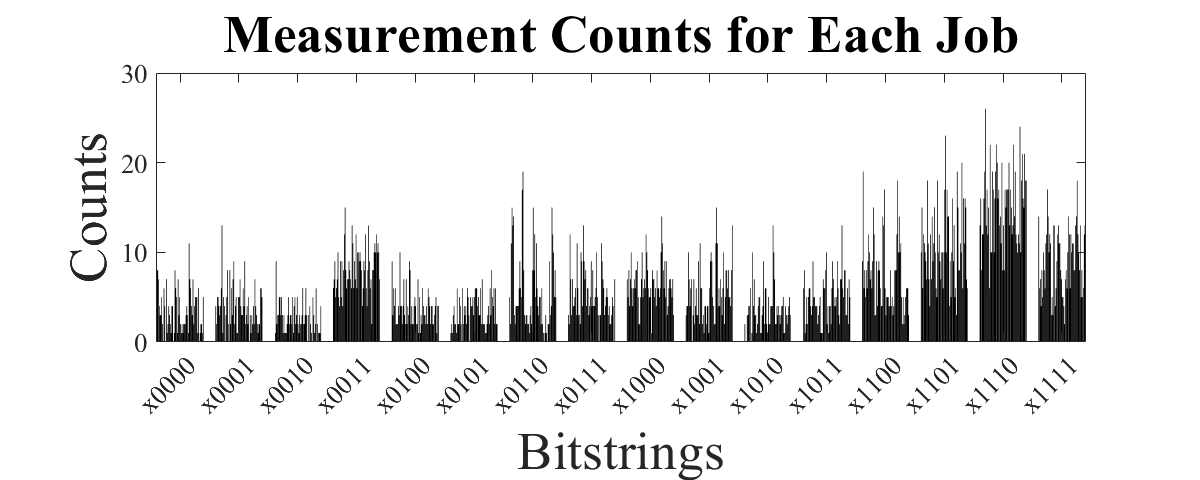}
    \caption{Measurements Counts}
    \label{fig:measurements_counts}
\end{figure}

In Figure \ref{fig:measurements_counts}, we can observe the measurement counts of all the jobs from IBM quantum platform that were created and completed during QFL experimentation on IBM computer.
In Figure \ref{fig:measurements_counts}, we observe the highest count for bitstring ``1110".

\subsection{QCNN}
A typical quantum classical neural network (\textit{QCNN}) consists of alternate layers of convolutional and pooling layers through which the image is passed through, where patterns are detected and associated with a particular label, as shown in Figure \ref{fig:QCNN}.
The important concept is the use of kernel by convolutional layer, which determines the patterns and features of a particular point.
However, pooling layers reduce the dimensionality of the input data which helps to reduce the computational complexity and the number of learning parameters required in \textit{QCNN}. For the classification fully connected layers are used.
The quantum convolutional neural network (\textit{QCNN}) is similar to \textit{CCNN}. 
However, with \textit{QCNN}, the image is first encoded using a feature map such as Qiskit's ZZFeatureMap.
After encoding, alternate convolutional and pooling layers are applied.
In \textit{QCNN}, the purpose of alternating layers is to reduce the dimensionality of the circuit until it is with only one qubit.
The classification is done by measuring the output on this remaining qubit.
In \textit{QCNN}, the convolution layer consists of series of two qubit unitary operators. The parameterized circuits are contained in each layer which are updated with new parameters during the training process.

\begin{figure}[!h]
    \centering
    \includegraphics[width=0.8\linewidth]{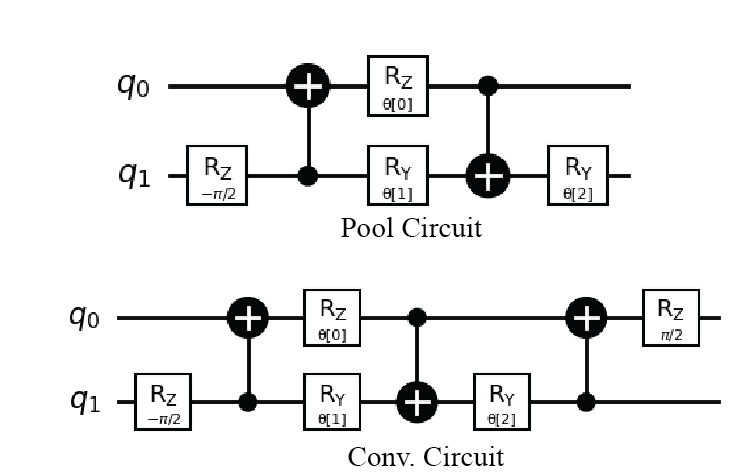}
    \caption{Convolutional and pooling circuit.}
    \label{fig:circuit}
\end{figure}

Each convolutional layer and the pooling layers consists of convolutional and pooling circuits, respectively, as shown in Figure \ref{fig:circuit}.
Both circuit consists of two qubit unitary circuit however, the purpose of the pooling layer is to reduce the dimension like transforming from a two qubit system to one and the convolution circuit consists of parameterized qubit unitary circuit.
Each layer in \textit{QCNN} is tuned with parameters during the training process to minimize loss function and train the \textit{QCNN} for the classification task.

\subsection{Variational Quantum Classifier (\textit{VQC})}
Variational Quantum Classifier (\textit{VQC}) is the simplest classifier available in Qiskit library which is composed of feature map for data encoding, ansatz with trainable parameters etc. as shown in Figure \ref{fig:VQC}.
It is a variational algorithm where the measured bitstrings are used to interpret as the output of the classifier.
\begin{figure}[!h]
    \centering
    \includegraphics[width=0.4\linewidth]{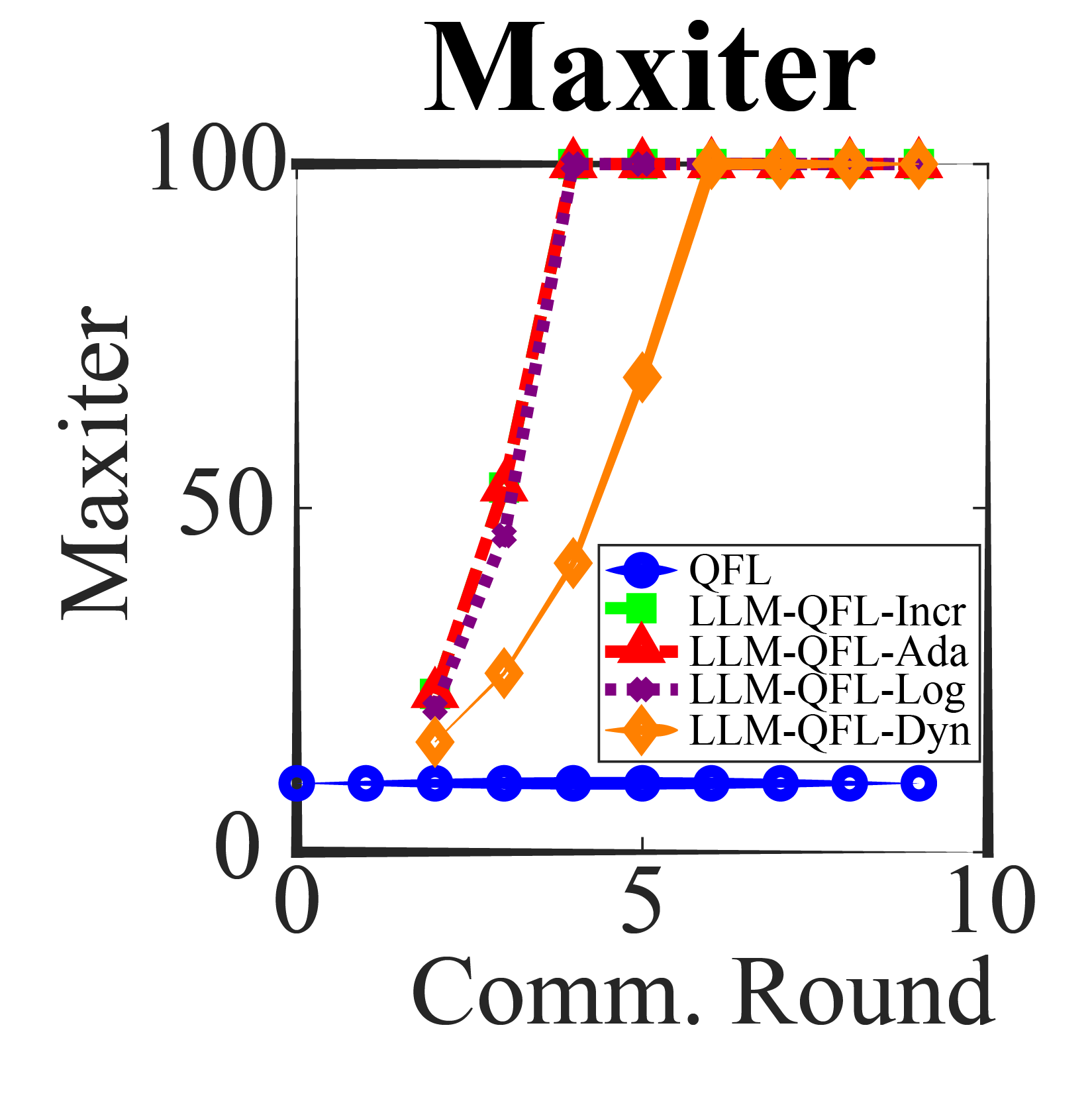}
    \caption{Impact of choice of regularization}
    \label{fig:maxiter_impact_all}
\end{figure}

\subsection{Variation of Maxiter Regularization}
Various adjustment approaches \textit{LLM-QFL-Incr, LLM-QFL-Ada, LLM-QFL-Log, LLM-QFL-Dyn} (representing incremental, adaptive, logarithmic, and dynamic approaches, respectively)
to regulate can be used in various ways such as incremental, dynamic weighted adjustment, logarithmic, adaptive, etc.  
With an incremental approach, we compute the ratio between the current performance of the device and the loss value of the LLM evaluation, calculate the increment to align performance, and increment $maxiter$ gradually.
In dynamic weighted adjustment, the weighted formula is used to dynamically adjust $maxiter$ based on the performance ratio. 
It combines the current $maxiter$ with a weighted adjustment proportional to the ratio with a smooth adjustment to the value.
The logarithmic approach adjusts $maxiter$ using the logarithmic function of the ratio where smaller adjustments are provided for larger ratios.
Finally, with the adaptive approach, it directly scales $maxiter$ by the performance ratio, which is the straightforward proportional adjustment.
The choice of method affects the system as shown in Figure \ref{fig:maxiter_impact_all}.

\subsection{Device Performance: Selection Methods}
\begin{figure}[!h]
    \centering
    \begin{subfigure}[b]{0.32\columnwidth}
        \centering
       \includegraphics[width=\linewidth]{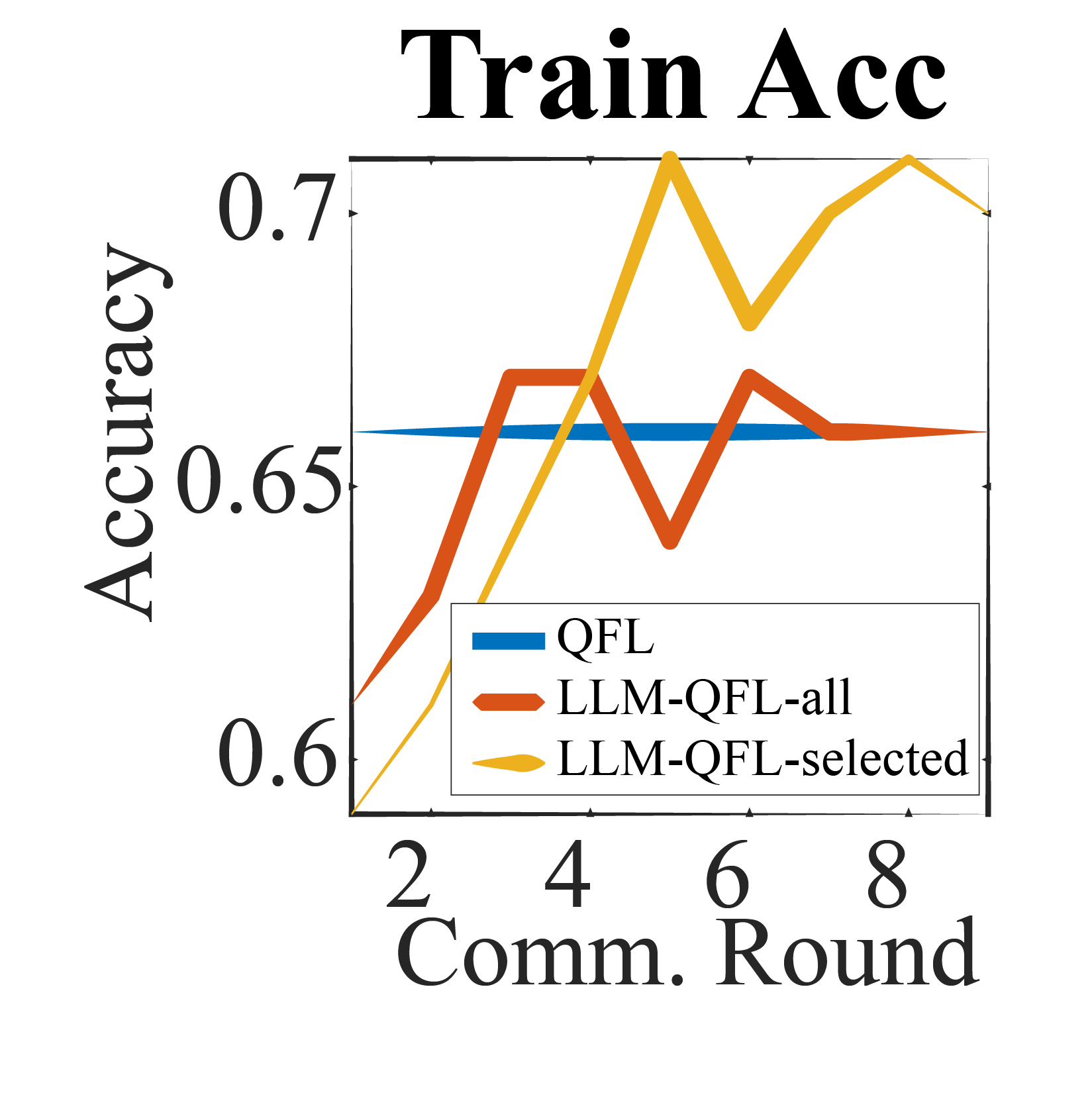}
    \caption{Train Acc.}
    \label{fig:train_acc_device_9}
    \end{subfigure}
    \begin{subfigure}[b]{0.32\columnwidth}
        \centering
       \includegraphics[width=\linewidth]{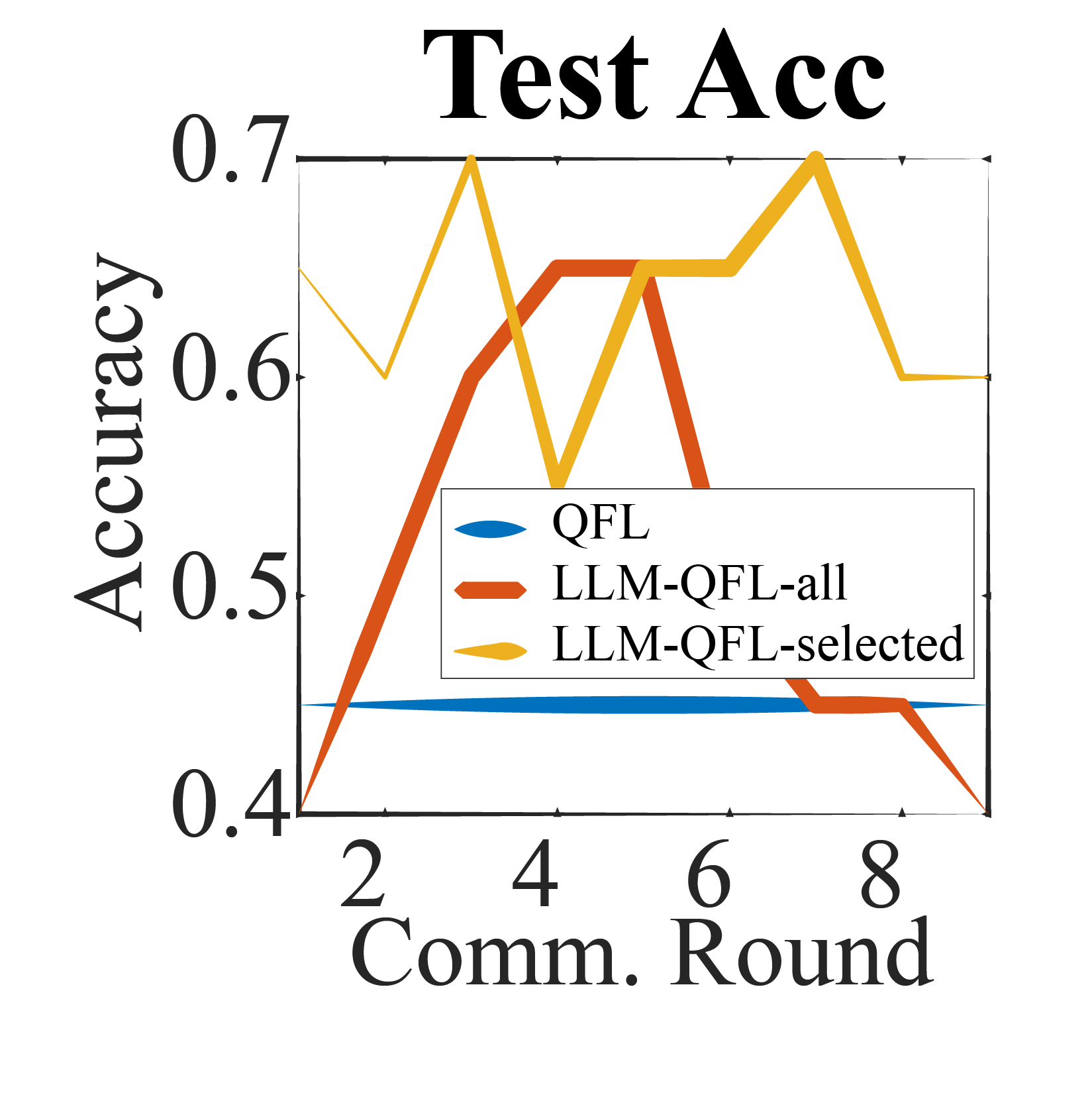}
    \caption{Test Acc.}
    \label{fig:test_acc_device9}
    \end{subfigure}
  \begin{subfigure}[b]{0.32\columnwidth}
        \centering
       \includegraphics[width=\linewidth]{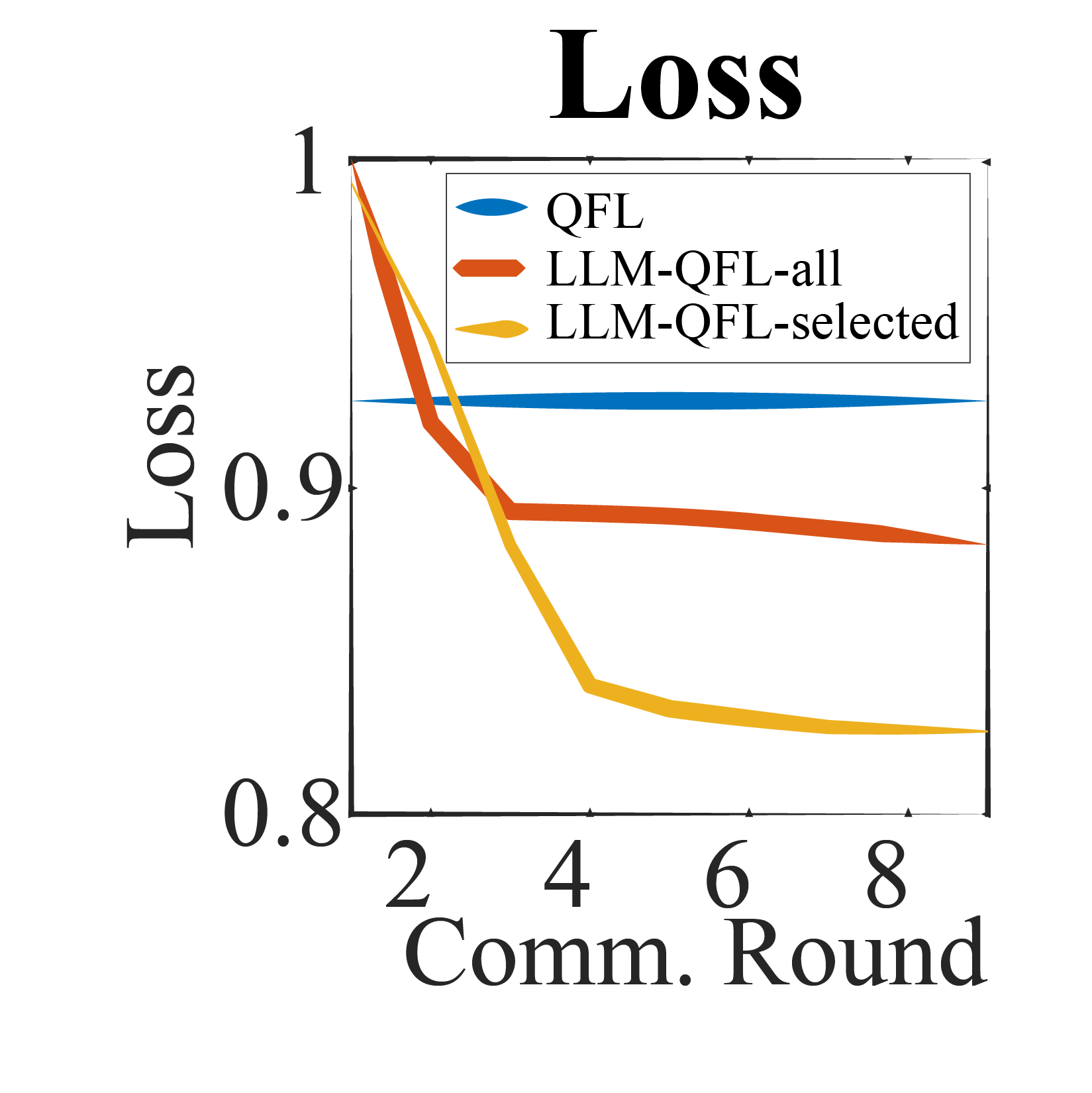}
    \caption{Train Loss.}
    \label{fig:loss_device_9}
    \end{subfigure}
    \caption{Device 9 performance; QFL, LLM-QFL-all vs. LLM-QFL-selected.}
    \label{fig:device_performance_device_3devices}
\end{figure}

\begin{figure}[!h]
    \centering
    \begin{subfigure}[b]{0.33\columnwidth}
        \centering
       \includegraphics[width=\linewidth]{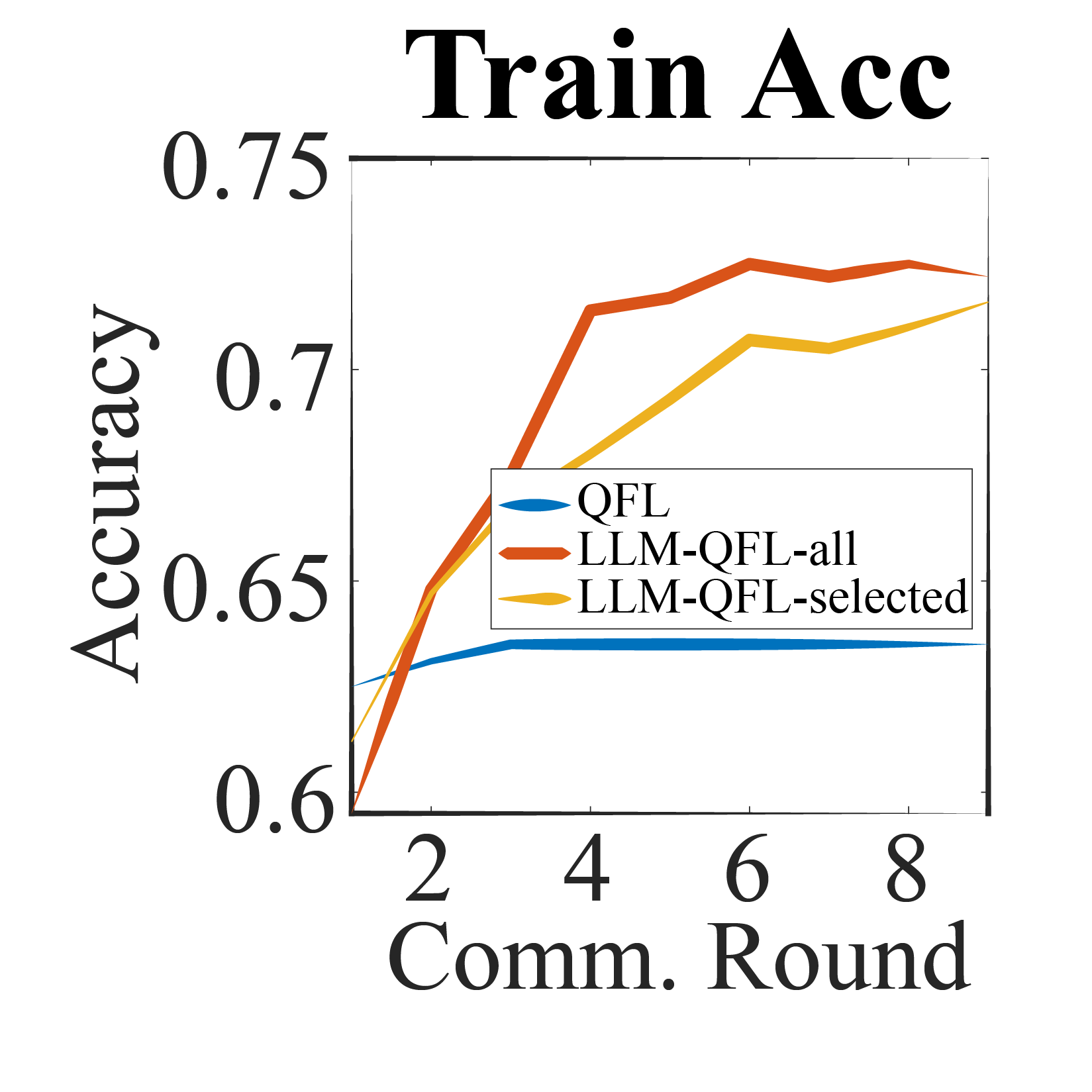}
    \caption{Train Acc.}
    \label{fig:avg_train_acc_devices_genomic}
    \end{subfigure}
    \hspace{-2mm}
    \begin{subfigure}[b]{0.33\columnwidth}
        \centering
       \includegraphics[width=\linewidth]{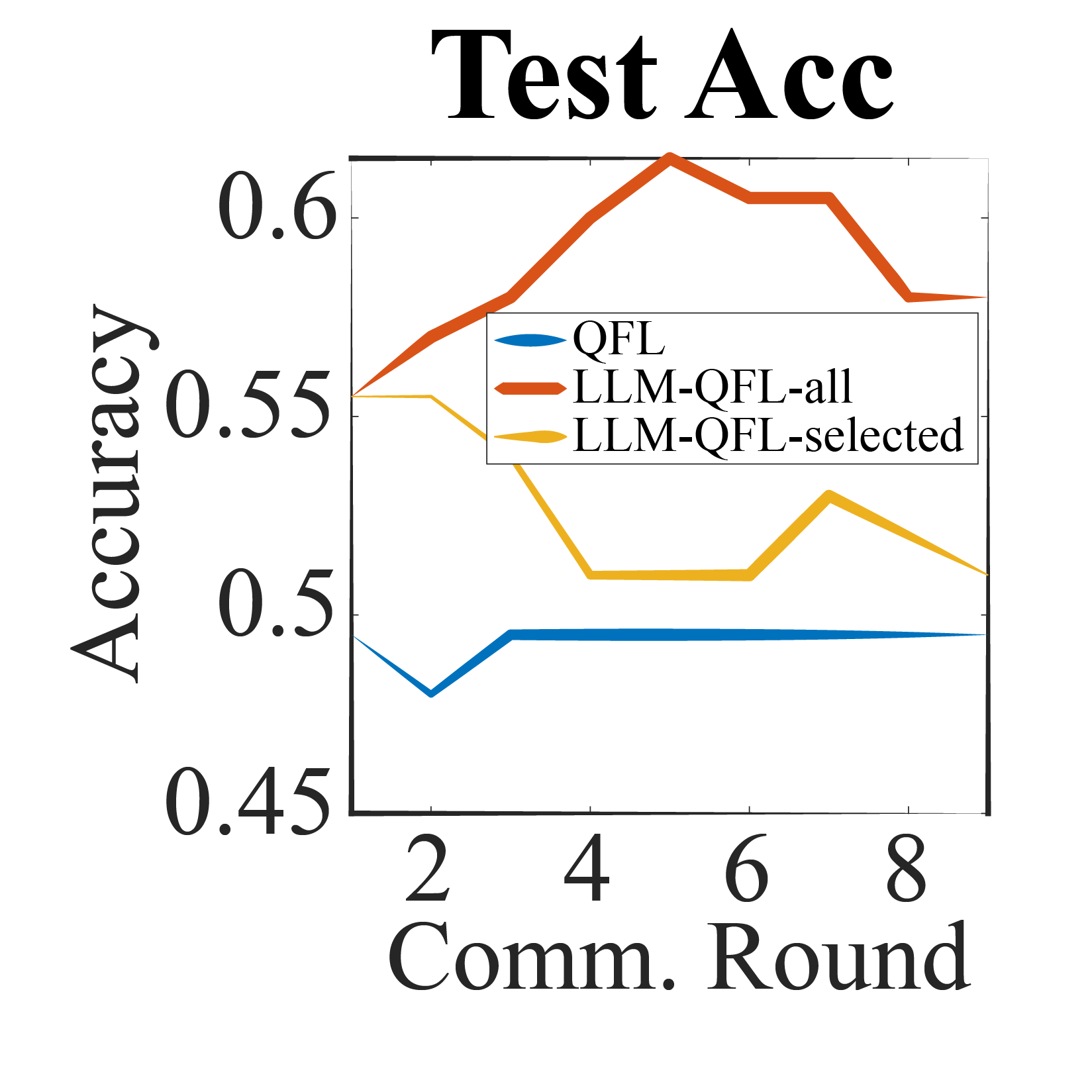}
    \caption{Test Acc.}
    \label{fig:avg_test_acc_devices_genomic}
    \end{subfigure}
    \hspace{-2mm}
  \begin{subfigure}[b]{0.33\columnwidth}
        \centering
       \includegraphics[width=\linewidth]{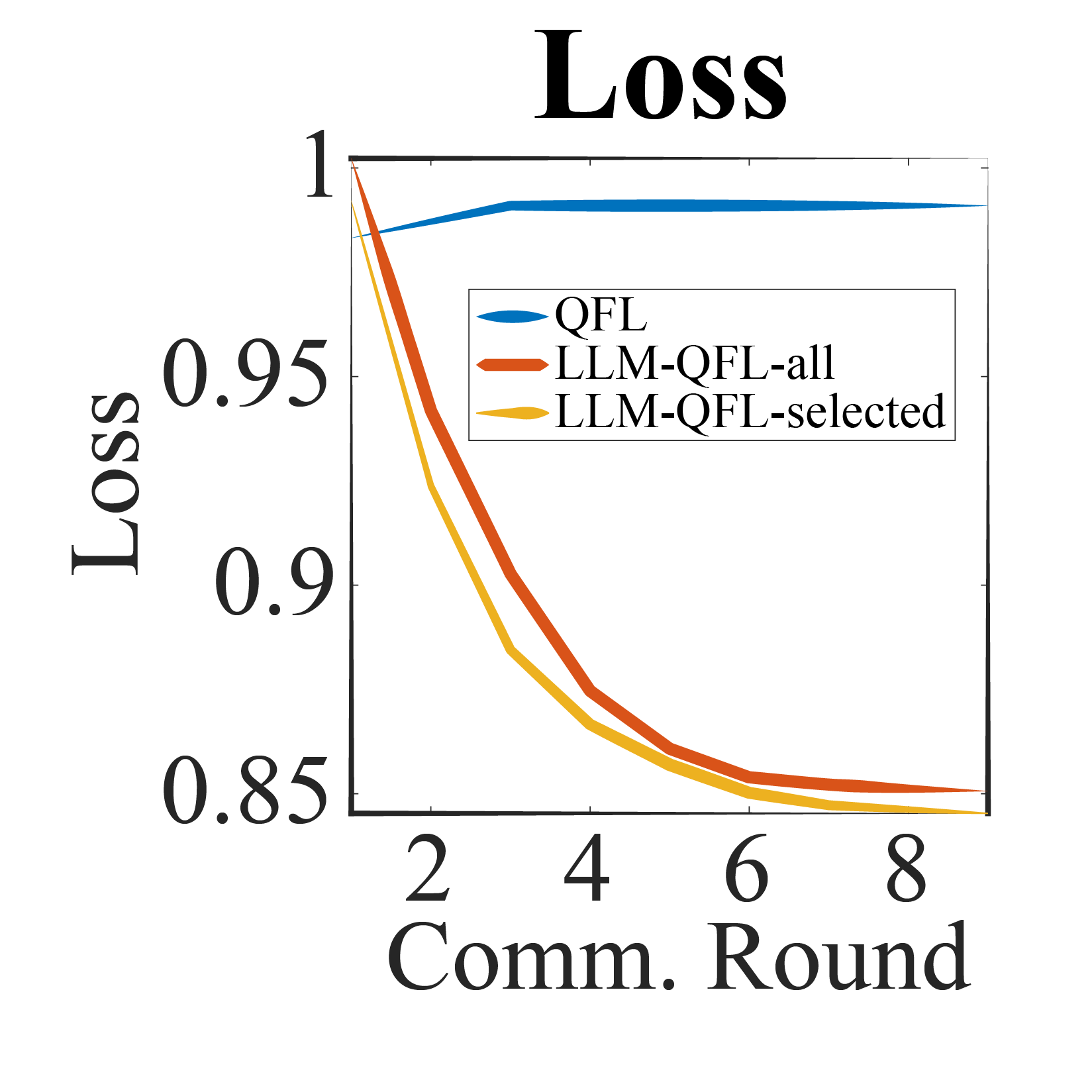}
    \caption{Train Loss.}
    \label{fig:avg_loss_devices_genomic}
    \end{subfigure}
    \caption{Average Device performance; QFL, LLM-QFL-all vs. LLM-QFL-selected.}
    \label{fig:avg_device_performance_device_3devices_11}
\end{figure}

In Figure \ref{fig:device_performance_device_3devices}, the results can be observed for Device 9 with LLM integrated QFL outperforming default QFL in terms of training, test accuracy, and training loss, while the average performance of all devices can be seen in Figure \ref{fig:avg_device_performance_device_3devices_11} following a similar pattern.

\subsection{LLM vs QFL comparison}
\begin{figure}[!h]
    \centering
    \begin{subfigure}[b]{0.4\columnwidth}
        \centering
       \includegraphics[width=\linewidth]{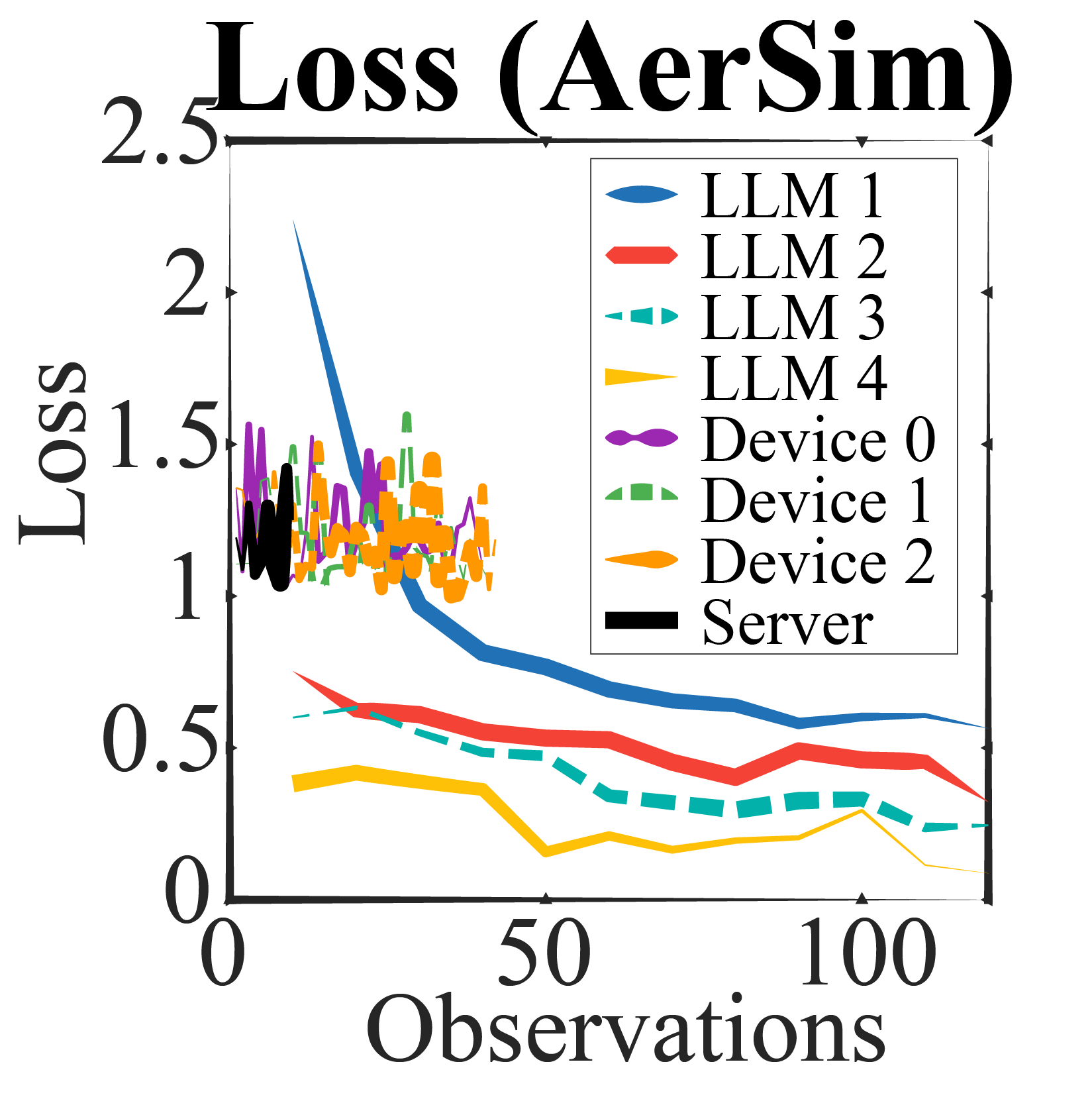}
    \caption{AerSimulator}
    \label{fig:loss_aersim}
    \end{subfigure}
    \begin{subfigure}[b]{0.4\columnwidth}
        \centering
       \includegraphics[width=\linewidth]{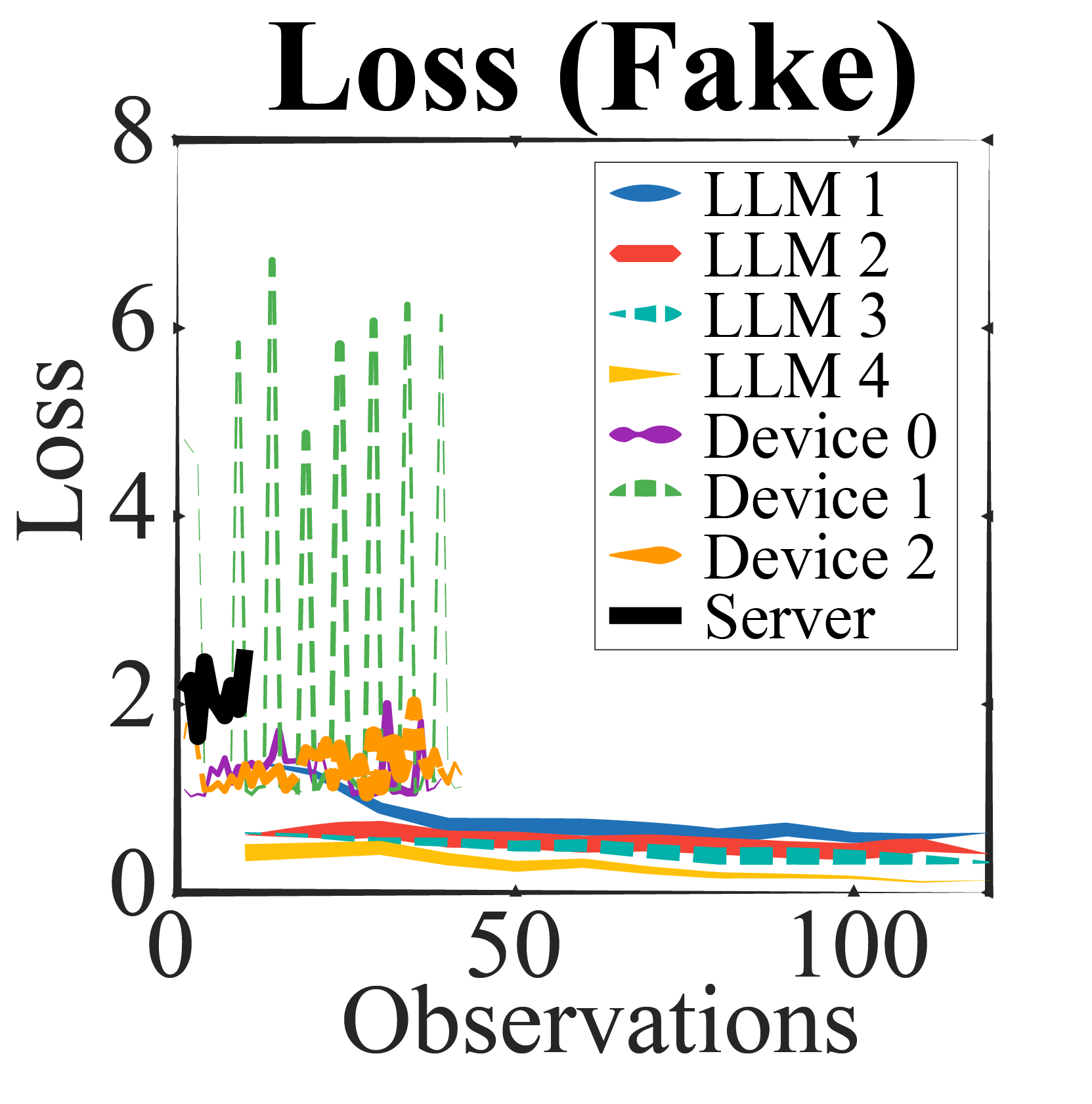}
    \caption{FakeManila}
    \label{fig:loss_fakemanila}
    \end{subfigure}
    \caption{Device performance comparison of results with and without LLM into QFL; Using LLaMA; AerSim, FakeManila}
    \label{fig:llm_performance_comparion_simulators}
\end{figure}

In this work, we fine-tune LLM models and use their performance to regulate and further manage the QFL algorithm autonomously in the direction and with the intention of improving the performance of the system.
In Figure \ref{fig:llm_performance_comparion_simulators}, we highlight the comparison between LLM performance, client device performance and server device performance in terms of objective values in all cases of simulation results in AerSimulator and FakeManila 
in Figures \ref{fig:loss_aersim} and \ref{fig:loss_fakemanila} 
respectively.
We can clearly observe that, for all three cases, the performance of LLM is unmatched. 
This demands more work towards study in the field of QFL or the field of quantum machine learning to advance the performance benchmarks to match classical LLM models.

\begin{figure}[!h]
    \centering
    \begin{subfigure}[b]{0.32\columnwidth}
        \centering
       \includegraphics[width=\linewidth]{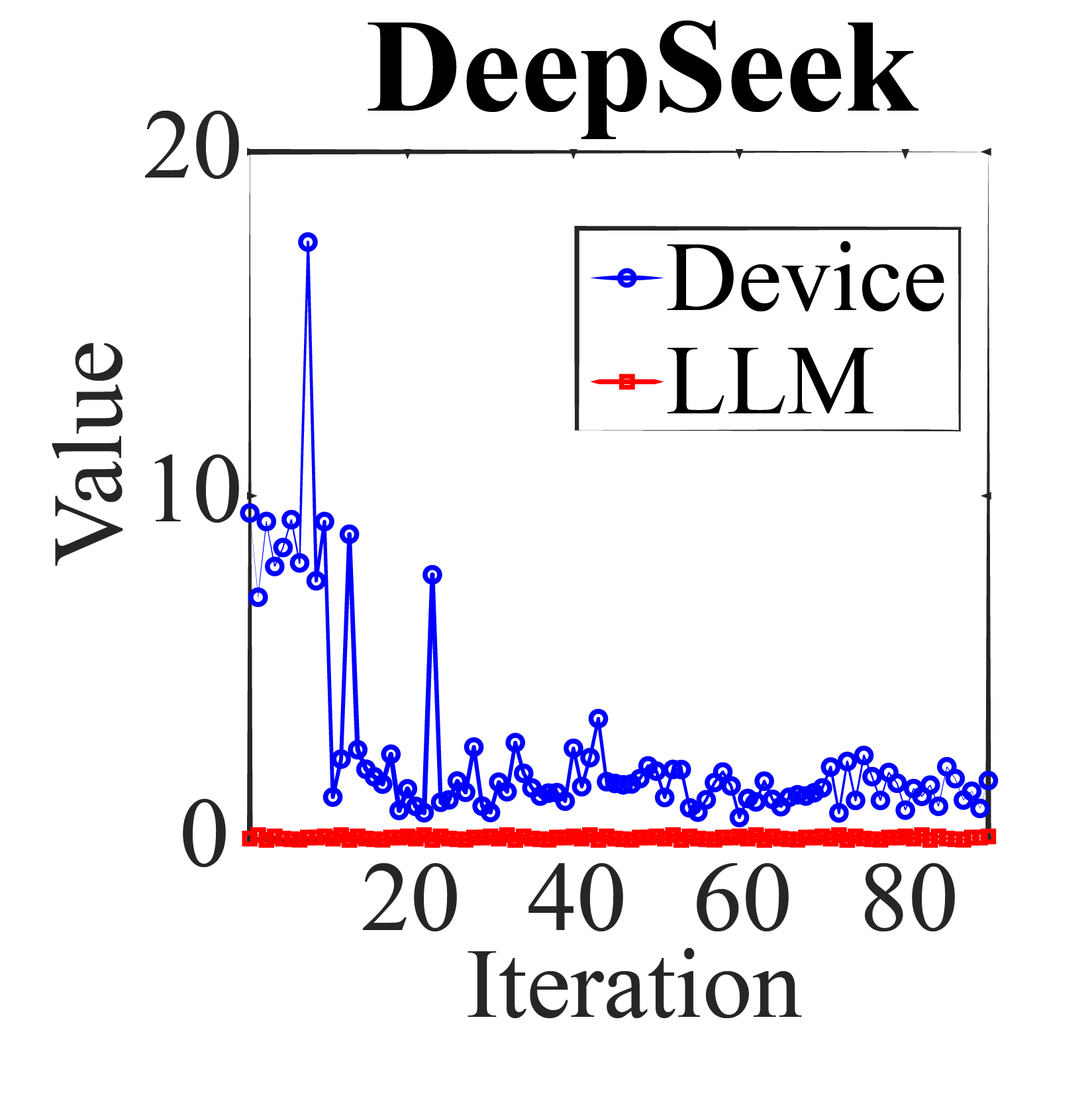}
    \caption{DeepSeek}
    \label{fig:llm_comparison_deepseek}
    \end{subfigure}
    \begin{subfigure}[b]{0.32\columnwidth}
        \centering
       \includegraphics[width=\linewidth]{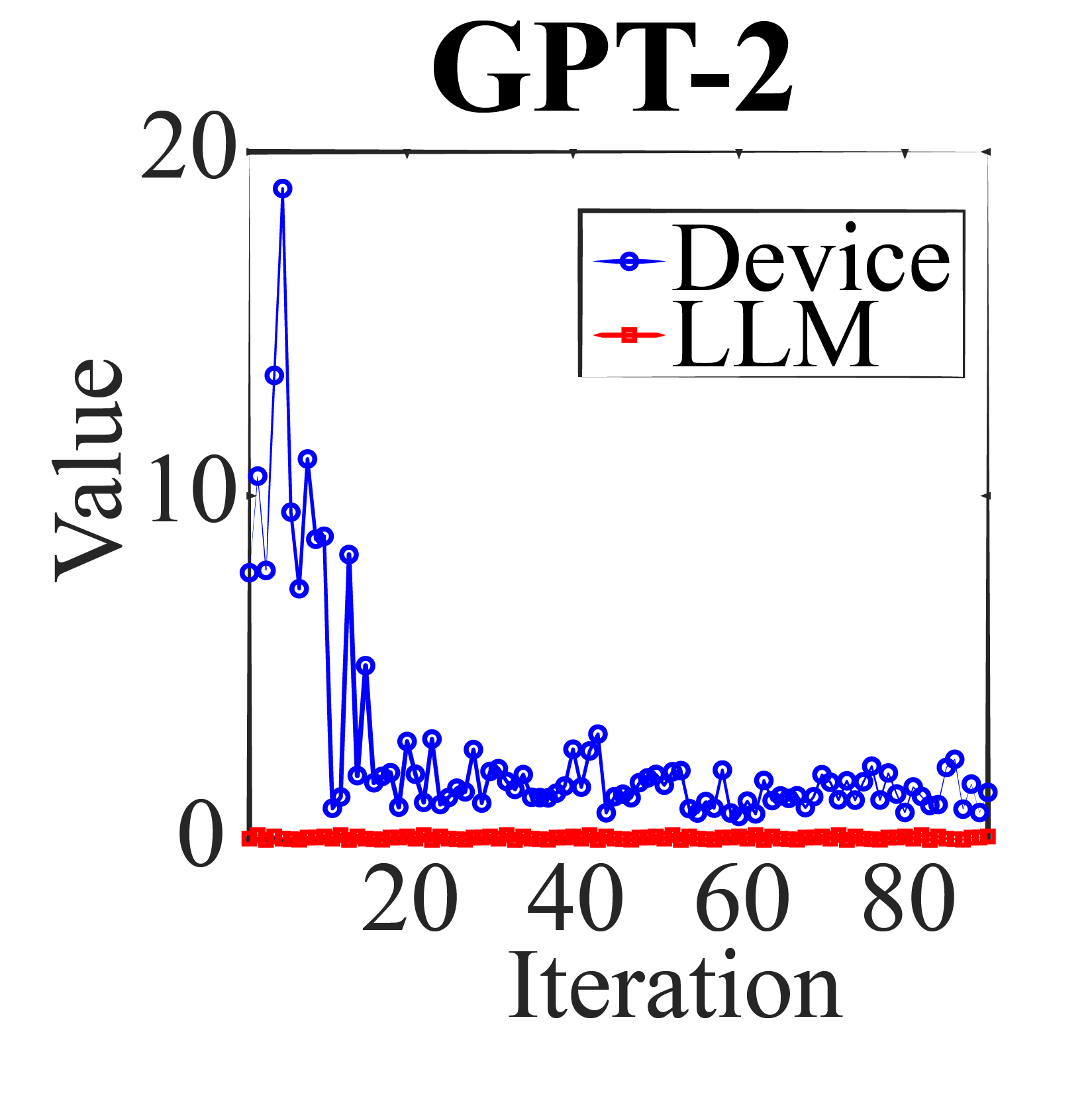}
    \caption{GPT-2}
    \label{fig:llm_comparison_gpt2}
    \end{subfigure}
  \begin{subfigure}[b]{0.32\columnwidth}
        \centering
       \includegraphics[width=\linewidth]{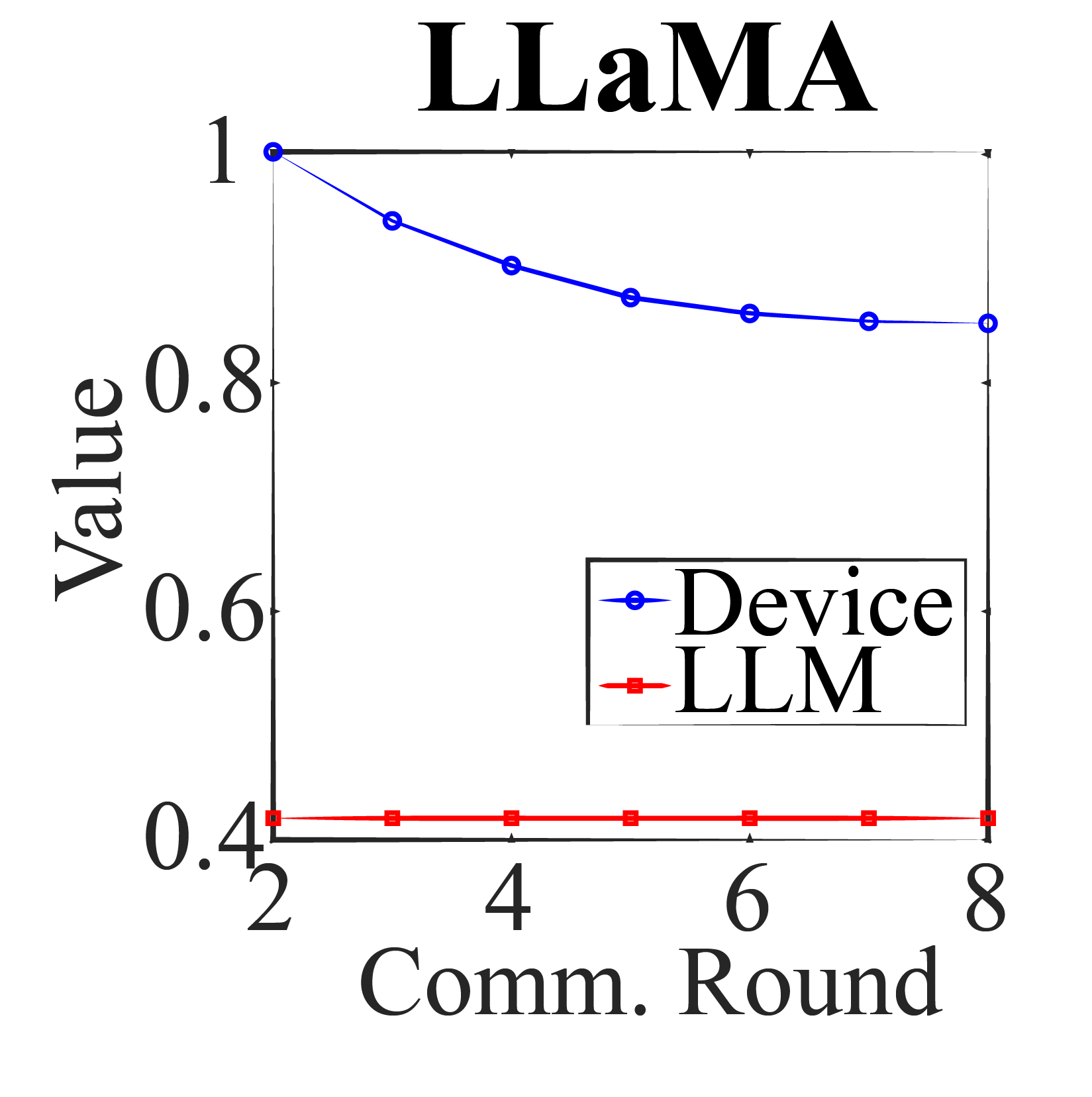}
    \caption{LLaMA}
    \label{fig:llm_comparison_llama}
    \end{subfigure}
    \caption{Device loss and LLM f1 Score.}
    \label{fig:llm_performance_comparion_loss}
\end{figure}

Further we present a comparison between fine-tuning pre-trained LLM models in QFL; LLaMA-3.2-1B, GPT-2 and DeepSeek-llm-7b-base.
We plot initial fine-tuning performance f1 score and their impact on device performance
as shown in Figure \ref{fig:llm_performance_comparion_loss}.

\subsection{Other Results}
\begin{figure}[!h]
    \centering
    \includegraphics[width=0.45\linewidth]{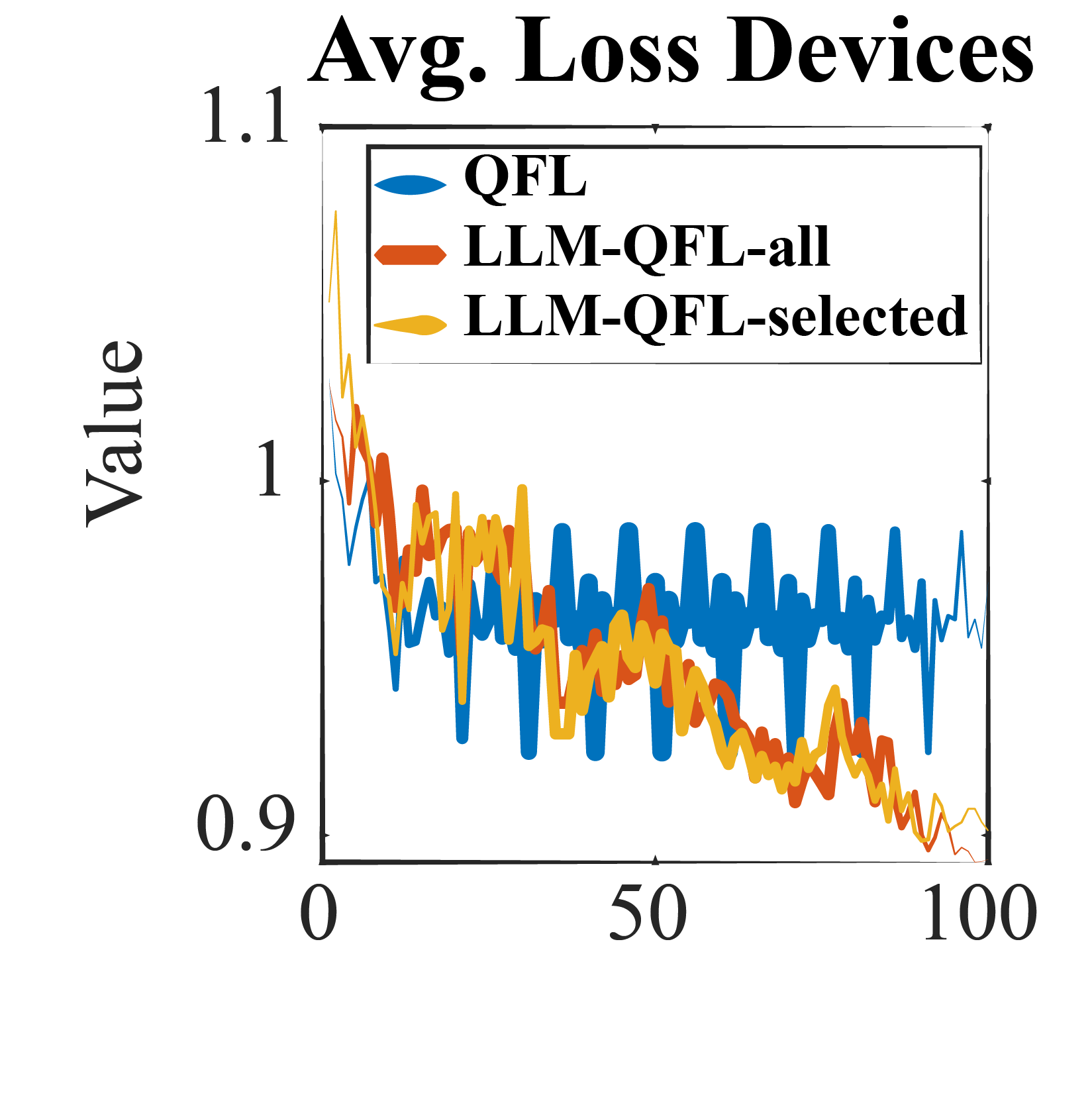}
    \caption{Average Device performance: QFL vs. LLM-QFL}
    \label{fig:device_objective_vqc_genomics}
\end{figure}

Figure \ref{fig:device_objective_vqc_genomics} shows a clear distinction in the impact of LLM integration into the QFL network.
By maintaining a fixed initial optimizer maxiter value of 10, the average device performance remains stagnant without achieving actual convergence throughout the communication rounds, necessitating additional rounds. On the other hand, LLM-QFL achieves better convergence with fewer communication rounds by adjusting the optimizer value as needed, allowing for more automatic iterations.

\begin{figure}
    \centering
    \includegraphics[width=0.5\linewidth]{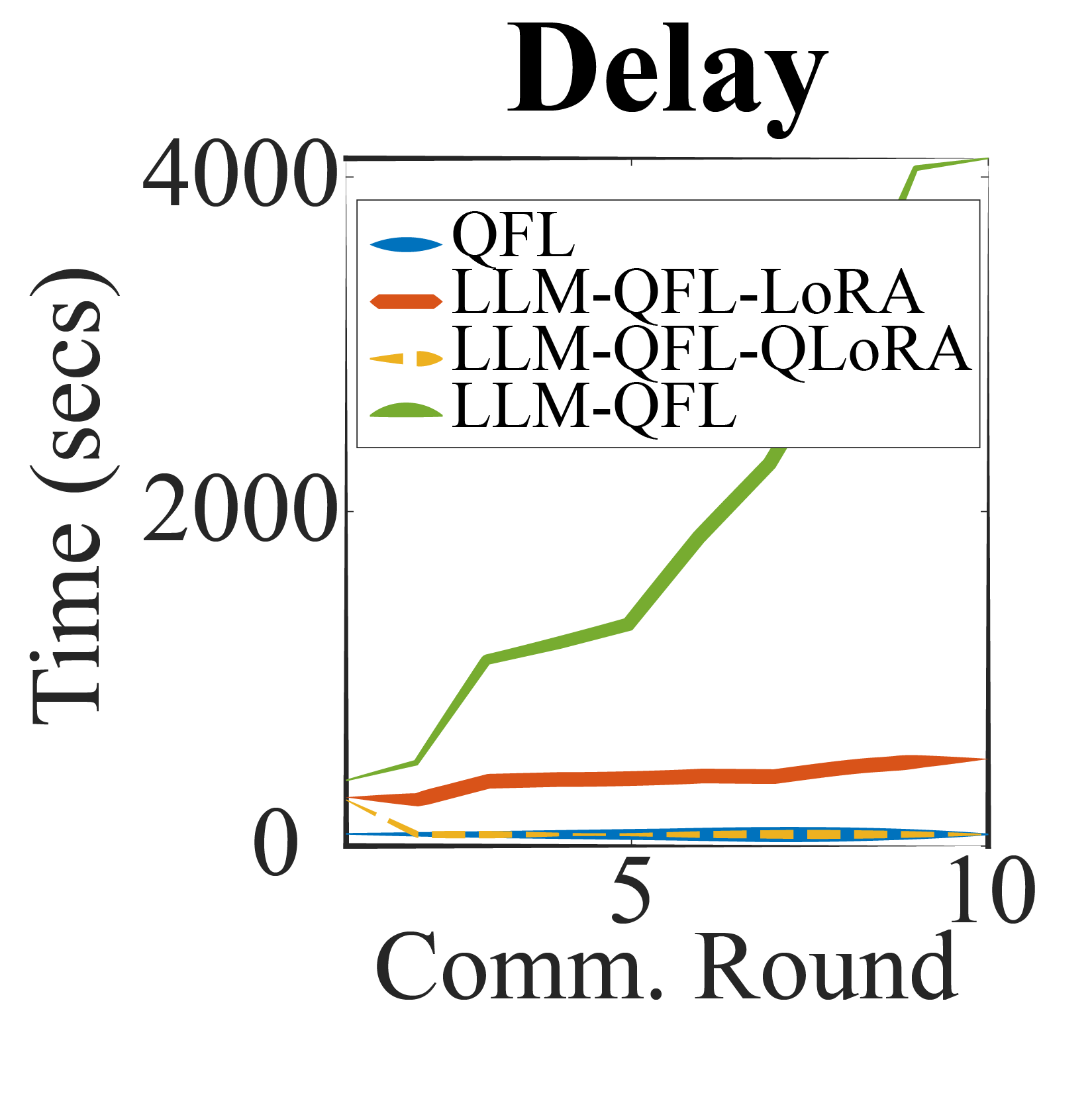}
    \caption{Communication Cost}
    \label{fig:communication_cost}
\end{figure}

In Figure \ref{fig:communication_cost}, we show the difference between QFL and LLM-QFL in terms of communication costs. With LLM-QFL we stop the communication round based on the convergence achieved. However, if we were to let all the communication rounds complete, then as shown in Figure \ref{fig:communication_cost}, we can observe the with LLM-QFL due to increase in the number of optimizer maxiter value, to complete each communication round requires more time than with default QFL.
Nevertheless, the result with LLM-QFL-QLoRA shows similar result with QFL indicating that faster fine tuning has less impact on the communication cost.
However, the performance achieved within first few communication rounds with LLM-QFL and with only QFL will be different with LLM-QFL performing better as concluded in this work.

\subsection{Selection of base models}
In this work, we have selected LLM based models as Meta-LLAMA 3.2-1B, GPT-2, and DeepSeek-LLM-7B-Base.
Their selection was based on their availability, licensing restrictions, computational need and their popularity.
In terms of size, we have chosen all of the models with the least parameters, such as LLAMA with 1B parameters, DeepSeek with 7B parameters. Similarly, GPT-2 model also have 1.5 billion parameters. One of the main reasons, we chose least parameters models is our experiments were carried out in Google Colab pro plus with
varying QPU/GPU configurations, including the A100 GPU
(40 GB) and T4 GPU (15 GB). 
LLaMA 3.2-1B is a lightweight yet efficient model is computationally efficient and suitable for resource constraint settings, GPT-2 is old widely recognized model but well documented model as standard reference point due to historical significance and DeepSeek-LLM-7B-Base was a representation model for DeepSeek, which is substantially larger.

\subsection{IBM Jobs}
\begin{figure}[!htb]
    \centering
    \includegraphics[width=\linewidth]{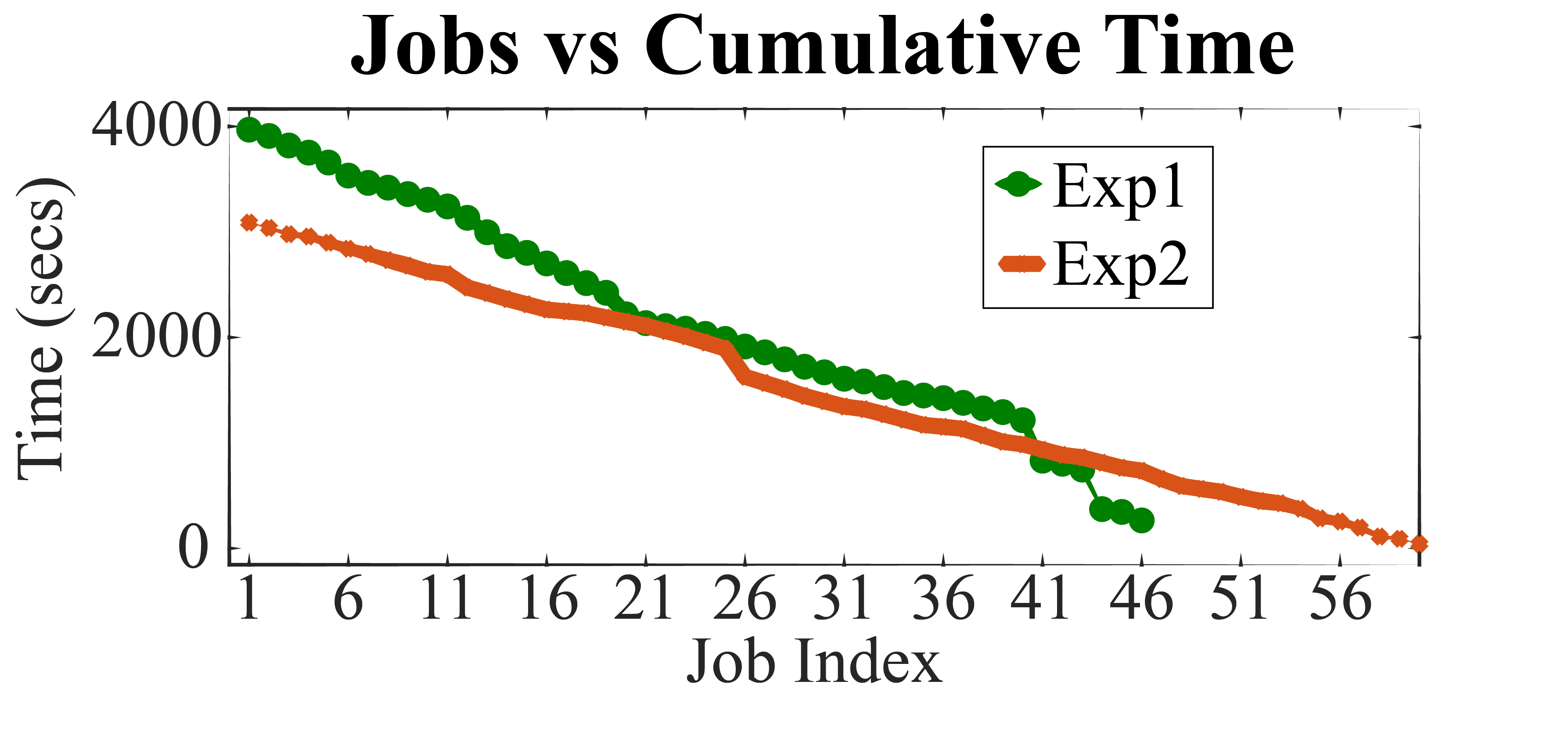}
    \caption{Cumulative Time comparison for two separate experiments on IBM}
    \label{fig:cumulative_time_both_exp}
\end{figure}

Figure \ref{fig:cumulative_time_both_exp} shows results in terms of the cumulative time taken by jobs to start and finish and their trend throughout job executions on the IBM computer.
This shows that there are subtle variations in the way jobs are executed on the IBM computer.

\end{document}